\documentclass[Afour,sageh,times]{sagej}

\usepackage{moreverb,url}

\usepackage[colorlinks,bookmarksopen,bookmarksnumbered,citecolor=red,urlcolor=red]{hyperref}

\usepackage{graphicx}
\usepackage{float}
\usepackage{epsfig} 
\usepackage{wrapfig}
\usepackage{amsmath} 
\usepackage{amssymb}  
\usepackage{amsthm} 
\usepackage{bm}  
\usepackage{subcaption}
\usepackage{caption}
\usepackage{varwidth}
\DeclareCaptionFormat{varwidth}{%
	\begin{varwidth}{\linewidth}#1#2#3\end{varwidth}%
}
\usepackage{color}
\usepackage[toc,page]{appendix}
\newcommand\BibTeX{{\rmfamily B\kern-.05em \textsc{i\kern-.025em b}\kern-.08em
T\kern-.1667em\lower.7ex\hbox{E}\kern-.125emX}}

\def\volumeyear{2019}

\graphicspath{{Figures/}}

\newcommand{\trsp}{\mathsf{T}}
\newcommand{\ty}[1]{{\scriptscriptstyle{\mathcal{#1}}}}
\newcommand{\tr}{\mathrm{tr}}

\begin{document}

\runninghead{Geometry-aware Manipulability Transfer}

\title{Geometry-aware Manipulability\\ Learning, Tracking and Transfer}

\author{No\'{e}mie Jaquier\affilnum{1}, Leonel Rozo\affilnum{2,3}, Darwin G. Caldwell\affilnum{3} and Sylvain Calinon\affilnum{1}}

\affiliation{\affilnum{1}Idiap Research Institute, CH-1920 Martigny, Switzerland\\
\affilnum{2}Bosch Center for Artificial Intelligence, 71272 Renningen, Germany\\
\affilnum{3}Department of Advanced Robotics, Istituto Italiano di Tecnologia, 16163 Genova, Italy}

\corrauth{No\'{e}mie Jaquier, Idiap Research Institute, CH-1920 Martigny, Switzerland}

\email{noemie.jaquier@idiap.ch}

\begin{abstract}
Body posture influences human and robots performance in manipulation tasks, as appropriate poses facilitate motion or force exertion along different axes. In robotics, manipulability ellipsoids arise as a powerful descriptor to analyze, control and design the robot dexterity as a function of the articulatory joint configuration. This descriptor can be designed according to different task requirements, such as tracking a desired position or apply a specific force. In this context, this article presents a novel \emph{manipulability transfer} framework, a method that allows robots to learn and reproduce manipulability ellipsoids from expert demonstrations. The proposed learning scheme is built on a tensor-based formulation of a Gaussian mixture model that takes into account that manipulability ellipsoids lie on the manifold of symmetric positive definite matrices. Learning is coupled with a geometry-aware tracking controller allowing robots to follow a desired profile of manipulability ellipsoids. Extensive evaluations in simulation with redundant manipulators, a robotic hand and humanoids agents, as well as an experiment with two real dual-arm systems validate the feasibility of the approach.
\end{abstract}


\maketitle

\section{Introduction}
\label{sec:Intro}
\begin{figure}[tbp]
	\centering
	\begin{subfigure}[b]{0.24\textwidth}
		\centering
		\includegraphics[width=0.67\textwidth]{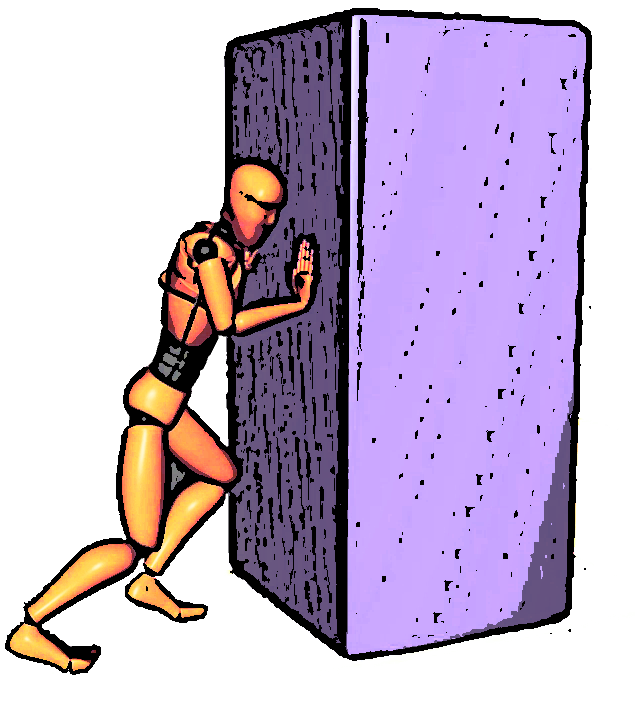}
		\includegraphics[width=0.92\textwidth]{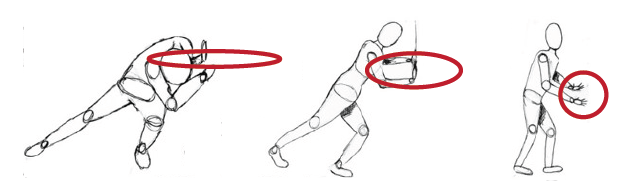}
		\caption{}
		\label{subFig:pushing}
	\end{subfigure}
	\begin{subfigure}[b]{0.24\textwidth}
		\centering
		\includegraphics[width=0.8\textwidth]{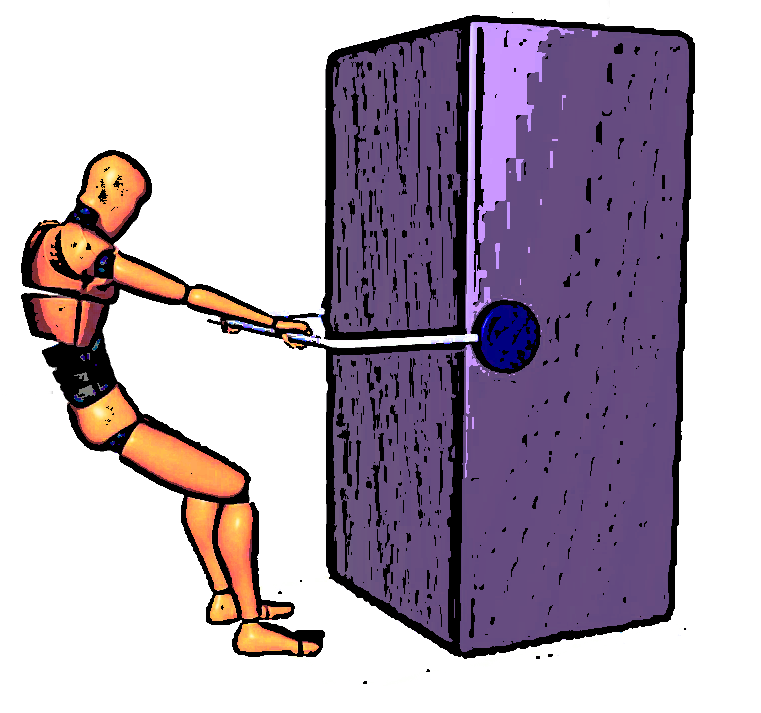}
		\includegraphics[width=0.92\textwidth]{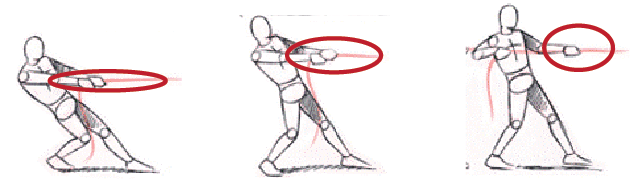}
		\caption{}
		\label{subFig:pulling}
	\end{subfigure}
	\caption{Illustration of pushing (\emph{a}) and pulling (\emph{b}) tasks for which the posture of the human significantly influences his/her ability to carry out the task.}
	\label{Fig:Motivation}
\end{figure}

When we perform a manipulation task, we naturally place our arms (and body) in a posture that is best suited to carry out the task at hand (see Fig.~\ref{Fig:Motivation}). Such posture variation is a means through which the motion and strength characteristics of the arms are made compatible with the task requirements. For example, human arm kinematics plays a central role when humans plan point-to-point reaching movements, where joint trajectory patterns arise as a function of the visual target~\citep{Morasso81:SpatialControl}, indicating that the task requirements influence the human arm posture. This insight was also identified in more complex situations, where not only kinematic but also other biomechanic factors affect the task planning~\citep{Cos11:BiomechanicsInfluence}. For example, the human central nervous system plans arm movements considering its directional sensitivity, which is directly related to the arm posture~\citep{Sabes97:MotorPlanning}. This allows humans to be mechanically resistant to potential perturbations coming from obstacles occupying the workspace. Interestingly, directional preferences of human arm movements are characterized by a tendency to exploit interaction torques for movement production at the shoulder or elbow, indicating that the preferred directions are largely determined by biomechanical factors~\citep{Dounskaia14:PreferDirections}. 

The robotics community has also been aware of the impact of robot posture on reaching movements and manipulation tasks (e.g., pushing, pulling, reaching). It is well known that by varying the posture of a robot, we can change the optimal directions for generating motion or applying specific forces. This has direct implications in hybrid control, since the controller capability can be fully realized when the optimal directions for controlling velocity and force coincide with those dictated by the task~\citep{Chiu87:RedundantRbtCtrl}. In this context, the so-called manipulability ellipsoid~\citep{Yoshikawa85:Manipulabilty} serves as a geometric descriptor that indicates the ability to arbitrarily perform motion and exert a force along the different task directions in a given joint configuration. 

Manipulability ellipsoids have been used to measure the compatibility of robot postures with respect to fine and coarse manipulation~\citep{Chiu87:RedundantRbtCtrl}, and to improve minimum-time trajectory planning using a manipulability-aware inverse kinematics algorithm~\citep{Chiacchio90:MinimumTimeME}. \citet{Vahrenkamp12:Manipulability} proposed a grasp selection process that favored high manipulability in the robot workspace. Other works have focused on maximizing the manipulability ellipsoid volume in trajectory generation algorithms~\citep{Guilamo06:ManipulabilityOpt}, and task-level robot programming frameworks~\citep{Somani16:TaskRobotProgram}, to obtain singularity-free joint trajectories and high task-space dexterity. Nevertheless, as stated in~\citep{Lee89:DualArmManipulability}, solely maximizing the ellipsoid volume to achieve high dexterity in motion may cause a reverse effect on the flexibility in force.

The aforementioned approaches do not specify a desired robot manipulability for the task. In contrast, \citet{LeeOh16:PostureSelection} proposed an optimization method to find reaching postures for a humanoid robot that achieved desired (manually-specified) manipulability volumes. Similarly, a series of desired manipulability ellipsoids was predefined according to Cartesian velocity and force requirements in dual-arm manipulation tasks~\citep{Lee89:DualArmManipulability}. Note that both~\citep{Lee89:DualArmManipulability} and~\citep{LeeOh16:PostureSelection} predetermined the task-dependent robot manipulability, which required a meticulous and demanding analysis about the motion that the robot needed to perform, which becomes impractical when the robot is required to carry out a large set of different tasks. Furthermore, these approaches overlooked an important characteristic of manipulability ellipsoids, namely, the fact that they lie on the manifold of symmetric positive definite (SPD) matrices. This may influence the optimal robot joint configuration for the task at hand. 

Other geometric descriptors have been proposed in the literature to evaluate the velocity or force performance of robots at a given joint configuration. In contrast to manipulability ellipsoids that do not fully account for boundary limits in the space of joint velocities or torques, manipulability polytopes provide a linear estimate of the exact joint constraints in task space~\citep{Chiacchio97:ForcePolytopes,Lee97}. Moreover,~\citet{Ajoudani15:SFR} introduced the concept of stiffness feasibility region (SFR) to represent the non-polytopic boundary where the realization of a desired Cartesian stiffness matrix is feasible. 
While the polytope approaches provide a more accurate estimate of the velocity or force generation capabilities of the robot compared to manipulability ellipsoids, their calculation is computationally expensive. SFR is a particular Cartesian stiffness descriptor and therefore does not generalize to other robot control settings. Manipulability ellipsoids are easy to compute, while representing an intuitive estimate of the robot ability to perform velocities, accelerations or exert forces along the different task directions.

In this article we introduce the novel idea that manipulability-based posture variation for task compatibility can be addressed from a robot learning from demonstration perspective. Specifically, we cast this problem as a \emph{manipulability transfer} between a teacher and a learner. The former demonstrates how to perform a task with a desired time-varying manipulability profile, while the latter reproduces the task by exploiting its own redundant kinematic structure so that its manipulability ellipsoid matches the demonstration. Unlike classical learning frameworks that encode reference position, velocity and force trajectories, our approach offers the possibility of transferring posture-dependent task requirements such as preferred directions for motion and force exertion in operational space, which are encapsulated in the demonstrated manipulability ellipsoids. 

This idea opens two main challenges, namely, \emph{(i)} how to encode and retrieve a sequence of manipulability ellipsoids, and \emph{(ii)} how to track a desired time-varying manipulability either as the main task of the robot or as a secondary objective. To address the former problem, we propose a tensor-based formulation of Gaussian mixture model (GMM) and Gaussian mixture regression (GMR) that takes into account that manipulability ellipsoids lie on the manifold of symmetric positive definite (SPD) matrices (see Section \ref{sec:Learning} for a full description of the model). The latter challenge is solved through a manipulability tracking formulation inspired by the classical inverse kinematics problem in robotics, where a first-order differential relationship between the robot manipulability ellipsoid and the robot joints is established, as explained in Section \ref{sec:ManTrack}. This relationship also demands to consider that manipulability ellipsoids lie on the SPD manifold, which is here tackled by exploiting tensor-based representations and differential geometry (see Section \ref{sec:Bckgr}). The geometry-awareness of our formulations is crucial for achieving successful manipulability transfer, as shown in Section \ref{sec:GeomImportance}. Note that Riemannian geometry has also been successfully exploited in robot motion optimization~\citep{Ratliff15:GeomMotionOpt} and manipulability analysis of closed chains~\citep{ParkKim98:ManipClosedKin}.
For sake of clarity, different aspects of the proposed learning and tracking approaches are illustrated with simple examples using simulated planar robots throughout the article.

The proposed approach can be straightforwardly applied to different types of kinetostatic and dynamic manipulability measures. This opens the door to manipulability transfer scenarios with various types of robots where different task requirements at kinematic and dynamic levels can be learned and successfully transfered between agents of different embodiments. The functionality of the proposed approach is evaluated in different simulated manipulability tracking tasks involving a 16-DoF robotic hand and two legged robots. The full manipulability transfer is showcased in a bimanual setup where an unplugging task is kinesthetically demonstrated to a 14-DoF dual-arm robot, which then transfers the learned model to a different dual-arm system that reproduces the unplugging task successfully, as described in Section \ref{sec:Experiment}.

Early contributions on our learning and tracking frameworks were presented in~\citep{Rozo17IROS:ManTransfer} and~\citep{Jaquier18}, respectively. In~\citep{Rozo17IROS:ManTransfer}, the learning approach provided a sequence of desired manipulability ellipsoids that a learner robot reproduced using gradient-based nullspace commands. Existing approaches built on the optimization of manipulability-based indices are not suitable as they do not allow the tracking of specific manipulability ellipsoids. In~\citep{Jaquier18}, the tracking framework used manually-specified robot manipulability ellipsoids for the task. As mentioned previously, this may be tedious and cumbersome when the robot needs to carry out different and complex tasks. Therefore, the integration of the proposed learning and tracking approaches solves the aforementioned problems and offers a complete geometry-aware manipulability transfer framework where manipulability ellipsoid profiles are learned from demonstrations and reproduced accurately. This opens the possibility to transfer posture-dependent task requirements between agents of dissimilar kinematic structures. In particular, this framework also permits to transfer other velocity, force or impedance specifications with any priority order with respect to the manipulability tracking controller.

Beyond the combination of our early contributions on manipulability learning and tracking, the other contributions of this article are: 
\emph{(i)} analyzing the role of the proposed differential geometry formulation of the geometry-aware tensor-based GMM/GMR adapted to manipulability ellipsoids; \emph{(ii)} extending the geometry-aware manipulability tracking control scheme initially designed for kinetostatic manipulability measures to dynamic measures; \emph{(iii)} demonstrating the exponential stability of the proposed manipulability tracking controllers; \emph{(iv)} introducing various novel types of geometry-aware manipulability tracking schemes and introducing methodologies to consider the robot actuators contribution and variability-based tracking precision; \emph{(v)} analyzing the importance of the geometry-awareness of the manipulability tracking controllers by comparison against state-of-the-art manipulability-based optimization methods.

A summary video, as well as videos of the illustrative planar examples and simulated and real experiments accompany the article and can be found at \url{https://sites.google.com/view/manipulability}. Related source codes are available at \url{https://github.com/NoemieJaquier/Manipulability}.

\section{Background}
\label{sec:Bckgr}

\subsection{Manipulability ellipsoids}
\label{subsec:ManEll}

Velocity and force manipulability ellipsoids introduced in~\citep{Yoshikawa85:Manipulabilty} are kinetostatic performance measures of robotic platforms. They indicate the preferred directions in which force or velocity control commands may be performed at a given joint configuration. More specifically, the velocity manipulability ellipsoid describes the characteristics of feasible motion in Cartesian space corresponding to all the unit norm joint velocities. The velocity manipulability of an $n$-DoF robot can be found by using the kinematic relationship between task velocities $\bm{\dot{x}}$ and joint velocities $\bm{\dot{q}}$,
\begin{equation}
\bm{\dot{x}} = \bm{J}(\bm{q}) \bm{\dot{q}},
\label{Eq:InvKin}
\end{equation} 
where $\bm{q}\!\in\!\mathbb{R}^n$ and $\bm{J}\!\in\!\mathbb{R}^{6 \times n}$ are the joint position and Jacobian of the robot, respectively. Moreover, consider the set of joint velocities of constant (unit) norm $\|\bm{\dot{q}}\|^2 \!=\! 1$ describing the points on the surface of a hypersphere in the joint velocity space, which is mapped into the Cartesian velocity space $\mathbb{R}^6$ with\footnote{Note that an additional scaling of the joint velocities may be included to consider actuator boundaries.}
\begin{align} 
\| \bm{\dot{q}} \|^2  = \bm{\dot{q}}^\trsp \bm{\dot{q}}  = \bm{\dot{x}}^\trsp(\bm{J}\bm{J}^\trsp)^{-1}\bm{\dot{x}},
\label{Eq:VelocityMapping}
\end{align}
by using the least-squares inverse kinematics relation ${\bm{\dot{q}}\!=\!\bm{J}^\dagger\bm{\dot{x}}\!=\!\bm{J}^\trsp(\bm{J}\bm{J}^\trsp)^{-1}\bm{\dot{x}}}$. Equation \eqref{Eq:VelocityMapping} represents the robot manipulability in terms of motion, indicating the flexibility of the manipulator in generating velocities in Cartesian space.\footnote{Dually, the force manipulability ellipsoid can be computed from the static relationship between joint torques and Cartesian forces~\citep{Yoshikawa85:Manipulabilty}.}

In the literature, the velocity manipulability ellipsoid is usually defined as $(\bm{J}\bm{J}^\trsp)^{-1}$, where the principal axes of the ellipsoid coincide with the eigenvectors and their length is equal to the inverse of the square root of the corresponding eigenvalues, i.e., $\frac{1}{\sqrt{\lambda_i}}$ (see e.g.~\citep{Chiu87:RedundantRbtCtrl}). For the sake of consistency, we here use an alternative definition of the velocity manipulability ellipsoid given by ${\bm{M}^{\bm{\dot{x}}} = \bm{J}\bm{J}^\trsp}$. 
In this case, the major axis of the manipulability ellipsoid is aligned to the eigenvector associated with the maximum eigenvalue $\lambda_{\mathrm{max}}$ of $\bm{M}^{\bm{\dot{x}}}$, whose length equals the square root of $\lambda_{\mathrm{max}}$. Thus, the interpretation and representation of the manipulability ellipsoid from the corresponding matrix are facilitated.
Note that the major axis of the velocity manipulability ellipsoid $\bm{M}^{\bm{\dot{x}}} = \bm{J}\bm{J}^\trsp$ indicates the direction in which the greater velocity can be generated, which is also the direction in which the robot is more sensitive to perturbations. This occurs due to the principal axes of the force manipulability being aligned with those of the velocity manipulability, with reciprocal lengths (eigenvalues) caused by the duality of velocity and force (see~\citep{Chiu87:RedundantRbtCtrl} for details). 

Other forms of manipulability ellipsoids exist, such as the dynamic manipulability~\citep{Yoshikawa85:DinamicMan}, which gives a measure of the ability of performing end-effector accelerations along each task-space direction for a given set of joint torques. This has shown to be useful when the robot dynamics cannot be neglected in highly dynamic manipulation tasks~\citep{Chiacchio91:DynManipulability}. Recent works have extended this measure to analyze the robot capacity to accelerate its center of mass for locomotion stability~\citep{Azad17:CoMmanipulability, Gu15:CoMmanipulability}, showing the applicability of the aforementioned tools beyond robotic manipulation.  

As mentioned previously, any manipulability ellipsoid $\bm{M}$ belongs to the set of symmetric positive definite (SPD) matrices $\mathcal{S}_{\ty{++}}^D$ which describe the interior of the convex cone. Consequently, our manipulability transfer formulation must consider this particular characteristic in order to properly encode, reproduce and track manipulability ellipsoids. To do so, we here propose geometry-aware formulations of both learning and tracking problems by exploiting Riemannian manifolds and tensor representations, which are introduced next.  

\subsection{Riemannian manifold of SPD matrices}
\label{subsec:Riemannian}
The set of $D\!\times\!D$ SPD matrices $\mathcal{S}_{\ty{++}}^D$ is not a vector space since it is not closed under addition and scalar product~\citep{Pennec06}, and thus the use of classical Euclidean space methods for treating and analyzing these matrices is inadequate. A compelling solution is to endow these matrices with a Riemannian metric so that these form a Riemannian manifold.\footnote{The original cone of SPD matrices has been changed into a regular and complete (but curved) manifold with an infinite development in each of its $D(D + 1)/2$ directions~\citep{Pennec06}.} This metric permits to define geodesics, which are the generalization of straight lines to Riemannian manifolds. Similarly to straight lines in Euclidean space, geodesics are the minimum-length curves between two points on the manifold.

Intuitively, a Riemannian manifold $\mathcal{M}$ is a mathematical space for which each point locally resembles a Euclidean space. For each point $\bm{\Sigma}\!\in\!\mathcal{M}$, there exists a tangent space $\mathcal{T}_{\bm{\Sigma}} \mathcal{M}$ equipped with a positive definite inner product. In the case of the SPD manifold, the tangent space at any point ${\bm{\Sigma}\in\mathcal{S}_{\ty{++}}^D}$ is identified by the space of symmetric matrices $\text{Sym}^D$ and the inner product between two matrices $\bm{T}_1$, $\bm{T}_2\in\mathcal{T}_{\bm{\Sigma}} \mathcal{M}$ is 
\begin{equation}
\langle\bm{T}_1,\bm{T}_2\rangle _{\bm{\Sigma}} \;\;=\;\; \tr(\bm{\Sigma}^{-\frac{1}{2}}\bm{T}_1\bm{\Sigma}^{-1}\bm{T}_2\bm{\Sigma}^{-\frac{1}{2}}).
\label{Eq:SPDinnerprod}
\end{equation}

The space of SPD matrices can be represented as the interior of a convex cone embedded in its tangent space $\text{Sym}^D$. To utilize these tangent spaces, we need mappings back and forth between $\mathcal{T}_{\bm{\Sigma}} \mathcal{M}$ and $\mathcal{M}$, which are known as exponential and logarithmic maps. 

\newsavebox{\spdmat}
\savebox{\spdmat}{$\left(\begin{smallmatrix}T_{11} & T_{12}\\ T_{12} & T_{22}\end{smallmatrix}\right)$}
\begin{figure}[tbp]
	\centering
	\begin{subfigure}[b]{0.24\textwidth}
		\includegraphics[width=\textwidth]{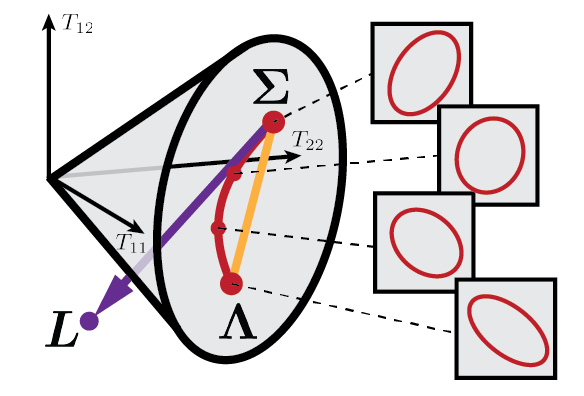}
		\caption{}
		\label{subFig:SPD_geodesic}
	\end{subfigure}
	\begin{subfigure}[b]{0.24\textwidth}
		\includegraphics[width=\textwidth]{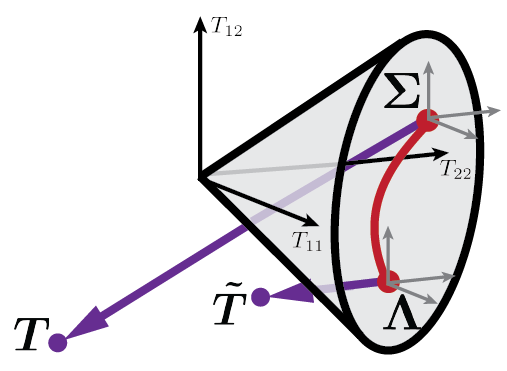}
		\caption{}
		\label{subFig:SPD_prltrsp}
	\end{subfigure}
	\caption{SPD manifold $\mathcal{S}_{\ty{++}}^2$ embedded in its tangent space $\text{Sym}^2$. One point corresponds to a matrix~\usebox{\spdmat} $\in \text{Sym}^2$. Points inside the cone, such as $\bm{\Sigma}$ and $\bm{\Lambda}$, belong to $\mathcal{S}_{\ty{++}}^2$. (\emph{a}) $\bm{L}$ lies on the tangent space of $\bm{\Sigma}$ such that $\bm{L} = \text{Log}_{\bm{\Sigma}}(\bm{\Lambda})$. The shortest path between $\bm{\Sigma}$ and $\bm{\Lambda}$ is the geodesic represented as a red curve, which differs from the Euclidean path depicted in yellow. (\emph{b}) $\bm{\tilde{T}}\in\mathcal{T}_{\bm{\Lambda}}\mathcal{M}$ is the result of the parallel transport of $\bm{T}\in\mathcal{T}_{\bm{\Sigma}}\mathcal{M}$ from $\mathcal{T}_{\bm{\Sigma}} \mathcal{M}$ to $\mathcal{T}_{\bm{\Lambda}} \mathcal{M}$. The coordinate axes of the tangent spaces $\mathcal{T}_{\bm{\Sigma}}\mathcal{M}$ and $\mathcal{T}_{\bm{\Lambda}}\mathcal{M}$ are represented in dark gray.}
	\label{Fig:SPD}
\end{figure}

The exponential map $\text{Exp}_{\bm{\Sigma}}: \mathcal{T}_{\bm{\Sigma}} \mathcal{M}\to \mathcal{M}$ maps a point $\bm{L}$ in the tangent space to a point $\bm{\Lambda}$ on the manifold, so that it lies on the geodesic starting at $\bm{\Sigma}$ in the direction $\bm{L}$ and such that the distance between $\bm{\Sigma}$ and $\bm{\Lambda}$ is equal to the norm of $\bm{L}$ in the tangent space. The inverse operation is called the logarithmic map $\text{Log}_{\bm{\Sigma}}:  \mathcal{M}\to \mathcal{T}_{\bm{\Sigma}}\mathcal{M}$. Both operations are illustrated in Fig.\ \ref{subFig:SPD_geodesic}. 

Specifically, the exponential and logarithmic maps on the SPD manifold corresponding to the affine-invariant distance 
\begin{equation}
d(\bm{\Lambda},\bm{\Sigma}) = \|\log(\bm{\Sigma}^{-\frac{1}{2}}\bm{\Lambda}\bm{\Sigma}^{-\frac{1}{2}})\|_\text{F},
\label{Eq:SPDdist}
\end{equation}
are computed as (see~\citep{Pennec06} for details)
\begin{align}
\bm{\Lambda} & = \text{Exp}_{\bm{\Sigma}}(\bm{L}) = \bm{\Sigma}^{\frac{1}{2}}\exp(\bm{\Sigma}^{-\frac{1}{2}}\bm{L}\bm{\Sigma}^{-\frac{1}{2}})\bm{\Sigma}^{\frac{1}{2}}, \\
\bm{L} & = \text{Log}_{\bm{\Sigma}}(\bm{\Lambda}) = \bm{\Sigma}^{\frac{1}{2}}\log(\bm{\Sigma}^{-\frac{1}{2}}\bm{\Lambda}\bm{\Sigma}^{-\frac{1}{2}})\bm{\Sigma}^{\frac{1}{2}},
\label{Eq:SPDmaps}
\end{align}
where $\exp(\cdot)$ and $\log(\cdot)$ are the matrix exponential and logarithm functions.

Another useful operation over manifolds is the parallel transport $\Gamma_{\bm{\Sigma}\to\bm{\Lambda}}: \mathcal{T}_{\bm{\Sigma}}\mathcal{M}\to\mathcal{T}_{\bm{\Lambda}}\mathcal{M}$, which moves elements between tangent spaces such that the angle between two elements in the tangent space remains constant (see Fig.~\ref{subFig:SPD_prltrsp}). The parallel transport of $\bm{T}\in\mathcal{T}_{\bm{\Sigma}}\mathcal{S}_{\ty{++}}^D$ to $\mathcal{T}_{\bm{\Lambda}}\mathcal{S}_{\ty{++}}^D$ is given by
\begin{equation}
\bm{\tilde{T}} = \Gamma_{\bm{\Sigma}\to\bm{\Lambda}} (\bm{T}) = \bm{A}_{\bm{\Sigma}\to\bm{\Lambda}} \; \bm{T} \; \bm{A}_{\bm{\Sigma}\to\bm{\Lambda}}^\trsp,
\label{Eq:spdPT}
\end{equation}
with $\bm{A}_{\bm{\Sigma}\to\bm{\Lambda}} = \bm{\Lambda}^{\frac{1}{2}}\bm{\Sigma}^{-\frac{1}{2}}$ (see~\citep{SraHosseini15:ConicOpt} for details). This operation is exploited when it is necessary to move SPD matrices along a curve on the nonlinear manifold. Finally, for a complete introduction to differential geometry and Riemannian manifolds, we refer the interested reader to e.g.,~\citep{DoCarmo92:RiemannianGeometry,Lee12:SmoothManifolds}.

In this article, we first exploit the Riemannian manifold framework to propose a probabilistic learning model that encodes and retrieves manipulability ellipsoids considering that these belong to $\mathcal{S}_{\ty{++}}^D$. Secondly, we take advantage of the Riemannian geometry to compute the difference between manipulability ellipsoids in the tracking problem, and consequently propose novel velocity- and acceleration-based controllers. This geometry-aware approach proves to be crucial for learning and tracking manipulability ellipsoids in terms of accuracy, stability and convergence, beyond providing an appropriate mathematical treatment of both problems.

\subsection{Tensor representation}
\label{subsec:TensorOp}
Tensors are generalization of matrices to arrays of higher dimensions~\citep{Kolda09}, where vectors and matrices may respectively be seen as 1st and 2nd-order tensors. Tensor representation permits to represent and exploit data structure of multidimensional arrays. In this article, such representation is first used in the learning process to encode a distribution of manipulability ellipsoids (as explained in Section~\ref{sec:Learning}). Then, tensor representation is also exploited in the proposed manipulability tracking formulation to find the first-order differential relationship between the robot joints and the robot manipulability ellipsoid (1st- and 2nd-order tensors, respectively), which results in a 3rd-order tensor (see Section~\ref{sec:ManTrack}). To do so, we first introduce the tensor operations needed for our mathematical treatment. 

\subsubsection{Tensor product}
The tensor product is a multilinear generalization of the outer product of two vectors $\bm{x} \otimes \bm{y}=\bm{x}\bm{y}^\trsp$. The tensor product of two tensors $\bm{\mathcal{X}}\in \mathbb{R}^{I_1\times\ldots\times I_M}$, $ \bm{\mathcal{Y}}\in \mathbb{R}^{J_1\times\ldots\times J_N}$ is  $\bm{\mathcal{X}}\otimes\bm{\mathcal{Y}} \in \mathbb{R}^{I_1\times\ldots\times I_M\times J_1\times\ldots\times J_N}$ with elements
\begin{equation}
(\bm{\mathcal{X}}\otimes\bm{\mathcal{Y}})_{i_1,\ldots,i_M,j_1,\ldots,j_N} \; = \; x_{i_1,\ldots,i_M} \; y_{j_1,\ldots,j_N}.
\label{Eq:OuterProd}
\end{equation}

\subsubsection{$n$-mode product}
The multiplication of a tensor ${\bm{\mathcal{X}}\!\in\!\mathbb{R}^{I_1 \times \ldots \times I_n \times \ldots \times I_N}}$ by a matrix $\bm{A}\!\in\!\mathbb{R}^{J \times I_n}$, known as the $n$-mode product is defined as
\begin{equation}
\bm{\mathcal{Y}} = \bm{\mathcal{X}} \times _n \bm{A} \iff \bm{Y}_{(n)} =\bm{A}\bm{X}_{(n)} ,
\label{Eq:TMprod}
\end{equation}
where $\bm{X}_{(n)}\!\in\! \mathbb{R}^{I_n \times I_1 I_2\ldots I_N}$ is the $n$-mode matricization or unfolding of tensor $\bm{\mathcal{X}}$. Element-wise, this $n$-mode product can be written as ${(\bm{\mathcal{X}} \times _n \bm{A})_{i_1\ldots i_{n-1} j_n i_{n+1}\ldots i_N} = \sum\limits_{i_n} a_{j_n i_n} x_{i_1\ldots i_{n-1}i_n i_{n+1}\ldots i_N}}$.

\subsubsection{Tensor contraction}
As described in~\citep{Tyagi08}, we denote the element $(i, j, k, l)$ of a 4th-order tensor $\bm{\mathcal{S}}$ by $\bm{\mathcal{S}}_{ij}^{kl}$ with two covariant indices $i$, $j$ and two contravariant indices $k$, $l$. The element ($k$,$l$) of a matrix $\bm{X}$ is denoted by $\bm{X}_{kl}$ with two covariant indices $k$, $l$. A tensor contraction between two tensors is performed when one or more contravariant and covariant indices are identical. For example, the tensor contraction of $\bm{\mathcal{S}}\in\mathbb{R}^{D\times D\times D\times D}$ and $\bm{X}\in\mathbb{R}^{D\times D}$ is written as
\begin{equation}
\bm{\mathcal{S}}\bm{X} = \sum_{k=1}^{D}\sum_{l=1}^{D} \bm{\mathcal{S}}_{ij}^{kl}\bm{X}_{kl}.
\label{Eq:tensorContraction}
\end{equation}

\subsubsection{Tensor covariance}
Similarly to the covariance of vectors, the $2M$th-order covariance tensor $\bm{\mathcal{S}}\in\mathbb{R}^{I_1\times\ldots\times I_M \times I_1\times\ldots\times I_M}$ of centered tensors $\bm{\mathcal{X}}_n\in\mathbb{R}^{I_1\times\ldots\times I_M}$ is given by
\begin{equation}
\bm{\mathcal{S}} = \frac{1}{N-1} \sum\limits_{n=1}^{N} \bm{\mathcal{X}}_n \otimes \bm{\mathcal{X}}_n,
\label{Eq:TensorCov}
\end{equation}
where $N$ is the total number of datapoints. This definition is used in the formulation of tensor-variate normal distributions.

\subsubsection{Normal distribution of symmetric matrices}
The tensor-variate normal distribution of a random 2nd-order symmetric matrix $\bm{X}\in\text{Sym}^D$ with mean $\bm{\Xi}\in\text{Sym}^D$ and covariance $\bm{\mathcal{S}}\in\mathbb{R}^{D\times D\times D\times D}$ is defined as~\citep{Basser07}
\begin{equation}
\mathcal{N}(\bm{X}|\bm{\Xi},\bm{\mathcal{S}}) = \frac{1}{\sqrt{(2\pi)^{\tilde{D}} | \bm{\mathcal{S}} |}} \; e^{-\frac{1}{2} (\bm{X}-\bm{\Xi})\bm{\mathcal{S}}^{-1}(\bm{X}-\bm{\Xi})},
\label{Eq:tensorPDF}
\end{equation} with ${\tilde{D} = D+D(D-1)/2}$.
This formulation is used in Section~\ref{sec:Learning} to formulate a normal distribution of SPD matrices necessary to adapt the formulations of GMM and GMR to encode and retrieve manipulability ellipsoids.

\subsubsection{Derivative of a matrix w.r.t a vector}
In the following identities, the matrix ${\bm{Y}\!\in\!\mathbb{R}^{I\times J}}$ is a function of ${\bm{x}\!\in\!\mathbb{R}^K}$, while $\bm{A}\!\in\!\mathbb{R}^{L\times I}$ and $\bm{B}\!\in\!\mathbb{R}^{J\times L}$ are constant matrices.
The derivative of a matrix function $\bm{Y}$ with respect to a vector $\bm{x}$ is a 3rd-order tensor $\frac{\partial \bm{Y}}{\partial \bm{x}}\!\in\!\mathbb{R}^{I\times J\times K}$ such that
\begin{equation}
\left(\frac{\partial \bm{Y}}{\partial \bm{x}}\right)_{ijk} = \frac{\partial y_{ij}}{\partial x_k}.
\label{Eq:MatVecDeriv}
\end{equation} 

Note that when the matrix function $\bm{Y}$ is multiplied by a constant matrix, the partial derivatives of $\bm{Y}$ are given by:

\paragraph{Left multiplication by a constant matrix}
\begin{equation}
\frac{\partial \bm{AY}}{\partial \bm{x}} = \frac{\partial \bm{Y}}{\partial \bm{x}} \times_1 \bm{A}
\label{Eq:LeftMultDeriv}
\end{equation} 
\paragraph{Right multiplication by a constant matrix}
\begin{equation}
\frac{\partial \bm{YB}}{\partial \bm{x}} = \frac{\partial \bm{Y}}{\partial \bm{x}} \times_2 \bm{B}^\trsp
\label{Eq:RightMultDeriv}
\end{equation} 

Finally, another useful operation for our manipulability tracking formulation is the derivative of the inverse of the matrix $\bm{Y}$ with respect to the vector $\bm{x}$, which results in a 3rd-order tensor, namely
\begin{equation}
\frac{\partial \bm{Y}^{-1}}{\partial \bm{x}} = - \frac{\partial \bm{Y}}{\partial \bm{x}}^\trsp \times_1 \bm{Y}^{-1} \times_2 \bm{Y}^{-\trsp}
\label{Eq:InvDeriv}
\end{equation}

Note that the proposed geometry-aware manipulability tracking, introduced in the section~\ref{sec:ManTrack}, takes inspiration from the computation of the robot Jacobian, which is computed from the 1st-order time derivative of the robot forward kinematics. We use the tensor representation to similarly compute the 1st-order derivative of the function that describes the relationship between a manipulability ellipsoid $\bm{M}$ and the robot joint configuration $\bm{q}$. Mathematical proofs for \eqref{Eq:LeftMultDeriv}, \eqref{Eq:RightMultDeriv} and \eqref{Eq:InvDeriv} are given in Appendix~\ref{sec:AppendixProofsDerivatives}.


\section{Learning Manipulability Ellipsoids}
\label{sec:Learning}
%
%
%
%

The first open problem in manipulability transfer is to appropriately encode sequences of demonstrated manipulability ellipsoids and subsequently retrieve a desired manipulability profile that encapsulates the patterns observed during the demonstrations. In order to describe how we tackle this problem, we first introduce the mathematical formulation of a Gaussian mixture model that encodes a set of demonstrated manipulability ellipsoids over the manifold of SPD matrices. This probabilistic formulation models the trend of the demonstrated manipulability sequences along with their variability, reflecting their dispersion through the different demonstrations. After, we describe how a distribution of the desired manipulability ellipsoids can be retrieved via Gaussian mixture regression on the SPD manifold.

\subsection{Gaussian Mixture Model on SPD manifolds}
\label{subsec:GMM}
Similarly to multivariate distribution (see~\citep{Zeestraten17,SimoSerra16,Dubbelman11}), we can extend the normal distribution~\eqref{Eq:tensorPDF} to the SPD manifold. Thus, a tensor-variate distribution maximizing the entropy in the tangent space is approximated by
\begin{equation}
\mathcal{N}_{\mathcal{M}}(\bm{X}|\bm{\Xi},\bm{\mathcal{S}}) = \frac{1}{\sqrt{(2\pi)^{\tilde{D}} | \bm{\mathcal{S}} |}} \; e^{-\frac{1}{2} \text{Log}_{\bm{\Xi}}\!(\bm{X}) \, \bm{\mathcal{S}}^{-1} \, \text{Log}_{\bm{\Xi}}\!(\bm{X})},
\label{Eq:tensorPDFmanifold}
\end{equation}
where $\bm{X}\in\mathcal{M}$, $\bm{\Xi}\in\mathcal{M}$ is the origin in the tangent space and $\bm{\mathcal{S}}\in\mathcal{T}_{\bm{\Xi}}\mathcal{M}$ is the covariance tensor.

Similarly to the Euclidean case, a GMM on the SPD manifold is defined by 
\begin{equation}
p(\bm{X}) = \sum_{k=1}^{K} \pi_k \mathcal{N}_{\mathcal{M}}(\bm{X}|\bm{\Xi}_k,\bm{\mathcal{S}}_k),
\label{Eq:SPD_GMM}
\end{equation}
with $K$ being the number of components of the model, and $\pi_k$ representing the priors such that $\sum_k \pi_k =1$. 

The parameters of a GMM on the manifold of SPD matrices are estimated by Expectation-Maximization (EM) algorithm. Specifically, the responsibility of each component $k$ is computed in the E-step as:
\begin{align}
p(k|\bm{X}_i) &= \frac{\pi_k\;\mathcal{N}_{\mathcal{M}}(\bm{X}_i|\bm{\Xi}_k,\bm{\mathcal{S}}_k)}{\sum_{j=1}^{K}\pi_j\;\mathcal{N}_{\mathcal{M}}(\bm{X}_i|\bm{\Xi}_j,\bm{\mathcal{S}}_j)}, \\
N_k &= \sum_{i=1}^{N} p(k|\bm{X}_i).
\label{Eq:SPD_Estep}
\end{align}
During the M-step, the mean $\bm{\Xi}_k$ is first updated iteratively until convergence for each component. The covariance tensor $\bm{\mathcal{S}}_k$ and prior $\pi_k$ are then updated using the new mean:
\begin{align}
\bm{\Xi}_k &\gets \frac{1}{N_k}\text{Exp}_{\bm{\Xi}_k}\!\left(\sum_{i=1}^{N}p(k|\bm{X}_i)\;\text{Log}_{\bm{\Xi}_k}\!(\bm{X}_i)\right), \\
\bm{\mathcal{S}}_k &\gets \frac{1}{N_k}\sum_{i=1}^{N}p(k|\bm{X}_i)\;\text{Log}_{\bm{\Xi}_k}\!(\bm{X}_i)\otimes \text{Log}_{\bm{\Xi}_k}\!(\bm{X}_i), \\
\pi_k &\gets \frac{N_k}{N}.
\end{align}

\subsection{Gaussian Mixture Regression on SPD manifolds}
\label{subsec:GMR}
GMR computes the conditional distribution $p(\bm{X}_\ty{OO}|\bm{X}_\ty{II})$ of the joint distribution $p(\bm{X})$, where the sub-indices $\mathcal{I}$ and $\mathcal{O}$ denote the sets of dimensions that span the input and output variables. We use the following block decomposition of the datapoints, means and covariances:
\begin{align}
&\bm{X} = \left(\begin{matrix}
\bm{X}_\ty{II} & \bm{0}\\
\bm{0} & \bm{X}_\ty{OO}\\
\end{matrix}\right),
\bm{\Xi} = \left(\begin{matrix}
\bm{\Xi}_{\ty{II}} & \bm{0}\\
\bm{0} & \bm{\Xi}_{\ty{OO}}\\
\end{matrix}\right),
\nonumber\\
&\bm{\mathcal{S}} = 
\left(\begin{array}{cc|cc}
\bm{\mathcal{S}}_{\ty{II}}^\ty{II} & \bm{0} & \bm{0} & \bm{0}\\
\bm{0} & \bm{\mathcal{S}}_{\ty{II}}^\ty{OO} & \bm{0} & \bm{0}\\
\hline 
\bm{0} & \bm{0} & \bm{\mathcal{S}}_{\ty{OO}}^\ty{II} & \bm{0}\\
\bm{0} & \bm{0} & \bm{0} & \bm{\mathcal{S}}_{\ty{OO}}^\ty{OO}
\end{array}\right),
\label{Eq:blockDec}
\end{align} where we represent the 4th-order tensor by separating the components of the 3rd- and 4th-modes with horizontal and vertical bars, respectively.
With this decomposition, manifold functions can be applied individually on input and output parts, for example the exponential map would be  
\begin{equation*}
\text{Exp}_{\bm{\Xi}_k}\!(\bm{X}) = \left(\begin{matrix}
\text{Exp}_{\bm{\Xi}_\ty{II}}\!(\bm{X}_\ty{II}) & \bm{0}\\
\bm{0} & \text{Exp}_{\bm{\Xi}_\ty{OO}}\!(\bm{X}_\ty{OO})\\
\end{matrix}\right).
\end{equation*} 

Similarly to GMR in Euclidean space~\citep{RozoEtAl16:Learning4HRC} and in manifolds where data are represented by vectors~\citep{Zeestraten17}, GMR on SPD manifold approximates the conditional distribution by a single Gaussian 
\begin{equation}
p(\bm{X}_\ty{OO}|\bm{X}_\ty{II}) \,\sim\, \mathcal{N}(\bm{\hat{\Xi}}_\ty{OO},\bm{\hat{\mathcal{S}}}_\ty{OO}^\ty{OO}),
\label{Eq:condGMR}
\end{equation}where the mean $\bm{\hat{\Xi}}_\ty{OO}$ is computed iteratively until convergence in its tangent space using
\begin{equation}
\bm{\Delta}_k = \text{Log}_{\bm{\hat{\Xi}}_{\ty{OO}}}\!(\bm{\Xi}_{\ty{OO},k})-\tilde{\bm{\mathcal{S}}}_{\ty{OO},k}^\ty{II}\;
\tilde{\bm{\mathcal{S}}}_{\ty{II},k}^{\ty{II}-1} \; \text{Log}_{\bm{X}_{\ty{II}}}\!(\bm{\Xi}_{\ty{II},k}),
\label{Eq:SPD_GMRmeanParam}
\end{equation}
\begin{equation}
\bm{\hat{\Xi}}_\ty{OO} \;\gets\; \text{Exp}_{\bm{\hat{\Xi}}_\ty{OO}}\!\Big(\sum\limits_{k}h_k\bm{\Delta}_k\Big),
\label{Eq:SPD_GMRmean}
\end{equation}
with $h_k$ describing the responsibilities of the GMM components in the regression, namely
\begin{equation}
h_k = \frac{\pi_k\;\mathcal{N}(\bm{X}_\ty{II}|\bm{\Xi}_{\ty{II},k},\bm{\mathcal{S}}_{\ty{II},k}^\ty{II})}{\sum_{j=1}^{K}\pi_j\;\mathcal{N}(\bm{X}_\ty{II}|\bm{\Xi}_{\ty{II},j},\bm{\mathcal{S}}_{\ty{II},j}^\ty{II})}.
\label{Eq:SPD_h}
\end{equation}
The covariance $\bm{\hat{\mathcal{S}}}_\ty{OO}^\ty{OO}$ is then computed in the tangent space of the mean
\begin{align}
\bm{\hat{\mathcal{S}}}_\ty{OO}^\ty{OO} =& \sum_{k} h_k \left(\tilde{\bm{\mathcal{S}}}_{\ty{OO},k}^\ty{OO}-\tilde{\bm{\mathcal{S}}}_{\ty{OO},k}^\ty{II}\tilde{\bm{\mathcal{S}}}_{\ty{II},k}^{\ty{II}-1}\tilde{\bm{\mathcal{S}}}_{\ty{II},k}^\ty{OO}+ \bm{\Delta}_k \otimes \bm{\Delta}_k\right) \nonumber\\
&- \bm{\hat{\Xi}}_\ty{OO} \otimes \bm{\hat{\Xi}}_\ty{OO}
\label{Eq:SPD_GMRcov},
\end{align}
where $\tilde{\bm{\mathcal{S}}}$ is the parallel transported covariance tensor
\begin{equation}
\tilde{\bm{\mathcal{S}}} = \Gamma_{\bm{\Xi}\to \bm{\hat{X}}} (\bm{\mathcal{S}}) \quad\text{with}\quad \bm{\hat{X}}=\left(\begin{matrix}
\bm{X}_\ty{II} & \bm{0}\\
\bm{0} & \bm{\hat{\Xi}}_\ty{OO}\\
\end{matrix}\right).
\label{Eq:SPD_GMRpt}
\end{equation}
This covariance has been typically used to define the controller gains of robotic systems for trajectory tracking problems (see also Section~\ref{subsec:4orderCov}).
Note that the definition of the tangent space $\mathcal{T}_{\bm{\Xi}} \mathcal{M}$ (which has the structure of a Euclidean vector space) is what allow us to compute the conditional distribution above. Also notice that to parallel transport a 4th-order covariance tensor ${\bm{\mathcal{S}}\in\mathbb{R}^{D\times D\times D\times D}}$, the covariance is first converted to a 2nd-order tensor $\bm{\Sigma}\in\mathbb{R}^{\tilde{D}\times \tilde{D}}$ with $\tilde{D} = D+D(D-1)/2$, as proposed in~\citep{Basser07}. We can then compute its eigentensors $\bm{V}_k$, which are used to parallel transport the covariance matrix between tangent spaces~\citep{Freifeld14}. Let ${\tilde{\bm{V}}_k = \Gamma_{\bm{\Xi}\to \bm{\hat{X}}}(\bm{V}_k)}$ be the $k$-th parallel transported eigentensor with~\eqref{Eq:spdPT} and $\lambda_k$ the $k$-th eigenvalue. The parallel transported 4th-order covariance tensor is then obtained with (see~\citep{Jaquier17} for more details)
\begin{equation}
\Gamma_{\bm{\Xi}\to\bm{\hat{X}}}(\bm{\mathcal{S}}) = \sum_{k} \lambda_k \tilde{\bm{V}}_k \otimes \tilde{\bm{V}}_k.
\label{Eq:PT4thcov}
\end{equation}

\subsection{Manipulability Learning Example with 2 Planar Robots}
In order to illustrate the functionality of the proposed learning approach, we carried out an experiment using a couple of simulated planar robots with dissimilar embodiments and a different number of joints. The central idea is to teach a redundant robot to track a reference trajectory in Cartesian space with a desired time-varying manipulability ellipsoid.
For the demonstration phase, a 3-DoF \textit{teacher} robot follows a C-shape trajectory four times, from which we extracted both the end-effector position $\bm{x}_t$ and robot manipulability ellipsoid $\bm{M}_t(\bm{q})$, at each time step $t$. The collected time-aligned data were split into two training datasets of time-driven trajectories, namely Cartesian position and manipulability. We trained a classical GMM over the time-driven Cartesian trajectories and a geometry-aware GMM over the time-driven manipulability ellipsoids, using models with five components, i.e.\ $K\!=\!5$ (the number was selected by the experimenter).

During the reproduction phase, a 5-DoF \textit{student} robot executed the time-driven task by following a desired Cartesian trajectory $\bm{\hat{x}}_t$ computed from a classical GMR as ${\bm{\hat{x}}_t \sim \mathcal{P}(\bm{x} \! \mid \!t)}$. As secondary task, the robot was also required to vary its joint configuration for matching desired manipulability ellipsoids $\bm{\hat{M}}_t \sim \mathcal{P}(\bm{M}|t)$, estimated by GMR over the SPD manifold. 

\begin{figure}[tbp]
	\centering
	\begin{subfigure}[b]{0.24\textwidth}
		\includegraphics[width=\textwidth]{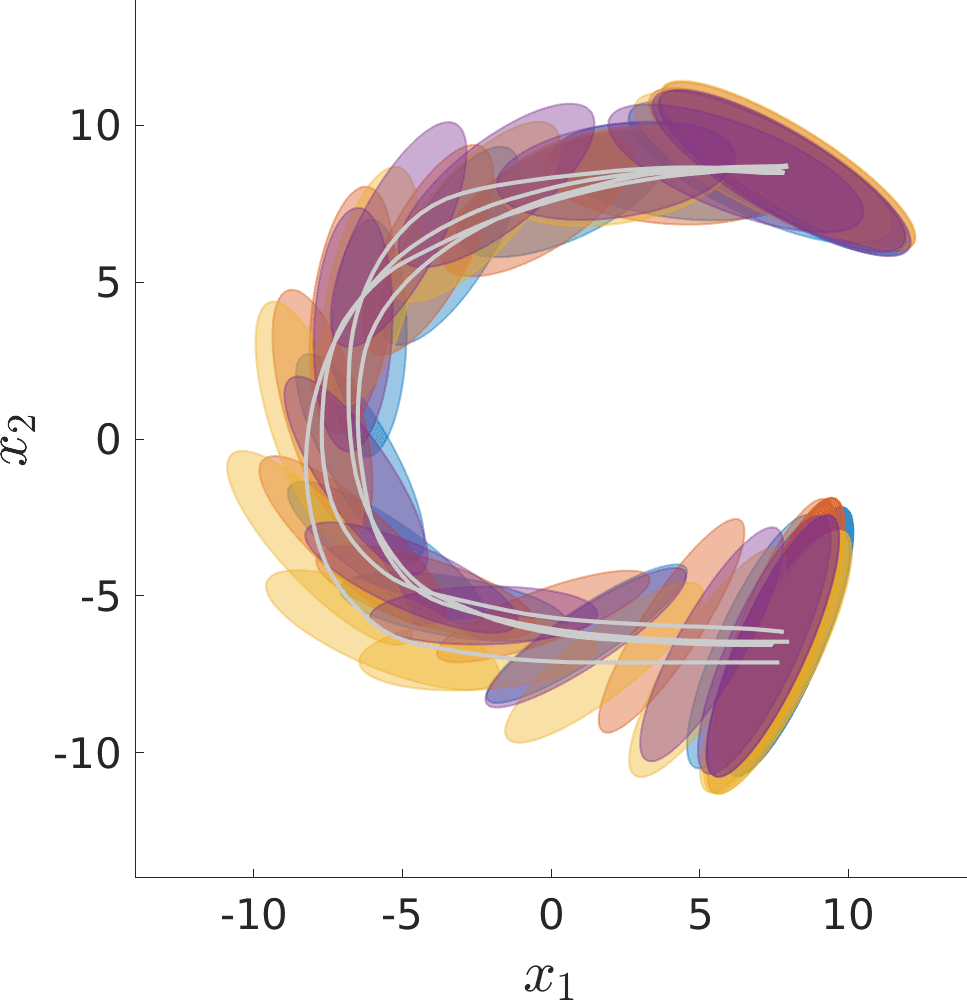}
		\caption{Demonstrations}
		\label{subFig:ManipTranferEx_Data}
	\end{subfigure}
	\begin{subfigure}[b]{0.24\textwidth}
		\includegraphics[width=\textwidth]{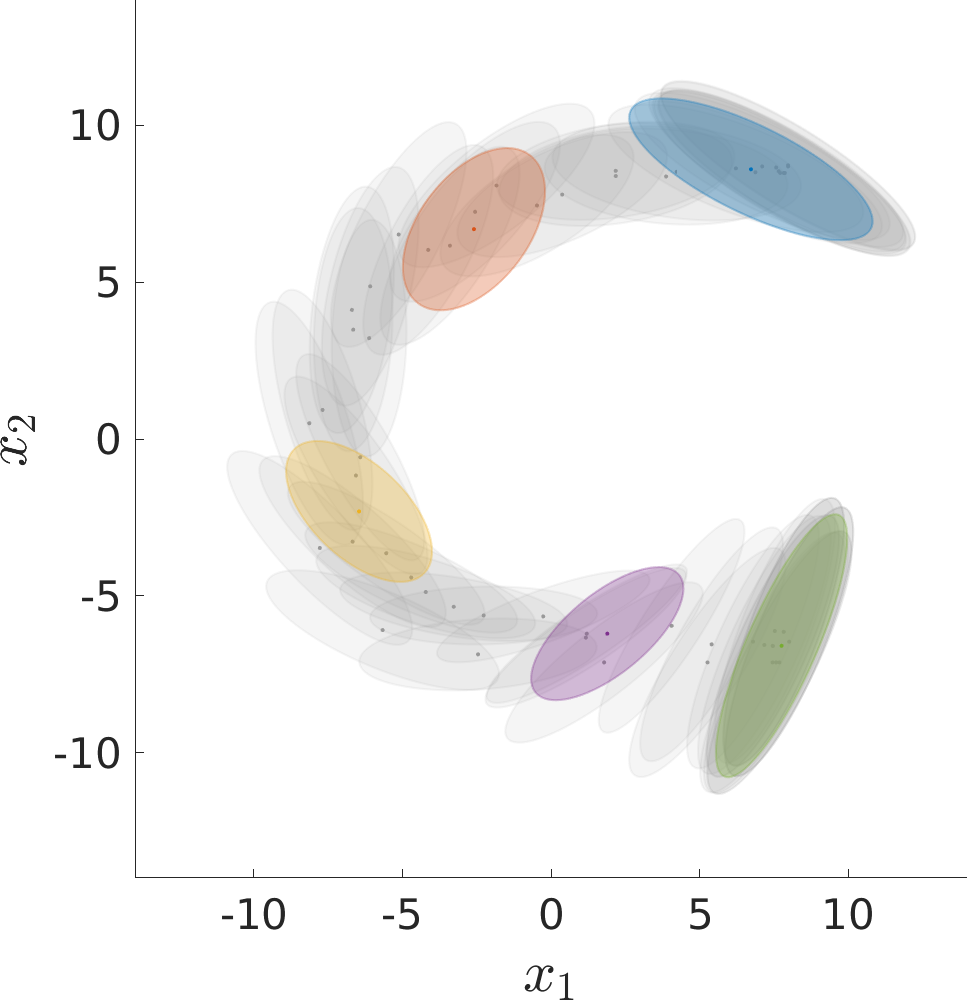}
		\caption{Learned model}
		\label{subFig:ManipTranferEx_GMM}
	\end{subfigure}
	\caption{(\emph{a}) Four demonstrations of a 3-DoF planar robot tracking a C-shape trajectory. The end-effector path (light gray solid lines) and the manipulability ellipsoids at different time steps are shown for all the demonstrations. (\emph{b}) Demonstrated manipulability ellipsoids (in gray) and centers $\bm{\Xi}_k$ of the 5-states GMM in the SPD manifold. Position $\bm{x}$ and time $t$ are given in centimeters and seconds, respectively.}
	\label{Fig:ManipTransferGMM}
\end{figure}

\begin{figure}[tbp]
	\centering
		\begin{subfigure}[b]{0.23\textwidth}
			\includegraphics[width=\textwidth]{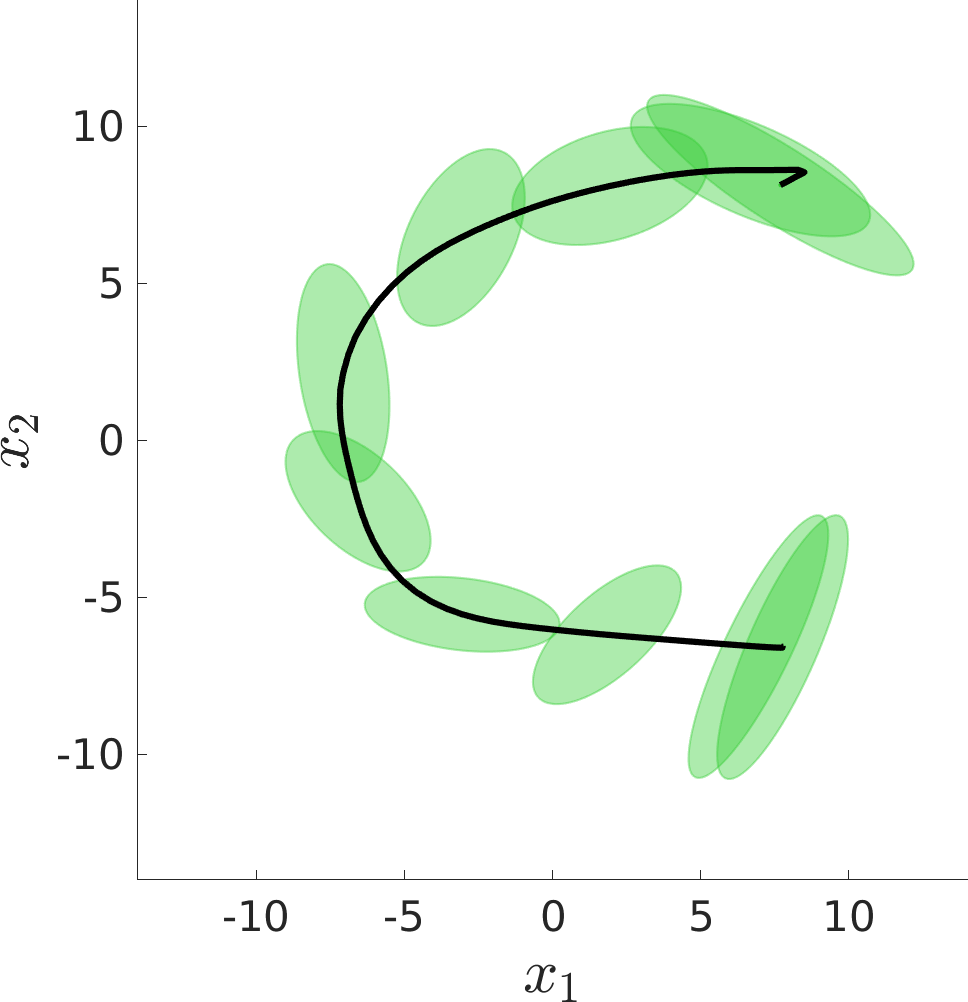}
			\caption{Desired execution}
			\label{subFig:ManipTranferEx_Desired}
		\end{subfigure}
		\begin{subfigure}[b]{0.23\textwidth}
			\includegraphics[width=\textwidth]{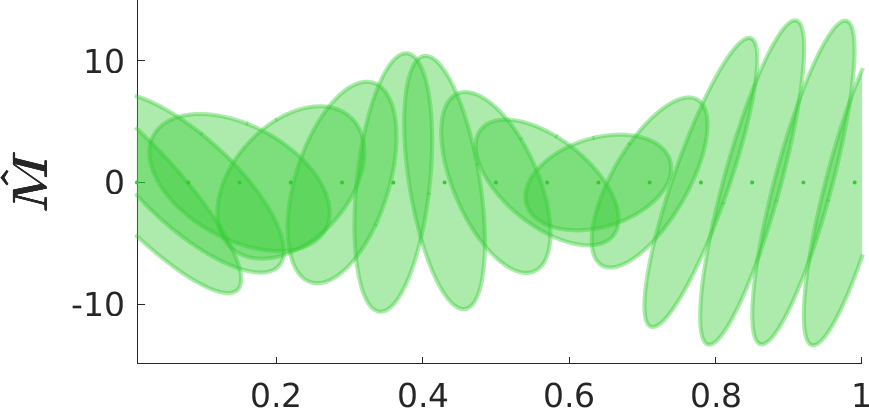}
			\includegraphics[width=\textwidth]{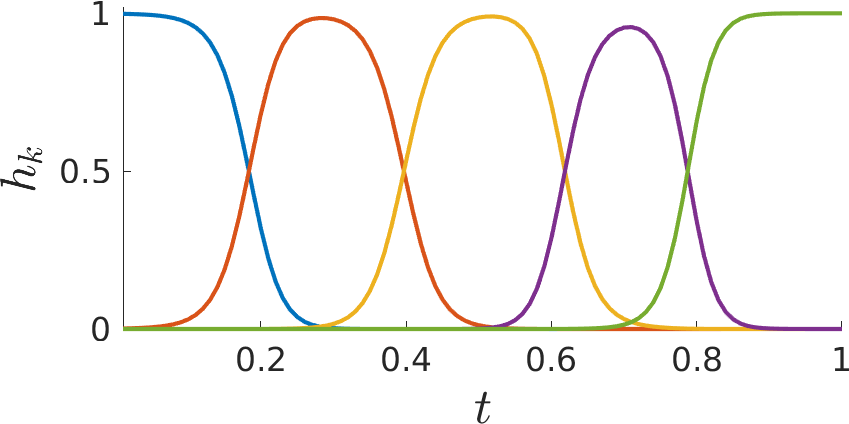}
			\caption{Time profile}
			\label{subFig:ManipTranferEx_GMR_states}
		\end{subfigure}	
	\caption{(\emph{a}) Desired execution of a C-shape tracking task. The desired Cartesian trajectory and manipulability profile are depicted as a black curve and green ellipsoids. (\emph{b})-\emph{top} Desired manipulability ellipsoids estimated by GMR. (\emph{b})-\emph{bottom} Influence of GMM components on the time-driven GMR. The colors match the distributions shown in Fig.~\ref{subFig:ManipTranferEx_GMM}.}
	
	\label{Fig:ManipTransferGMR}
\end{figure}

Figure \ref{subFig:ManipTranferEx_Data} shows the four demonstrations carried out by the 3-DoF robot, where both the Cartesian trajectory and manipulability ellipsoids are displayed. Note that the recorded manipulability ellipsoids slightly change across demonstrations as a side effect of the variation observed in both the initial end-effector position and the generated trajectory. Figure \ref{subFig:ManipTranferEx_GMM} displays the demonstrated ellipsoids (in gray) along with the center $\bm{\Xi}_k$ of the five components of the GMM encoding $\bm{X}^{\bm{M}}$. These are centered at the end-effector position recovered by the classical GMR for the corresponding time steps represented in the geometry-aware GMM. Figure~\ref{Fig:ManipTransferGMR} shows the desired Cartesian trajectory and manipulability ellipsoid profile respectively estimated by classical GMR and GMR in the SPD manifold. Both manipulability and Cartesian path are references to be tracked by the student robot.

These results validate that the proposed learning framework permits to learn and plan the reproduction of reference trajectories, while fulfilling additional task requirements encapsulated in a profile of desired manipulability ellipsoids. In Section~\ref{sec:ManTrack}, we develop a manipulability tracking formulation that will then be used by the 5-DoF student robot to track the desired manipulability profile obtained in the learning phase.

\section{Tracking Manipulability Ellipsoids}
\label{sec:ManTrack}
Several robotic manipulation tasks may demand the robot to track a desired trajectory with certain velocity specifications, or apply forces along different task-related axes. These requirements are more easily achieved if the robot adopts a posture that suits velocity or force control commands. In other tasks, the robot may be required to adopt a posture that comply several aligned velocity or force requirements. These problems can be viewed as matching a set of desired manipulability ellipsoids that are compatible with the task requirements.
In this section, we introduce an approach that addresses this problem by exploiting the mathematical concepts presented in Section \ref{sec:Bckgr}. 

\subsection{Manipulability Jacobian}
\label{subsec:ManJac}
Given a desired profile of manipulability ellipsoids, the goal of the robot is to adapt its posture to match the desired manipulability, either as its main task or as a secondary objective. We here propose a formulation inspired by the classical inverse kinematics problem in robotics, which permits to compute the joint angle commands to track a desired manipulability ellipsoid.

First, the manipulability ellipsoid is expressed as a function of time
\begin{equation}
\bm{M}(t) = f\Big(\bm{J}\big(\bm{q}(t)\big)\Big) ,
\label{Eq:ForwardManip}
\end{equation}for which we can compute the first-order time derivative by applying the chain rule as 
\begin{equation}
\frac{\partial \bm{M}(t)}{\partial t} = \frac{\partial f(\bm{J}(\bm{q}))}{\partial \bm{q}} \times_3 \frac{\partial \bm{q}(t)}{\partial t}^\trsp= \bm{\mathcal{J}}(\bm{q}) \times_3 \bm{\dot{q}}^\trsp,
\label{Eq:ManipJacobian}
\end{equation}
where $\bm{\mathcal{J}}\in\mathbb{R}^{6\times 6 \times n}$ is the \emph{manipulability Jacobian} of an $n$-DoF robot, representing the linear sensitivity of the changes in the robot manipulability ellipsoid $\bm{\dot{M}} = \frac{\partial \bm{M}(t)}{\partial t}$  to the joint velocity $\bm{\dot{q}} = \frac{\partial \bm{q}(t)}{\partial t}$. Note that the computation of the manipulability Jacobian depends on the type of manipulability ellipsoid that is used. We develop here the expressions for the force, velocity and dynamic manipulability ellipsoids.

The derivation of the manipulability Jacobian $\bm{\mathcal{J}}^{\bm{\dot{x}}}$ corresponding to the velocity manipulability ellipsoid $\bm{M}^{\bm{\dot{x}}} = \bm{J}\bm{J}^\trsp$ is straightforward by using \eqref{Eq:LeftMultDeriv} and \eqref{Eq:RightMultDeriv} \footnote{In the remainder of the article we drop dependencies on $\bm{q}$ to simplify the notation.}
\begin{equation}
\bm{\mathcal{J}}^{\bm{\dot{x}}} = \frac{\partial\bm{J}}{\partial \bm{q}}\times_2\bm{J} + \frac{\partial\bm{J}^\trsp}{\partial \bm{q}}\times_1\bm{J}.
\label{Eq:VelocityManipJacobian}
\end{equation}

Similarly, the manipulability Jacobian $\bm{\mathcal{J}}^{\bm{F}}$ corresponding to the force manipulability ellipsoid $\bm{M}^{\bm{F}} = (\bm{J}\bm{J}^\trsp)^{-1}$ is obtained using \eqref{Eq:LeftMultDeriv}, \eqref{Eq:RightMultDeriv} and \eqref{Eq:InvDeriv}, 
\begin{equation}
\bm{\mathcal{J}}^{\bm{F}} = -\left(\frac{\partial\bm{J}}{\partial \bm{q}}\times_2\bm{J} + \frac{\partial\bm{J}^\trsp}{\partial \bm{q}}\times_1\bm{J}\right)\times_1\bm{M}^{\bm{\dot{x}}}\times_2\bm{M}^{\bm{\dot{x}}}.
\label{Eq:ForceManipJacobian}
\end{equation}

In a similar fashion, the manipulability Jacobian $\bm{\mathcal{J}}^{\ddot{\bm{x}}}$ corresponding to the dynamic manipulability ellipsoid $\bm{M}^{\ddot{\bm{x}}} = \bm{\Upsilon}\bm{\Upsilon}^{\trsp}$ with $\bm{\Upsilon} = \bm{J}\bm{\Lambda}(\bm{q})^{-1}$ (as defined in~\citep{Yoshikawa85:DinamicMan}, where $\bm{\Lambda}(\bm{q})$ is the robot inertia matrix), is computed as follows 
\begin{equation}
\bm{\mathcal{J}}^{\ddot{\bm{x}}} = \frac{\partial\bm{\Upsilon}}{\partial \bm{q}}\times_2 \bm{\Upsilon} + \frac{\partial\bm{\Upsilon}^\trsp}{\partial \bm{q}}\times_1 \bm{\Upsilon} ,
\label{Eq:DynManipJacobian}
\end{equation}	
where
\begin{align*}
\frac{\partial\bm{\Upsilon}}{\partial \bm{q}} &=
\frac{\partial\bm{J}}{\partial \bm{q}} \times_2 \bm{\Lambda}^{-\trsp}
+ \frac{\partial\bm{\Lambda}^{-1}}{\partial \bm{q}} \times_1 \bm{J} \\
&= \frac{\partial\bm{J}}{\partial \bm{q}} \times_2 \bm{\Lambda}^{-\trsp}
- \frac{\partial\bm{\Lambda}}{\partial \bm{q}} \times_1 \bm{\Upsilon} \times_2 \bm{\Lambda}^{-\trsp}.
\end{align*} 

Details on the computation of the derivative of the Jacobian and inertia matrix w.r.t the joint angles are given in Appendices~\ref{sec:AppendixManipJacobian} and~\ref{sec:AppendixDynManipJacobian}.

\subsection{Geometry-aware manipulability tracking formulation}
\label{subsec:ManTrackForm}
\subsubsection{Velocity-based controller}
A solution to control a robot so that it tracks a desired end-effector trajectory is to compute the desired joint velocities using the inverse kinematics formulation derived from \eqref{Eq:InvKin}. We use here a similar approach to compute the joint velocities $\bm{\dot{q}}$ to track a desired manipulability profile. More specifically, by minimizing the $\ell^2$ norm of the residuals 
\begin{equation*}
\min_{\bm{\dot{q}}} \|\bm{\dot{M}} - \bm{\mathcal{J}} \times_3 \bm{\dot{q}}^\trsp\| = \min_{\bm{\dot{q}}} \|\text{vec}(\bm{\dot{M}}) - \bm{\mathcal{J}}_{(3)}^\trsp \bm{\dot{q}}\| ,
\end{equation*} 
we can compute the required joint velocities of the robot to track a profile of desired manipulability ellipsoids as its main task with
\begin{equation}
\bm{\dot{q}} = (\bm{\mathcal{J}}_{(3)}^\dagger)^\trsp \text{vec}(\bm{\dot{M}}) ,
\label{Eq:InvManip}
\end{equation} 
where $\text{vec}(\bm{\dot{M}})$ is the vectorization of the matrix $\bm{\dot{M}}$.

Note that \eqref{Eq:InvManip} allows us to define a controller to track a reference manipulability ellipsoid as main task, similarly as the classical velocity-based control that tracks a desired task-space velocity. To do so, we propose to use a geometry-aware similarity measure to compute the joint velocities necessary to move the robot towards a posture where the match between the current manipulability ellipsoid $\bm{M}_t$ and the desired one $\bm{\hat{M}}_t$ is maximum. Specifically, the difference between manipulability ellipsoids is computed using the logarithmic map \eqref{Eq:SPDmaps} on the SPD manifold. Therefore, the corresponding controller is given by
\begin{equation}
\bm{\dot{q}}_t = (\bm{\mathcal{J}}_{(3)}^\dagger)^\trsp \, \bm{K}_{\bm{M}} \, \text{vec}\Big(\text{Log}_{\bm{M}_t}(\bm{\hat{M}}_t)\Big) ,
\label{Eq:ManipTrackFirstTask}
\end{equation} 
where $\bm{K}_{\bm{M}}$ is a gain matrix.

Alternatively, for the case in which the main task of the robot is to track reference trajectories in the form of Cartesian positions or force profiles, the tracking of a profile of manipulability ellipsoids is assigned a secondary role. Thus, the robot task objectives are to track the reference trajectories while exploiting the kinematic redundancy to minimize the difference between current and desired manipulability ellipsoids. In this situation, a manipulability-based redundancy resolution is carried out by computing a nullspace velocity that similarly exploits the geometry of the SPD manifold. Thus, the corresponding controller is given by
\begin{multline}
\bm{\dot{q}}_t = \bm{J}^\dagger \, \bm{K}_{\bm{x}} \, (\bm{\hat{x}}_t-\bm{x}_t) \\
+ (\bm{I}-\bm{J}^\dagger\bm{J}) \, (\bm{\mathcal{J}}_{(3)}^\dagger)^\trsp \, \bm{K}_{\bm{M}} \, \text{vec}\Big(\text{Log}_{\bm{M}_t}(\bm{\hat{M}}_t)\Big) .
\label{Eq:ManipTrackSecTask}
\end{multline} 

\begin{figure}[tpb]
	\centering
	\begin{subfigure}[b]{0.24\textwidth}
		\centering
		\includegraphics[width=0.45\textwidth]{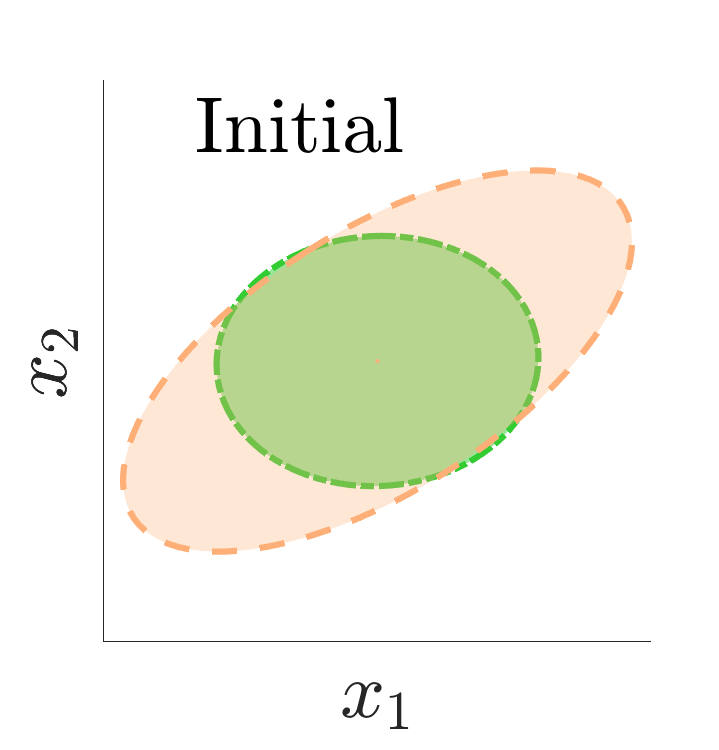}
		\includegraphics[width=0.45\textwidth]{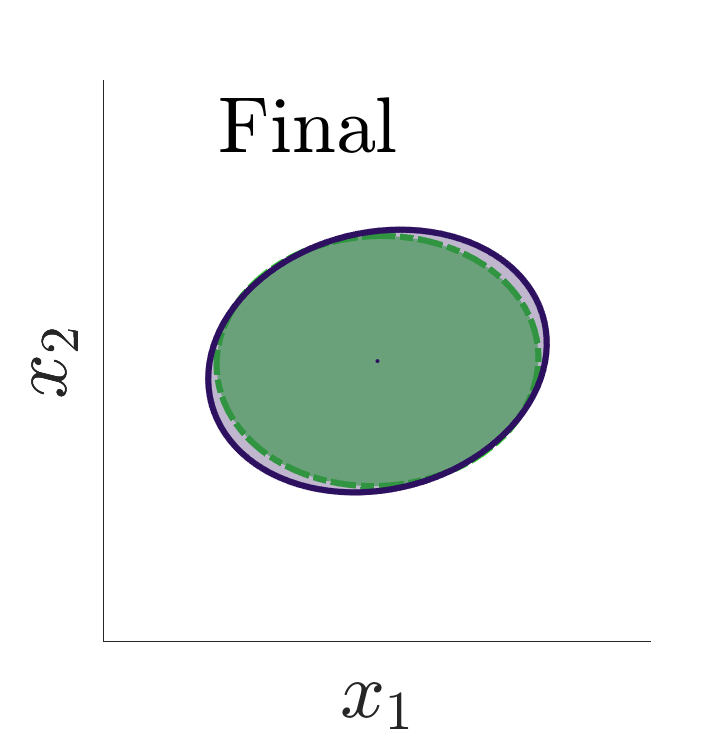}
		\includegraphics[width=\textwidth]{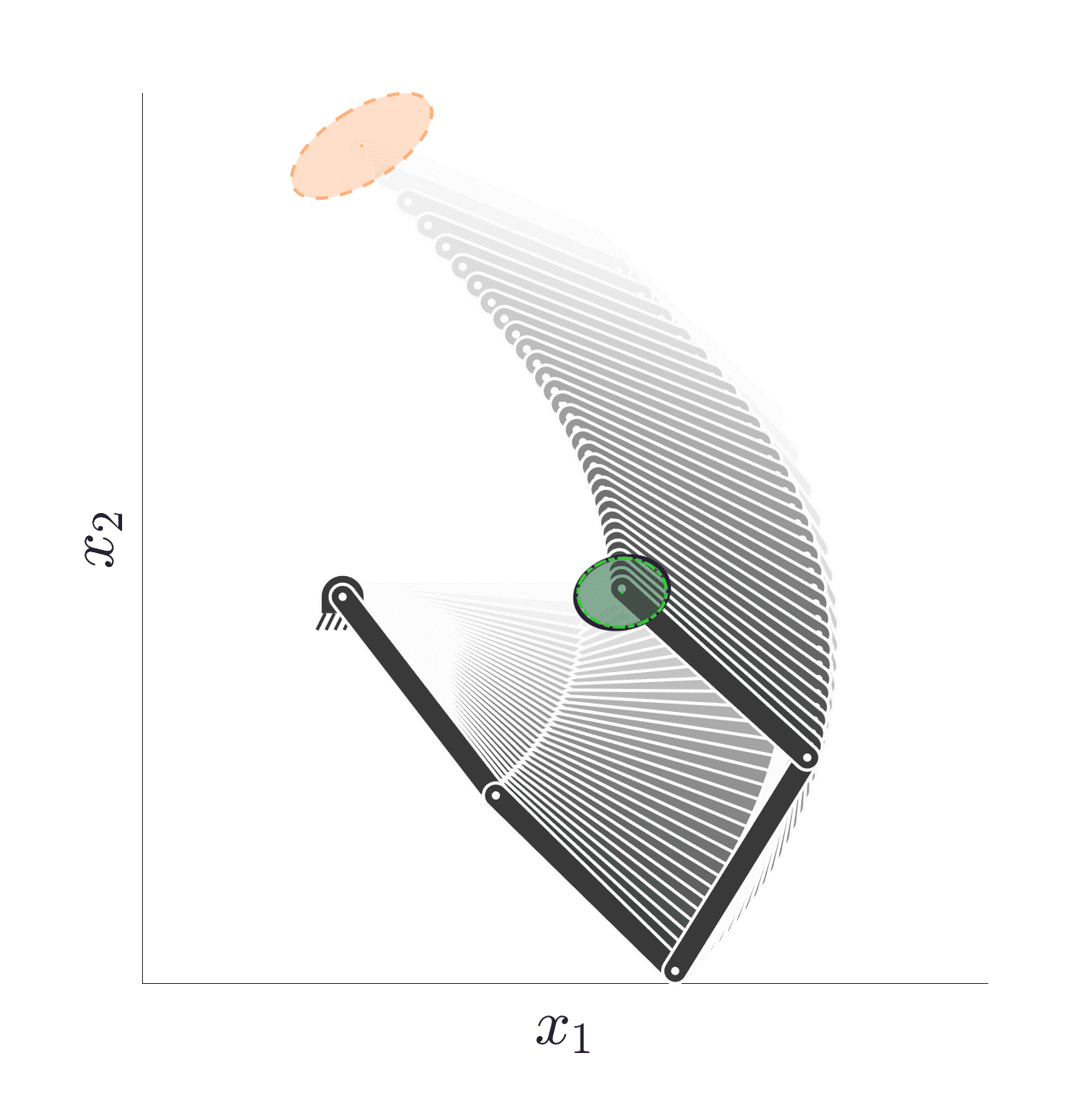}
		\caption{}
		\label{subFig:FirstTask}	
	\end{subfigure}	
	\begin{subfigure}[b]{0.24\textwidth}
		\centering
		\includegraphics[width=0.45\textwidth]{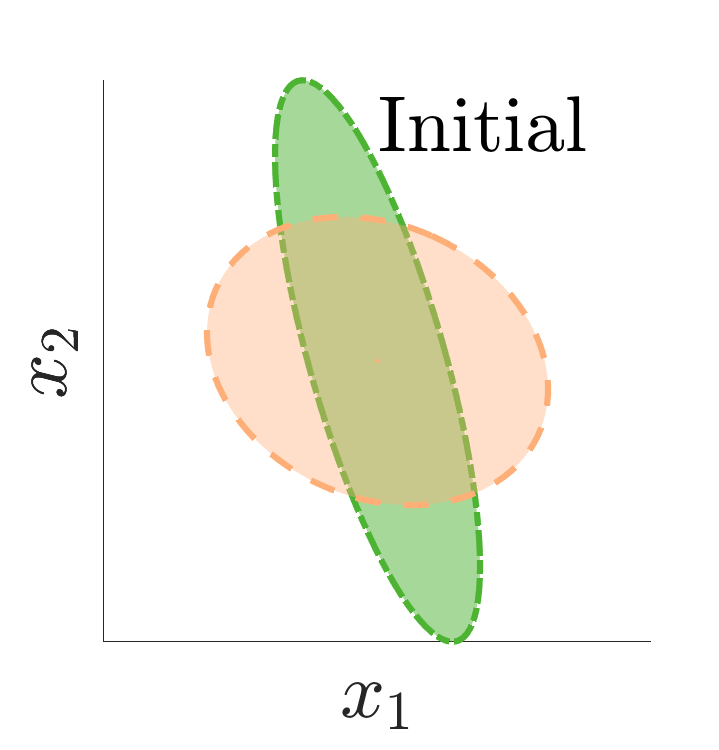}
		\includegraphics[width=0.45\textwidth]{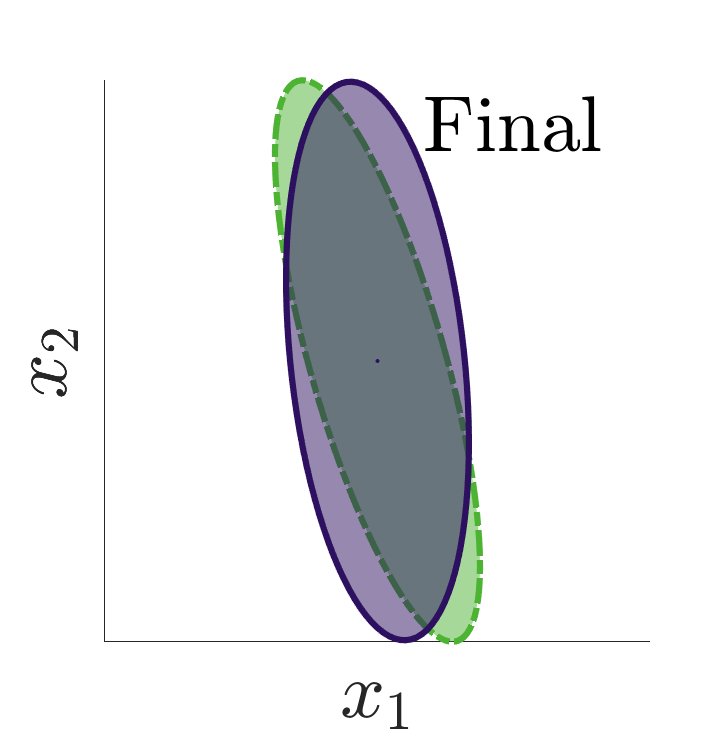}
		\includegraphics[width=\textwidth]{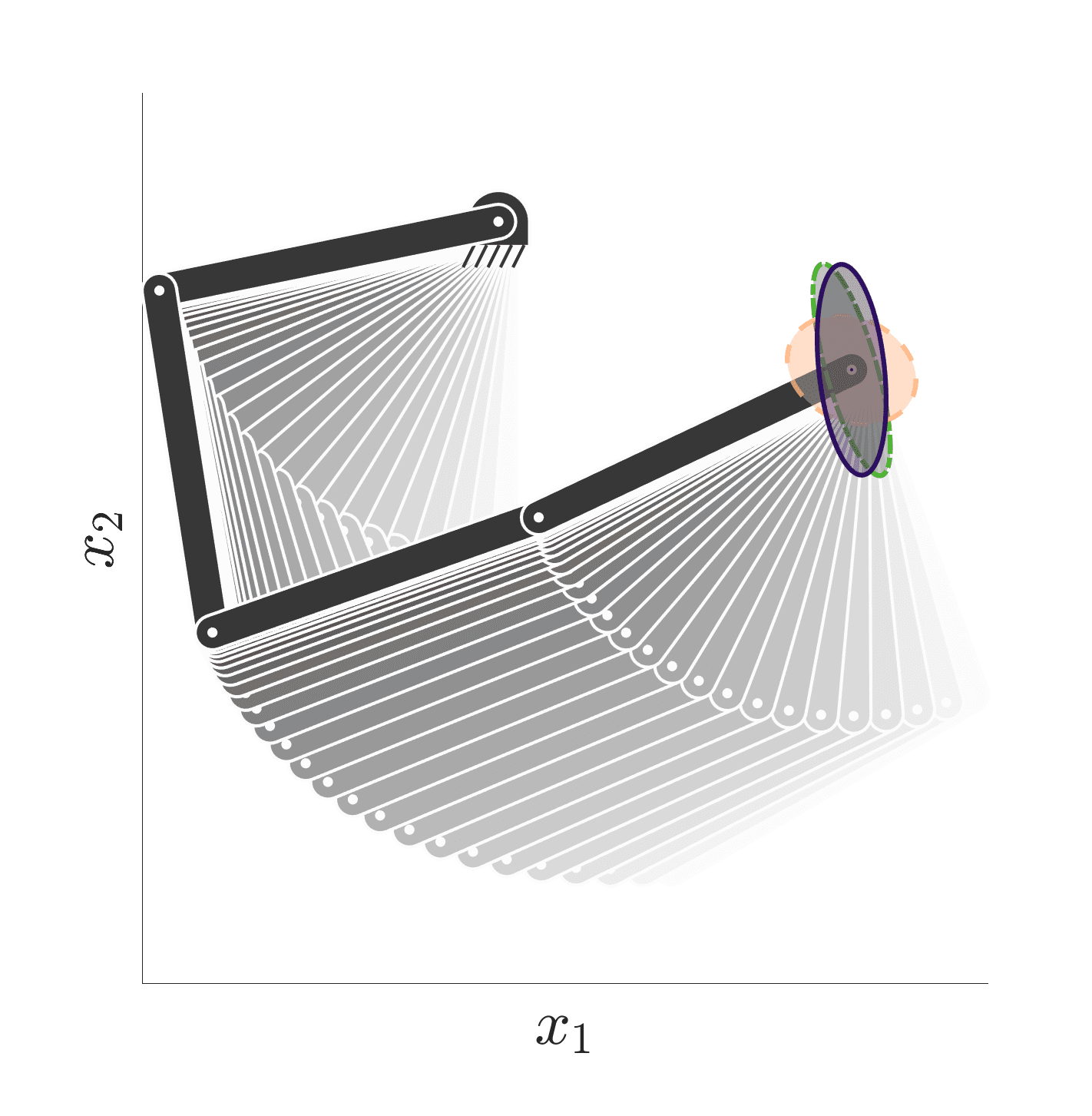}
		\caption{}
		\label{subFig:SecondTask}
	\end{subfigure}	
	\caption{(\emph{a}) Manipulability tracking as main task. (\emph{b}) Manipulability-based redundancy resolution with Cartesian position control. The robot color goes from light gray to black to show the evolution of the posture. Initial, final, and desired manipulability ellipsoids are respectively depicted in yellow, dark purple, and green. The \emph{top} rows show close-up plots corresponding to the initial and final manipulability.}
	\label{Fig:FirstTask}
	\vspace{-0.2cm}
\end{figure}
\begin{table}[t]
	\renewcommand*{\arraystretch}{1.2}
	\caption{Initial and final distances $d(\bm{\hat{M}},\bm{M}_t)$ between the current and desired manipulability for the experiments illustrated in Fig.~\ref{Fig:FirstTask}.}
	\label{Tab:TrackingDistances}
	\begin{center}
		\begin{tabular}{c|c|c|}
			& Initial & Final\\
			\hline
			Main task & $1.342$ & $0.199$\\
			Redundancy resolution & $2.194$ & $0.955$\\
			\hline
		\end{tabular}
	\end{center}
\end{table}
Note that matricization and vectorization operations can be defined using Mandel notation to alleviate the computational cost of the controllers using tensor representations, such that
\begin{equation}
\mathcal{X}_{(3)} \!=\! \begin{pmatrix}\text{vec}\left(\bm{\mathcal{X}}_{:,:,1}\right)^\trsp\\  \vdots\\ \text{vec}\left(\bm{\mathcal{X}}_{:,:,K}\right)^\trsp \end{pmatrix}
\text{ and } 
\text{vec}\Bigg(\!\!\begin{pmatrix}\alpha & \beta\\ \beta & \gamma\end{pmatrix}\!\!\Bigg) \!=\! \begin{pmatrix}\alpha\\ \gamma\\ \sqrt{2}\beta \end{pmatrix} ,
\label{Eq:MandelVector}
\end{equation}
for $2\!\times\!2\!\times\!K$ third-order tensors and $2\!\times\!2$ matrices.

In order to show the functionality of the proposed approach where the goal of the robot is to reproduce a given manipulability ellipsoid either as its main task or as a secondary objective, we carried out experiments with a simulated 4-DoF planar robot. In the first case, the robot is required to vary its joint configuration to make its manipulability ellipsoid $\bm{M}_t$ coincide with the desired one $\bm{\hat{M}}$, without any task requirement at the level of its end-effector. In the second case, the robot needs to keep its end-effector at a fixed Cartesian position while moving its joints to match the desired manipulability ellipsoid. Fig.~\ref{Fig:FirstTask} shows how the manipulator configuration is successfully adjusted so that $\bm{M}_t\simeq\bm{\hat{M}}$ when the manipulability ellipsoid tracking is considered as the main task or as a secondary objective (see Table~\ref{Tab:TrackingDistances}). These results show that our geometry-aware controllers inspired by the inverse kinematics formulation are suitable to solve the manipulability ellipsoid tracking problem.

\subsubsection*{Stability analysis}
We here analyze the stability properties of the proposed manipulability tracking controller given the geometry of the underlying manifold. First of all, note that the dynamical system operated by the controller~\eqref{Eq:ManipTrackFirstTask} corresponds to
\begin{equation}
\bm{\dot{M}} = k_{\bm{M}} \, \text{Log}_{\bm{M}}(\bm{\hat{M}}) ,
\label{Eq:ManipDynSys}
\end{equation}
where the controller gain is assumed to be a positive scalar value for sake of simplicity. Then, we select the Lyapunov function $V$ as
\begin{equation}
V(\bm{M}) \;\;=\;\; \langle\bm{F},\bm{F}\rangle_{\bm{\hat{M}}},
\label{Eq:LyapFun}
\end{equation}
where $\bm{F}=\text{Log}_{\bm{\hat{M}}}(\bm{M})$ is a vector field composed of the initial velocities of all geodesics departing from the origin $\bm{\hat{M}}$, and $\langle \cdot,\cdot \rangle_{\bm{\hat{M}}}$ is the inner product \eqref{Eq:SPDinnerprod}. As proved in~\citep{Pait10}, the function \eqref{Eq:LyapFun} is a Lyapunov function for a dynamical system $\bm{\dot{M}} = h(\bm{M})$ such that $h(\bm{\hat{M}})=0$ if the Lie derivative $\mathcal{L}_h V(\bm{M}) = 2\langle h,\bm{F}\rangle_{\bm{\hat{M}}}$ is negative everywhere except at the origin $\bm{\hat{M}}$. To verify this condition, we first express the velocity of the dynamical system \eqref{Eq:ManipDynSys} in the tangent space of $\bm{\hat{M}}$ using parallel transport as
\begin{equation}
\Gamma_{\bm{M}\to\bm{\hat{M}}}(\bm{\dot{M}}) = -k_{\bm{\hat{M}}} \, \text{Log}_{\bm{\hat{M}}}(\bm{M}).
\label{Eq:ManipDynSys2}
\end{equation}
The Lie derivative $\mathcal{L}_h V$ of the proposed Lyapunov function for the dynamical system \eqref{Eq:ManipDynSys2} is given by
\begin{align}
\mathcal{L}_h V(\bm{M}) \;\; &=\;\; 2\langle -k_{\bm{\hat{M}}} \, \text{Log}_{\bm{\hat{M}}}(\bm{M}), \text{Log}_{\bm{\hat{M}}}(\bm{M}) \rangle_{\bm{\hat{M}}} \nonumber\\
&= -2k_{\bm{\hat{M}}} \;\langle \text{Log}_{\bm{\hat{M}}}(\bm{M}), \text{Log}_{\bm{\hat{M}}}(\bm{M}) \rangle_{\bm{\hat{M}}} \nonumber\\
&= -2k_{\bm{\hat{M}}} V. 
\label{Eq:LyapDeriv}
\end{align}
Therefore, we have
\begin{align*}
&V(\bm{M})>0,\;\;\mathcal{L}_h V(\bm{M}) <0 \qquad\forall\;\;\bm{M}\neq \bm{\hat{M}},\\
&V(\bm{M})=\mathcal{L}_h V(\bm{M})=0 \iff \bm{M}= \bm{\hat{M}},
\label{Eq:LyapStab}
\end{align*}
so that the function \eqref{Eq:LyapFun} is a valid Lyapunov function. 
Moreover, by observing that $V(\bm{M}) = d^2(\bm{M},\bm{\hat{M}})$ with $d(\cdot, \cdot)$ the affine-invariant distance~\eqref{Eq:SPDdist}, we have
\begin{align*}
&c_1 d^2(\bm{M},\bm{\hat{M}}) \leq V(\bm{M}) \leq c_2 d^2(\bm{M},\bm{\hat{M}}),\\
&\mathcal{L}_h V(\bm{M}) \leq -c_3 d^2(\bm{M},\bm{\hat{M}}),
\end{align*}
where $0 < c_1 \leq 1$, $c_2\geq 1$, $c_3 = 2k_{\bm{\hat{M}}} > 0$. This implies that the controller~\eqref{Eq:ManipTrackFirstTask} is exponentially stable (see e.g.,~\citep{Wu20:LyapunovManifolds}). It can be easily shown that this result holds with ${c_3 = 2 \lambda_{\min}(\bm{K}_{\bm{\hat{M}}})}$ for a positive-definite controller gain matrix $\bm{K}_{\bm{\hat{M}}}$, where $\lambda_{\min}(\cdot)$ returns the minimum eigenvalue of the matrix.

Note that the Lyapunov function \eqref{Eq:LyapFun} is similar to the one usually defined to demonstrate the exponential stability of the classical inverse kinematic-based velocity controller $\bm{\dot{q}}_t = \bm{J}^\dagger \, \bm{K}_{\bm{x}} \, (\bm{\hat{x}}_t-\bm{x}_t)$. In that case, the Lyapunov function is defined as $V(\bm{x}) \!=\! (\bm{\hat{x}}-\bm{x})^\trsp(\bm{\hat{x}}-\bm{x})$, which is equivalent to the inner product $\langle \bm{e},\bm{e}\rangle$ with the error $\bm{e}=\bm{\hat{x}}-\bm{x}$. In the case of manipulability tracking, the inner product $\langle \cdot,\cdot \rangle$ is defined in the SPD manifold and the error $\bm{e}$ is computed as $\text{Log}_{\bm{\hat{M}}}(\bm{M})$. Finally, it is worth highlighting that when the manipulability tracking is assigned a secondary role, the controller~\eqref{Eq:ManipTrackSecTask} does not influence the stability of the main task of the robot as the manipulability-based redundancy resolution is carried out in the corresponding nullspace.

\subsubsection{Acceleration-based controller}
Similarly to the velocity-based controller, we propose a geometry-aware acceleration-based controller that allows the computation of the joint accelerations $\bm{\ddot{q}}$ required to track a desired manipulability trajectory (i.e. desired manipulability and manipulability velocity profiles). The approach is inspired by the inverse kinematics formulation and its differential relationships used to compute the joint accelerations necessary to track desired end-effector positions and velocities.

To formalize the acceleration-based controller, let us first define the second-order time derivative of the manipulability ellipsoid computed from \eqref{Eq:ManipJacobian} by applying the product rule
\begin{equation}
\frac{\partial^2 \bm{M}(t)}{\partial t^2} = \bm{\mathcal{J}}(\bm{q}) \times_3 \ddot{\bm{q}}^\trsp + \bm{\mathcal{\dot{J}}}(\bm{q}) \times_3 \bm{\dot{q}}^\trsp,
\label{Eq:ManipSecDeriv}
\end{equation}
(see Appendix~\ref{sec:AppendixDiffManipJacobian} for details on the computation of $\bm{\mathcal{\dot{J}}}(\bm{q})$).
So, by minimizing the $\ell^2$-norm of the residuals, we can compute the required joint accelerations of the robot to track a desired trajectory of manipulability ellipsoids as its main task with
\begin{equation}
\ddot{\bm{q}} = 
(\bm{\mathcal{J}}_{(3)}^\dagger)^\trsp 
\Big(\text{vec}(\ddot{\bm{M}}) -\bm{\mathcal{\dot{J}}}_{(3)}^\trsp\bm{\dot{q}}\Big).
\label{Eq:InvManipAcc}
\end{equation}
Similarly as in the classical acceleration-based controller that tracks a desired end-effector trajectory, we can define a controller to track a reference manipulability ellipsoid trajectory based on \eqref{Eq:InvManipAcc}. To do so, we exploit the geometry of the SPD manifold to compute the difference between the current manipulability ellipsoid $\bm{M}_t$ and the desired one $\bm{\hat{M}}_t$, as previously specified for the velocity-based controller. Moreover, since the first-order time derivative of manipulability ellipsoids lies on the tangent space of the SPD manifold (i.e. the space of symmetric matrices $\text{Sym}^D$), the difference between the current manipulability velocity $\bm{\dot{M}}_t$ and the desired one $\bm{\widehat{\dot{M}}}_t$ is computed as a subtraction in the Euclidean space. Therefore, a reference manipulability acceleration command can be specified by
\begin{equation}
\text{vec}(\bm{\ddot{M}}_t) = \bm{K}_p \text{vec}\Big(\text{Log}_{\bm{M}_t}(\bm{\hat{M}}_t)\Big) + \bm{K}_d \text{vec}\big(\widehat{\bm{\dot{M}}_t} - \bm{\dot{M}}_t \big),
\label{Eq:RefManipAcc}
\end{equation}
which resembles a proportional-derivative controller where $\bm{K}_p$ and $\bm{K}_d$ are gain matrices. Then, the reference joint acceleration $\ddot{\bm{q}}$ can be computed using \eqref{Eq:InvManipAcc} and \eqref{Eq:RefManipAcc}. Note that this reference joint acceleration can correspond to a main task of the robot or to a secondary tracking objective. In the latter case, a manipulability-based redundancy resolution can also be implemented in a similar way as \eqref{Eq:ManipTrackSecTask}. 


\subsection{Actuators contribution}
\label{subsec:ActuatorsContrib}
In many practical applications, the joint velocities of the robot are limited. The definition of manipulability ellipsoid can then be extended to include these actuation constraints, as shown in~\citep{Lee97}. We here provide the definition of the force, velocity and dynamic manipulability ellipsoids and the corresponding manipulability Jacobians considering joint actuation constraints.

\begin{figure}[tbp]
	\begin{subfigure}[b]{0.26\textwidth}
		\centering
		\includegraphics[width=\textwidth]{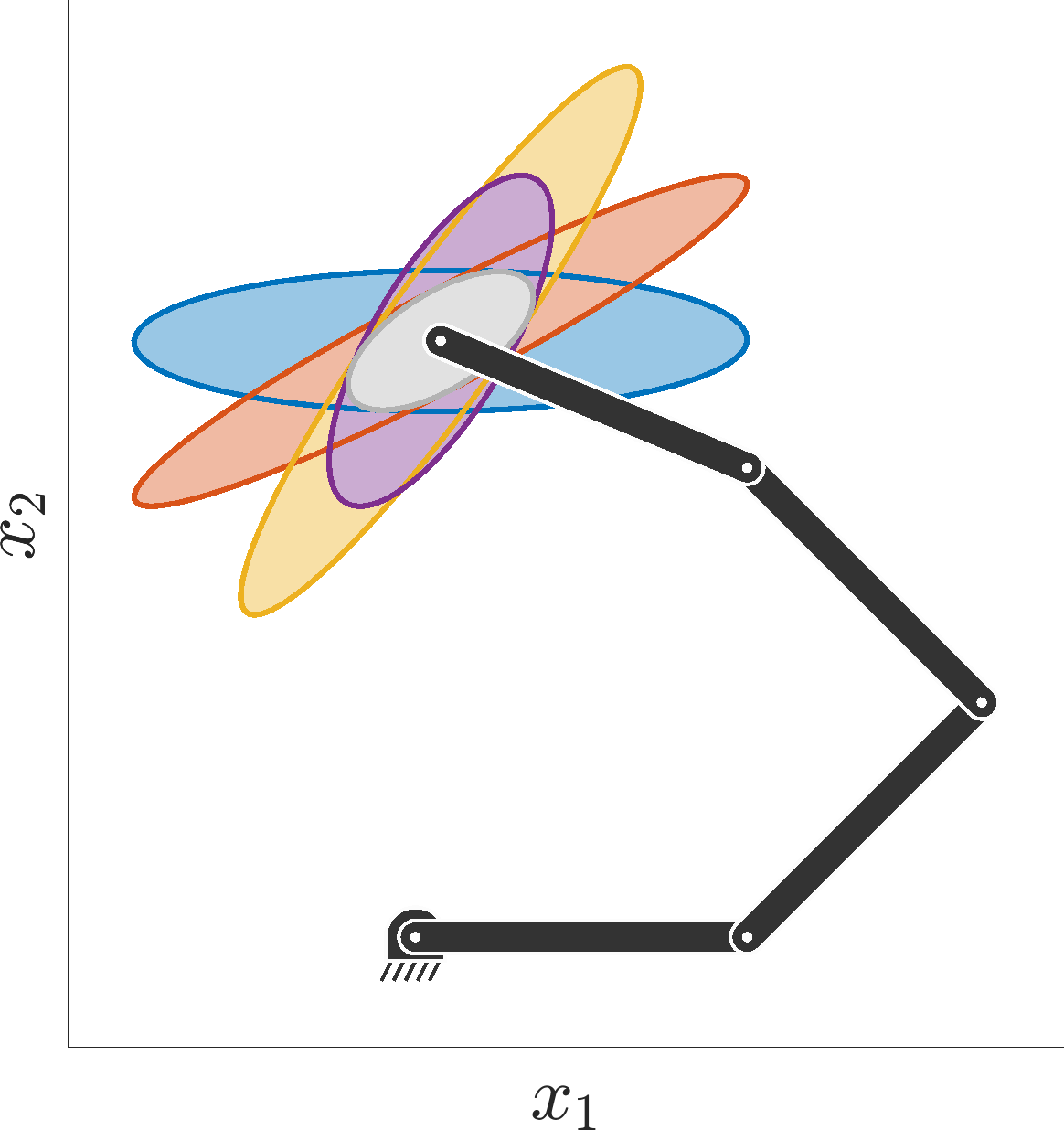}
		\caption{}
		\label{subFig:WeightedManipulability}
	\end{subfigure}
	\begin{subfigure}[b]{0.22\textwidth}
		\centering
		\includegraphics[width=.95\textwidth]{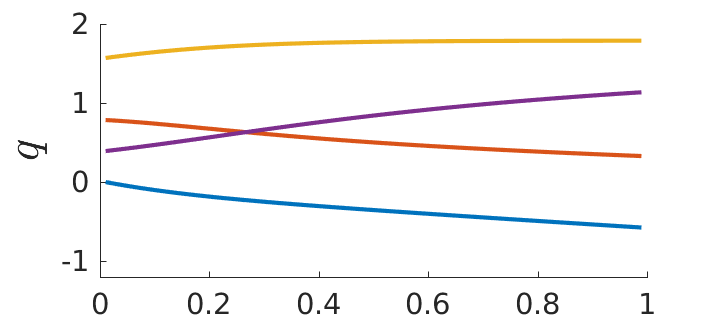}
		\includegraphics[width=.95\textwidth]{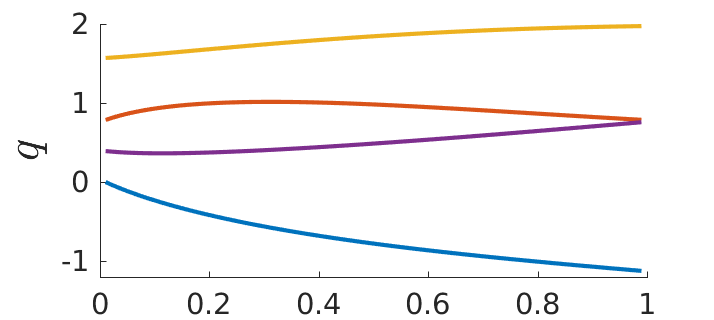}
		\includegraphics[width=.95\textwidth]{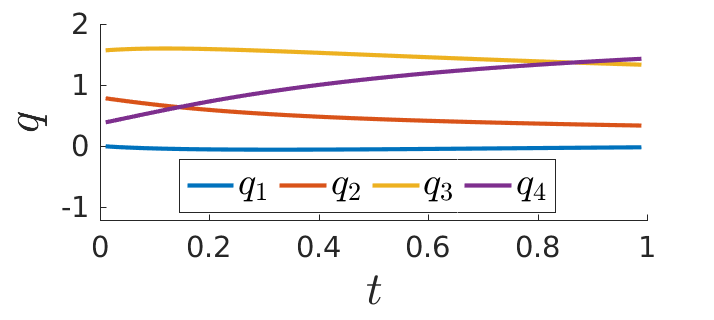}
		\caption{}
		\label{subFig:WeightedManipulability_joints}
	\end{subfigure}
	\caption{Illustration of actuators contribution. (\emph{a}) Velocity manipulability ellipsoids obtained when setting a maximum joint velocity, for each joint, five times higher than the rest. The manipulability corresponding to equal maximum joint velocity is shown in gray. (\emph{b}) Joint trajectories obtained with manipulability tracking (as in Fig.~\ref{subFig:FirstTask}) for equal maximum joint velocities (\emph{top}), highest velocity limit for $q_1$ (\emph{middle}), and lowest velocity limit for $q_1$ (\emph{bottom}).}
	\label{Fig:WeightedManipulability}
\end{figure}

To include the joint velocity constraints of the robot in the definition of the velocity manipulability ellipsoid, we use the following weighted forward kinematics formulation
\begin{equation}
\bm{\dot{x}} = \underbrace{( \bm{J} \bm{W}^{\bm{\dot{q}}} )}_{\bm{\tilde{J}}} \underbrace{( \bm{W}^{\bm{\dot{q}}-1} \bm{\dot{q}} )}_{\bm{\tilde{\dot{q}}}} ,
\label{Eq:WeightedFK}
\end{equation} 
where $\bm{W}^{\bm{\dot{q}}} = \text{diag}(\dot{q}_{1,\max}, \ldots, \dot{q}_{n,\max})$ is a diagonal matrix whose elements correspond to the maximum joint velocities of the robot. Then, considering the set of joint velocities of constant unit norm $\|\bm{\tilde{\dot{q}}}\|=1$ mapped into the Cartesian velocity space through
\begin{align} 
\| \bm{\tilde{\dot{q}}} \|^2  = \bm{\tilde{\dot{q}}}^\trsp \bm{\tilde{\dot{q}}}  = \bm{\dot{x}}^\trsp(\bm{\tilde{J}}\bm{\tilde{J}}^\trsp)^{-1}\bm{\dot{x}},
\label{Eq:WeightedVelocityMapping}
\end{align}
the velocity manipulability ellipsoid is given by $\bm{\tilde{M}}^{\bm{\dot{x}}} = \bm{\tilde{J}}\bm{\tilde{J}}^\trsp = \bm{J} \bm{W}^{\bm{\dot{q}}}\bm{W}^{\bm{\dot{q}}\trsp} \bm{J}^\trsp$, which represents the flexibility of the manipulator in generating velocities in Cartesian space considering its maximum joint velocities as illustrated in Figure~\ref{subFig:WeightedManipulability}. Note that the actuators contribution $\bm{W}^{\bm{\dot{q}}}\bm{W}^{\bm{\dot{q}}\trsp}$ also has a geometrical interpretation based on the fact that the robot joint position $\bm{q}$ lies on the flat $n$-torus manifold~\citep{Park95}.

By following the methodology of Section~\ref{subsec:ManJac}, the change in the robot manipulability ellipsoid is related to the joint velocity via
\begin{equation}
\frac{\partial \bm{\tilde{M}}(t)}{\partial t} = 
\bm{\mathcal{\tilde{J}}}(\bm{q}) \times_3 \bm{\dot{q}}^\trsp.
\label{Eq:WeightedManipJacobian}
\end{equation}
Therefore, the velocity manipulability Jacobian including joint velocity limits is given by
\begin{equation}
\bm{\mathcal{\tilde{J}}}^{\bm{\dot{x}}} = \frac{\partial\bm{J}}{\partial \bm{q}}\times_2\bm{J} \bm{W}^{\bm{\dot{q}}}\bm{W}^{\bm{\dot{q}}\trsp} + \frac{\partial\bm{J}^\trsp}{\partial \bm{q}}\times_1\bm{J}\bm{W}^{\bm{\dot{q}}}\bm{W}^{\bm{\dot{q}}\trsp}.
\label{Eq:WeightedVelocityManipJacobian}
\end{equation} 
Figure~\ref{subFig:WeightedManipulability_joints} shows the effect of including the actuator contribution when tracking a velocity manipulability ellipsoid. Notice that the robot joint $q_1$ significantly moves when given the highest velocity limit. In contrast, its influence on the manipulability tracking task is minimal when given the lowest velocity limit. This demonstrates the importance of considering the robot actuator specifications when tracking manipulability ellipsoids in real platforms.  

In a similar way, the force manipulability ellipsoid considering the maximum joint torques is defined as ${\bm{\tilde{M}}^{\bm{F}}\!=\!(\bm{J} \bm{\Omega^\tau} \bm{J}^\trsp)^{-1}}$, where $\bm{\Omega^\tau}\!=\!(\bm{W}^{\bm{\tau}}\bm{W}^{\bm{\tau}\trsp})^{-1}$ and $\bm{W}^{\bm{\tau}} = \text{diag}(\tau_{1,\max}, \ldots, \tau_{n,\max})$. Then, the corresponding manipulability Jacobian is given by
\begin{equation}
\footnotesize
\bm{\mathcal{\tilde{J}}}^{\bm{F}}\!=\! -\left(\frac{\partial\bm{J}}{\partial \bm{q}} \times_2\bm{J}\bm{\Omega^\tau} + \frac{\partial\bm{J}^\trsp}{\partial \bm{q}} \times_1\bm{J}\bm{\Omega^\tau} \right) \nonumber \times_1 (\bm{\tilde{M}}^{{\bm{F}}})^{-1} \times_2 (\bm{\tilde{M}}^{{\bm{F}}})^{-1}.
\label{Eq:WeightedForceManipJacobian}
\end{equation}
Finally, the dynamic manipulability ellipsoid considering the maximum joint torques is $\bm{\tilde{M}}^{\ddot{\bm{x}}} = \bm{\Upsilon} {\bm{\Omega^{\tau}}}^{-1} \bm{\Upsilon}^{\trsp}$ with corresponding manipulability Jacobian defined as 
\begin{equation}
\bm{\mathcal{\tilde{J}}}^{\ddot{\bm{x}}} = \frac{\partial\bm{\Upsilon}}{\partial \bm{q}}\times_2 \bm{\Upsilon} {\bm{\Omega^{\tau}}}^{-1} + \frac{\partial\bm{\Upsilon}^\trsp}{\partial \bm{q}}\times_1 \bm{\Upsilon} {\bm{\Omega^{\tau}}}^{-1}.
\label{Eq:WeightedDynManipJacobian}
\end{equation}	

\begin{figure*}[tbp]
	\centering
	\begin{subfigure}[b]{1\textwidth}
		\centering
		\begin{subfigure}[b]{0.19\textwidth}
			\centering
			\includegraphics[width=.88\textwidth]{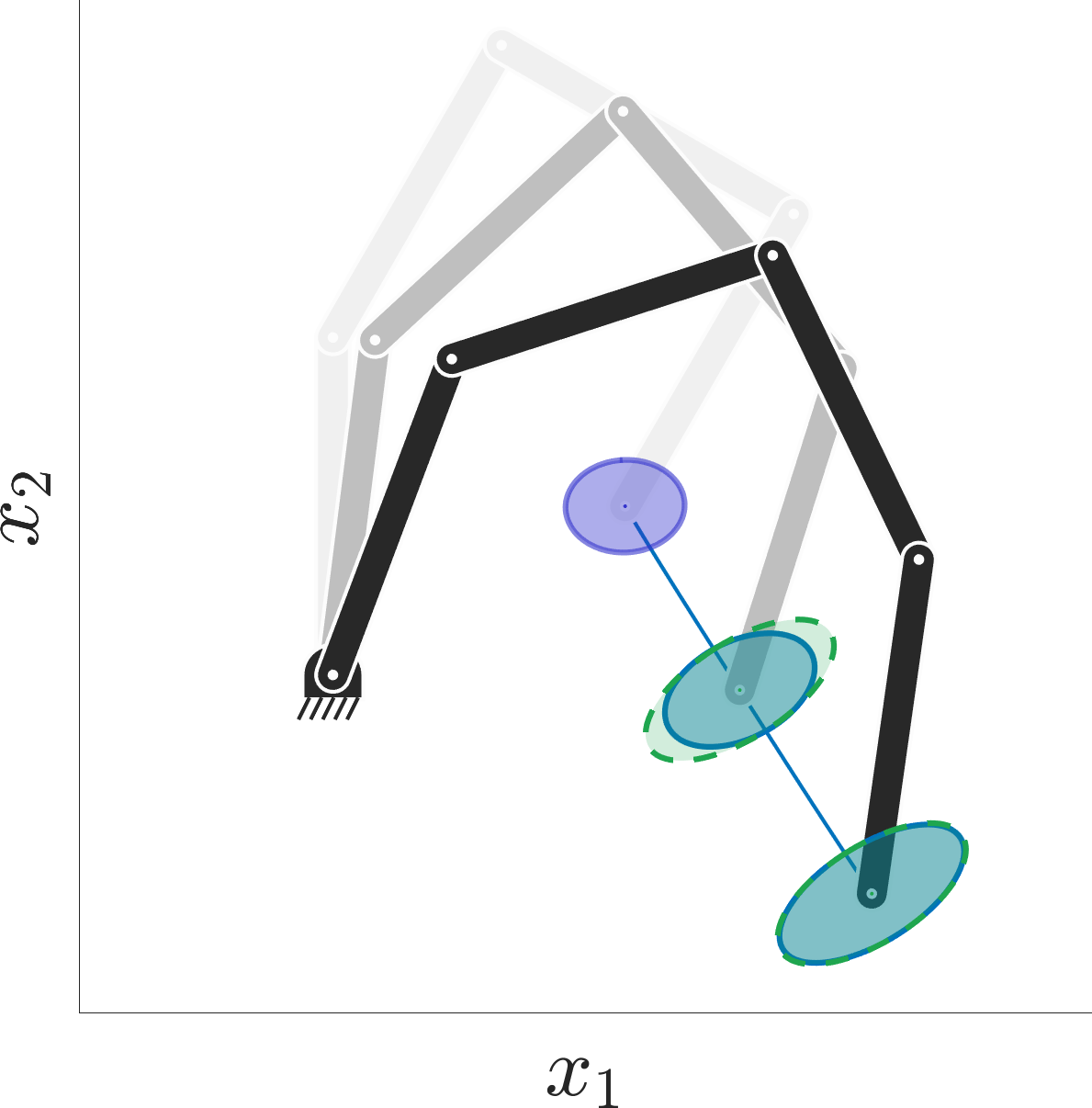}
			\includegraphics[width=.95\textwidth]{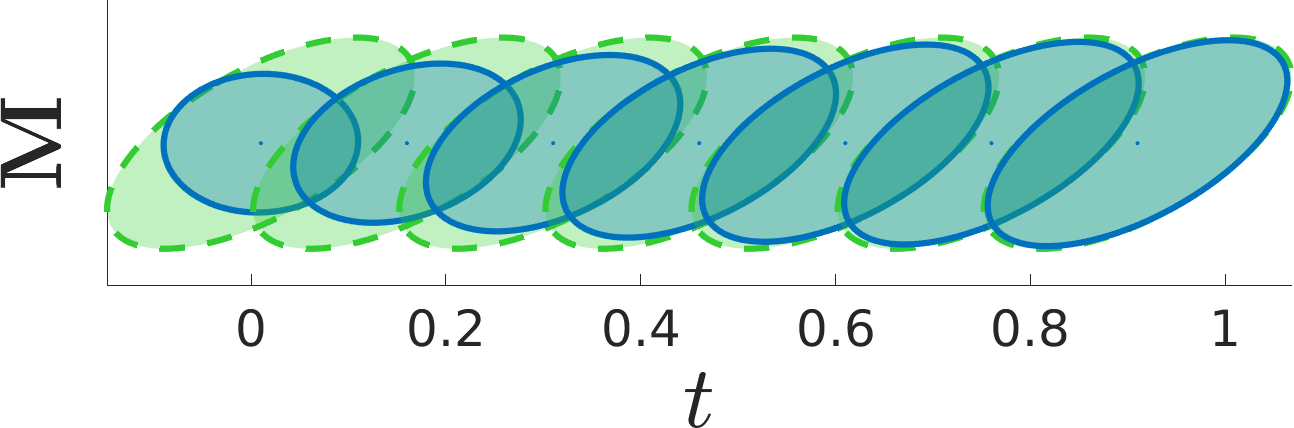}
			\caption{Baseline}
			\label{subFig:Precision_bsln}
		\end{subfigure}
		\begin{subfigure}[b]{0.19\textwidth}
			\centering
			\includegraphics[width=.88\textwidth]{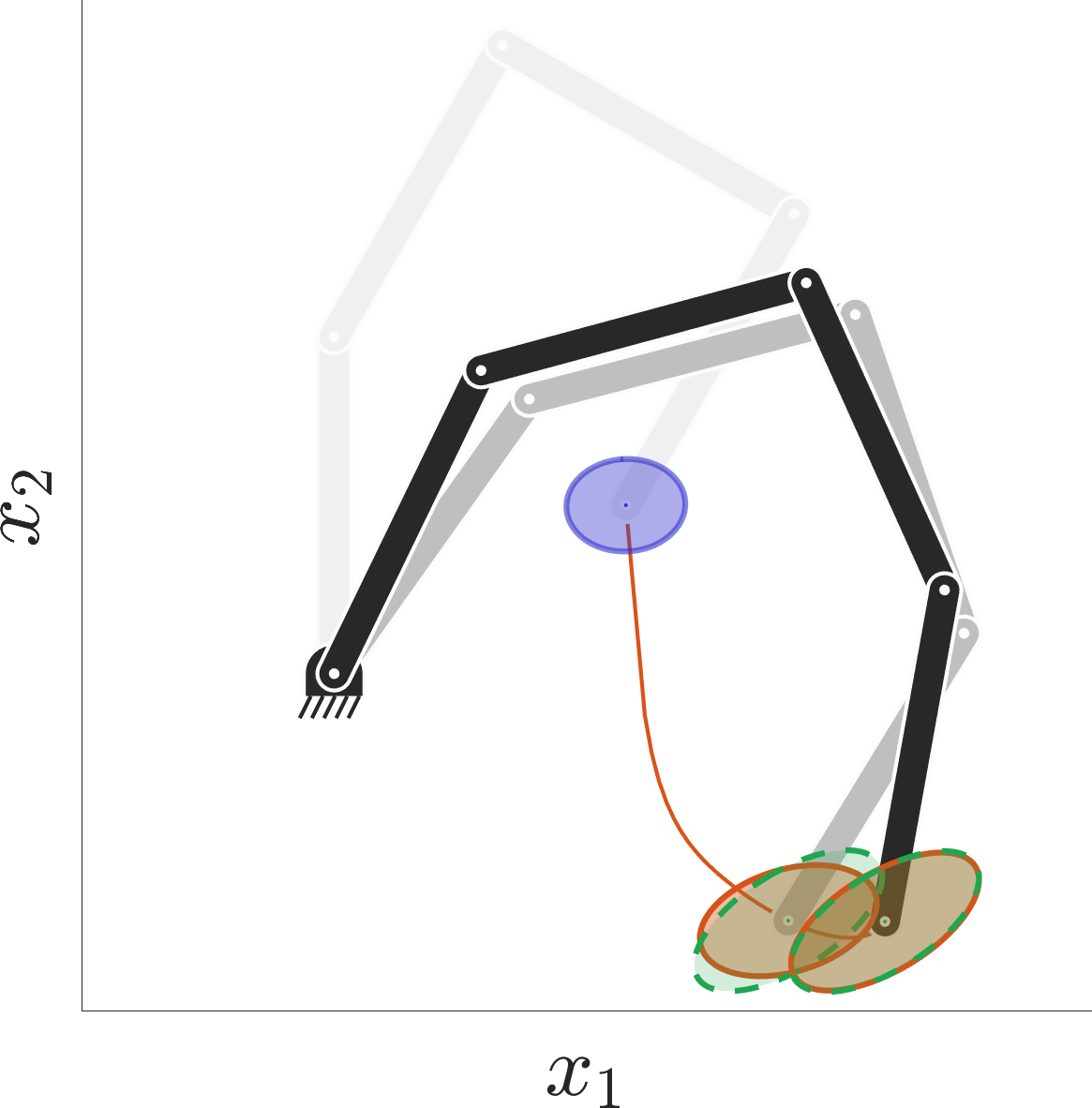}
			\includegraphics[width=.95\textwidth]{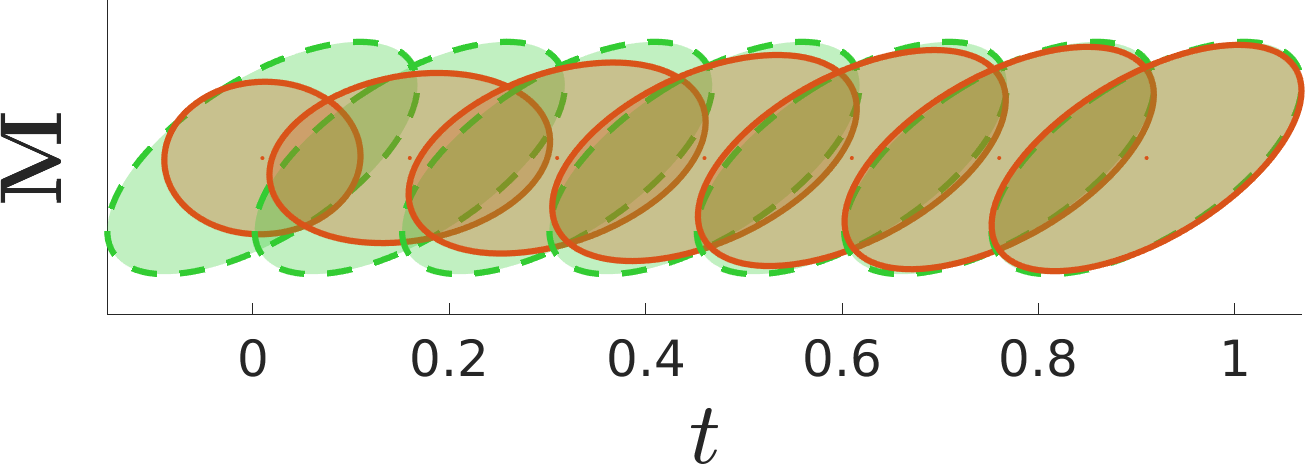}
			\caption{Precision along $x_1$}
			\label{subFig:Precision_x}
		\end{subfigure}
		\begin{subfigure}[b]{0.19\textwidth}
			\centering
			\includegraphics[width=.88\textwidth]{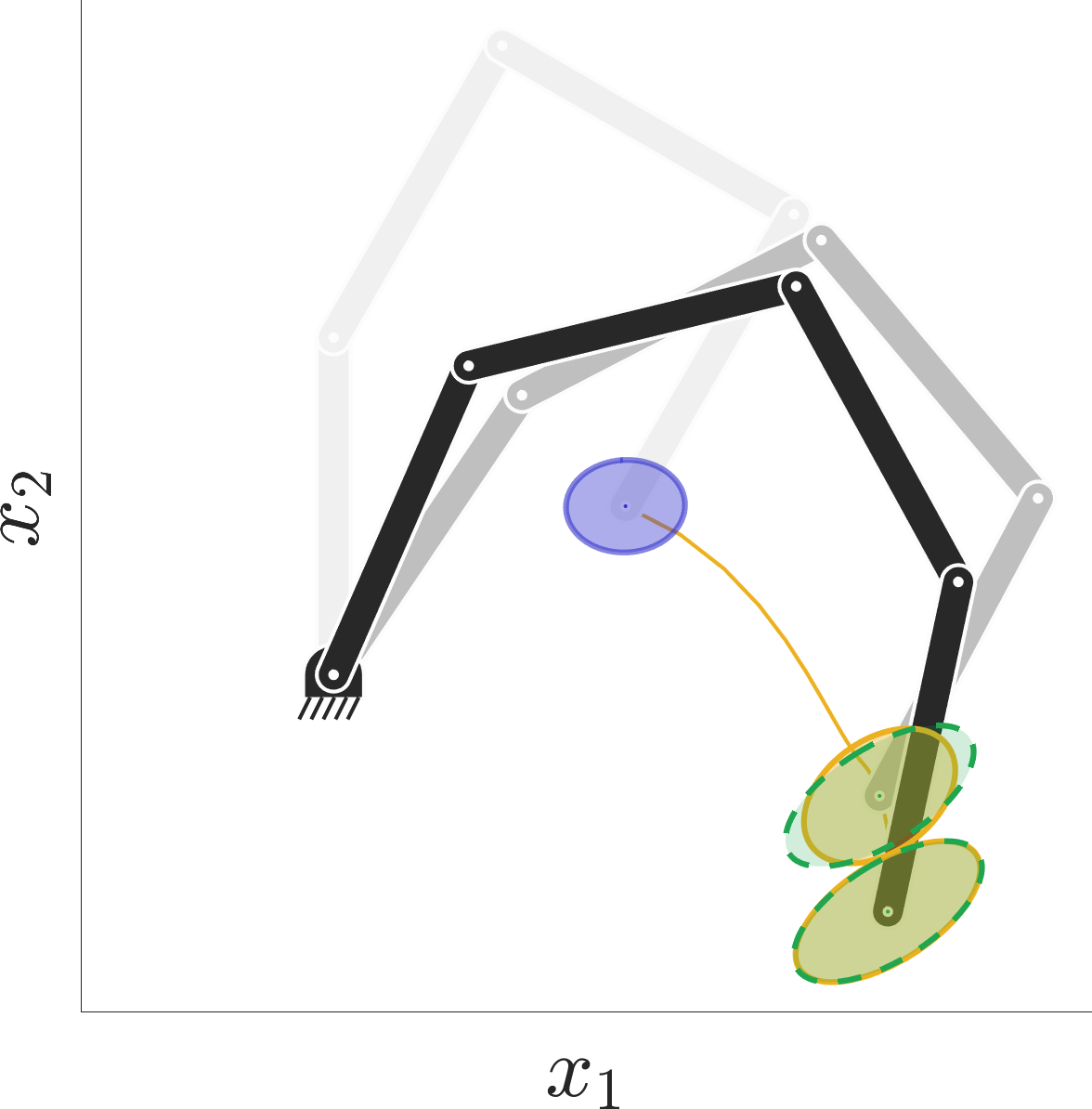}
			\includegraphics[width=.95\textwidth]{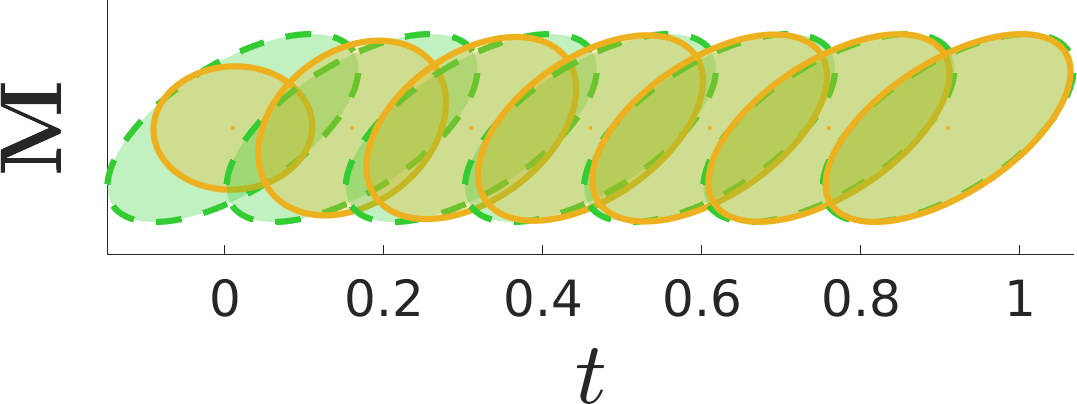}
			\caption{Precision along $x_2$}
			\label{subFig:Precision_y}
		\end{subfigure}
		\begin{subfigure}[b]{0.19\textwidth}
			\centering
			\includegraphics[width=.88\textwidth]{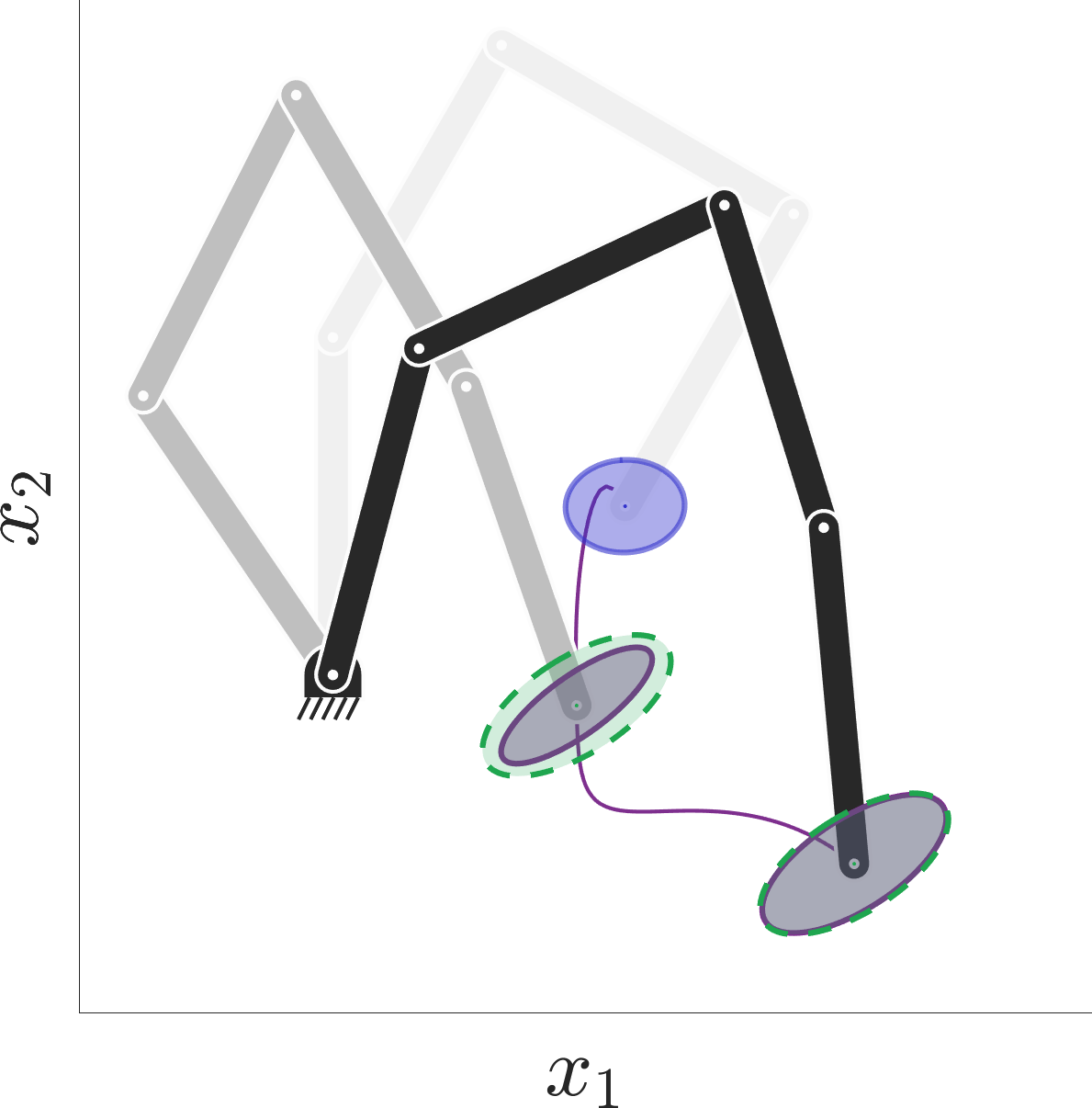}
			\includegraphics[width=.95\textwidth]{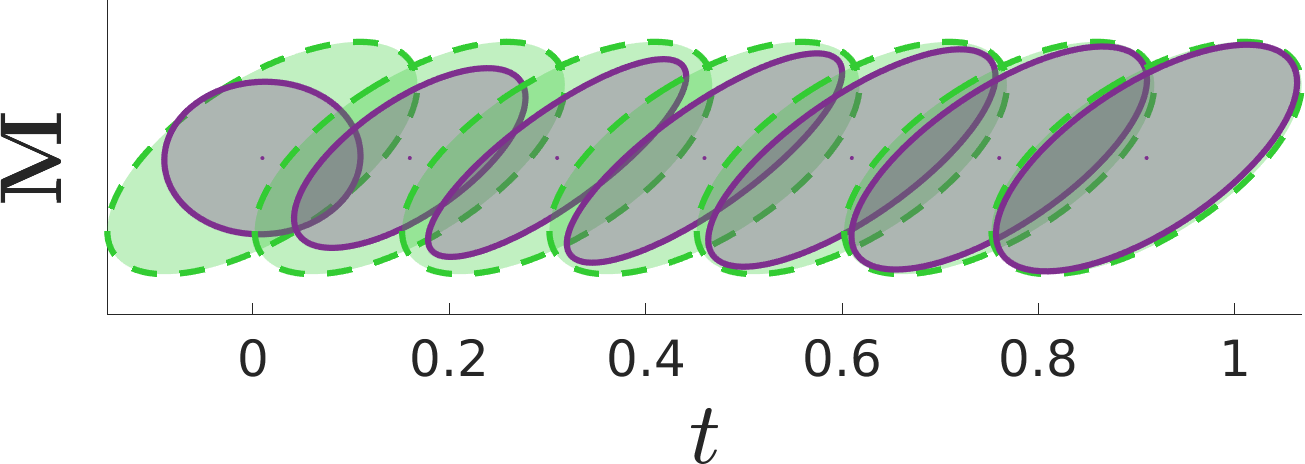}
			\caption{Diagonal precision}
			\label{subFig:Precision_xy}
		\end{subfigure}
		\begin{subfigure}[b]{0.19\textwidth}
			\centering
			\includegraphics[width=.88\textwidth]{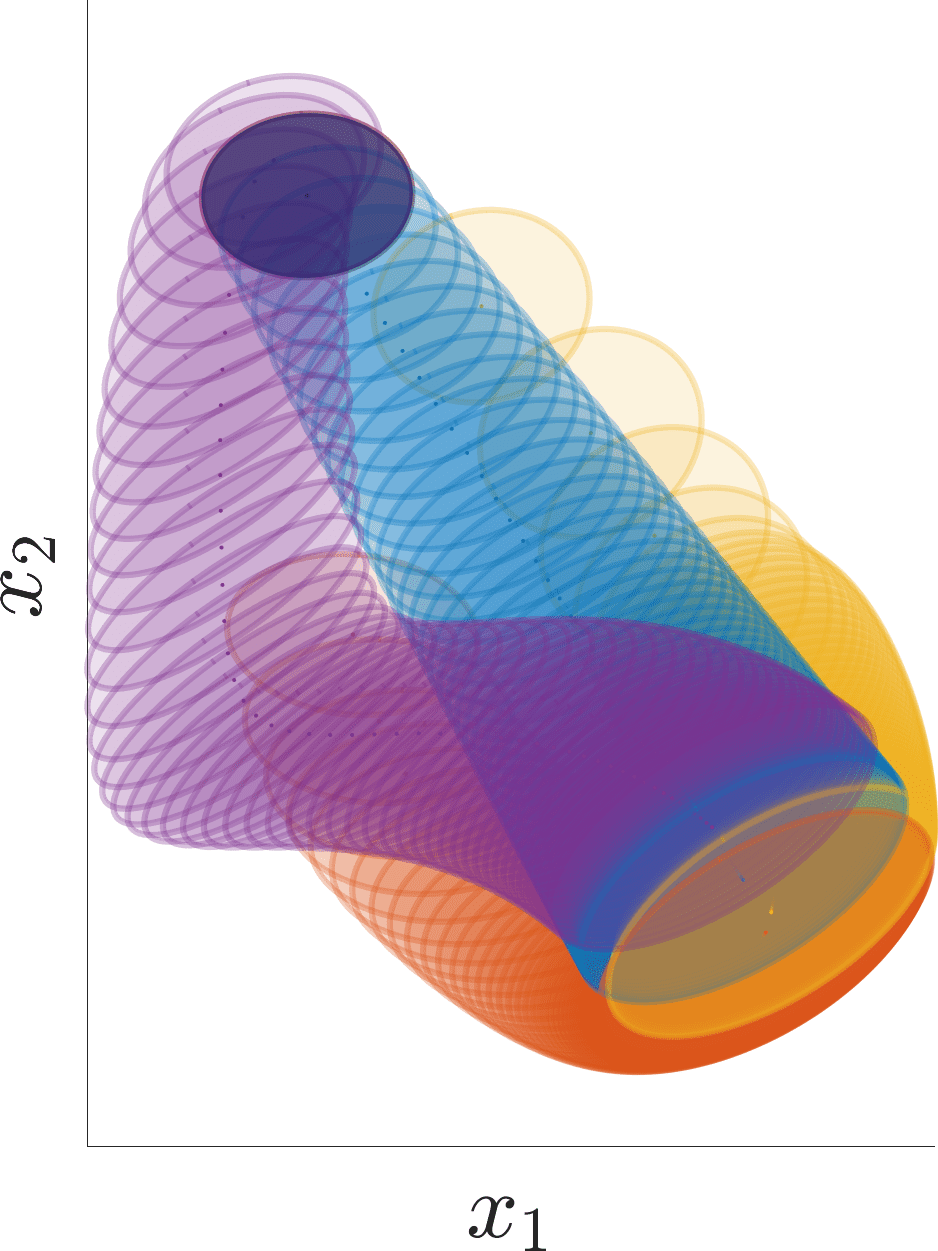}
			\caption{Trajectories}
			\label{Fig:Precision_trajectories}
		\end{subfigure}
	\end{subfigure}
	\caption{Manipulability tracking as main task with diagonal gain matrices defined from different precision tensors. The \emph{top} plots depict the end-effector trajectory (solid colored line) and the posture of the robot along with the corresponding manipulability at time $t=0$, $0.25$ and $1$s. The evolution of the manipulability along time is shown in the \emph{bottom} plots. (\emph{a}): equal tracking precision for all components. (\emph{b}) and (\emph{c}): tracking precision is $10\!:\!1$ higher for $x_1$ and $x_2$, respectively. (\emph{d}): correlation between $x_1$ and $x_2$ axes is assigned a high tracking accuracy. (\emph{e}) Evolution of the robot manipulability and end-effector trajectory for the gain matrices used in (\emph{a})-(\emph{d}). The colors match those of the previous graphs. Initial and desired manipulability ellipsoids are depicted in dark blue and green on all graphs. Time $t$ is in seconds.}
	\label{Fig:TaskWithPrecision}
\end{figure*}


\subsection{Exploiting 4th-order precision matrix as controller gain}
\label{subsec:4orderCov}
An open problem regarding the proposed tracking approach is how to specify the values of the gain matrix $\bm{K}_{\bm{M}}$, which basically determines how the manipulability tracking error affects the resulting joint velocities. In this sense, we propose to define $\bm{K}_{\bm{M}}$ as a precision matrix, which describes how accurately the robot should track a desired manipulability ellipsoid. In learning from demonstration applications, such gain matrix would typically be set as proportional to the inverse of the observed covariance $\bm{\mathcal{S}}$ (see Section~\ref{subsec:GMR}). This encapsulates variability information of the task to be learned. Our goal here is to exploit this information to demand the robot a high precision tracking for directions in which low variability is observed, and vice-versa. 

We therefore introduce the required precision $\bm{\mathcal{S}}^{-1}$ for a given manipulability tracking task into the controllers defined in Section \ref{subsec:ManTrackForm}. To do so, we define the gain matrix $\bm{K}_{\bm{M}}$ as a function of the precision tensor. Specifically, we define the controller gain matrix as a full SPD matrix, which is computed from the matricization of the precision tensor $\bm{\mathcal{S}}^{-1}$ along its two first dimensions, with a proportion defined by
\begin{equation}
\bm{K}_{\bm{M}} \,\propto\, \mathcal{S}^{-1}_{(1,2)}.
\label{Eq:FullGainMat}
\end{equation}

To show how precision matrices work as controller gains in our manipulability tracking problem, we tested different forms of $\bm{K}_{\bm{M}}$ aimed at reproducing a given manipulability ellipsoid as a main task with a simulated 4-DoF planar robot. The robot is required to move its joints to track a desired manipulability ellipsoid, where the controller gain matrix $\bm{K}_{\bm{M}}$ is a diagonal matrix with the diagonal elements of ~\eqref{Eq:FullGainMat} to take into account the variation of each component of the manipulability ellipsoid. We tested four different precision tensors. First, equal variability for all components of the manipulability ellipsoid matrix is given. Then, the variability along the first or the second main axis of the manipulability ellipsoid, corresponding to the first and second diagonal elements of the gain matrix $\bm{K}_{\bm{M}}$, is reduced. This means that the robot needs to prioritize the tracking of one of the ellipsoid main axes over the other. In the fourth test, the variability of the correlation between the two main axes of the manipulability ellipsoid is lowered. In this last case, the manipulability controller prioritizes the tracking of the ellipsoid orientation over the shape. 

Figure~\ref{Fig:TaskWithPrecision} shows how the manipulator posture is adapted to track the desired manipulability ellipsoid with a priority on the component with the lowest variability. Note that when high tracking precision is required for one of the main axes of the ellipsoid, the robot initially seeks to fit the shape of the ellipsoid along that specific axis, and subsequently it matches the whole manipulability ellipsoid. In this case, the precision ratio between the prioritized and the rest of components of the gain matrix is $10\!:\!1$. When high tracking precision is assigned to the correlation of the ellipsoid axes, the robot first tries to align its manipulability with the orientation of the desired ellipsoid, and afterwards the whole manipulability is matched. In this case, the precision ratio between the prioritized correlation and the other components of the gain matrix is $3\!:\!1$. Notice that the precision tensor naturally affects the computed joint velocities required to track a given ellipsoid, which consequently influences the resulting motion of the end-effector as a function of the precision constraints, as shown in Fig.~\ref{Fig:Precision_trajectories}. After convergence, the desired manipulability ellipsoid is successfully matched for all experiments. These results show that our geometry-aware tracking permits to take into account the variability information of a task to define the manipulability tracking precision. 

\begin{figure}[!tbp]
	\centering
	\begin{subfigure}[b]{0.24\textwidth}
		\includegraphics[width=\textwidth]{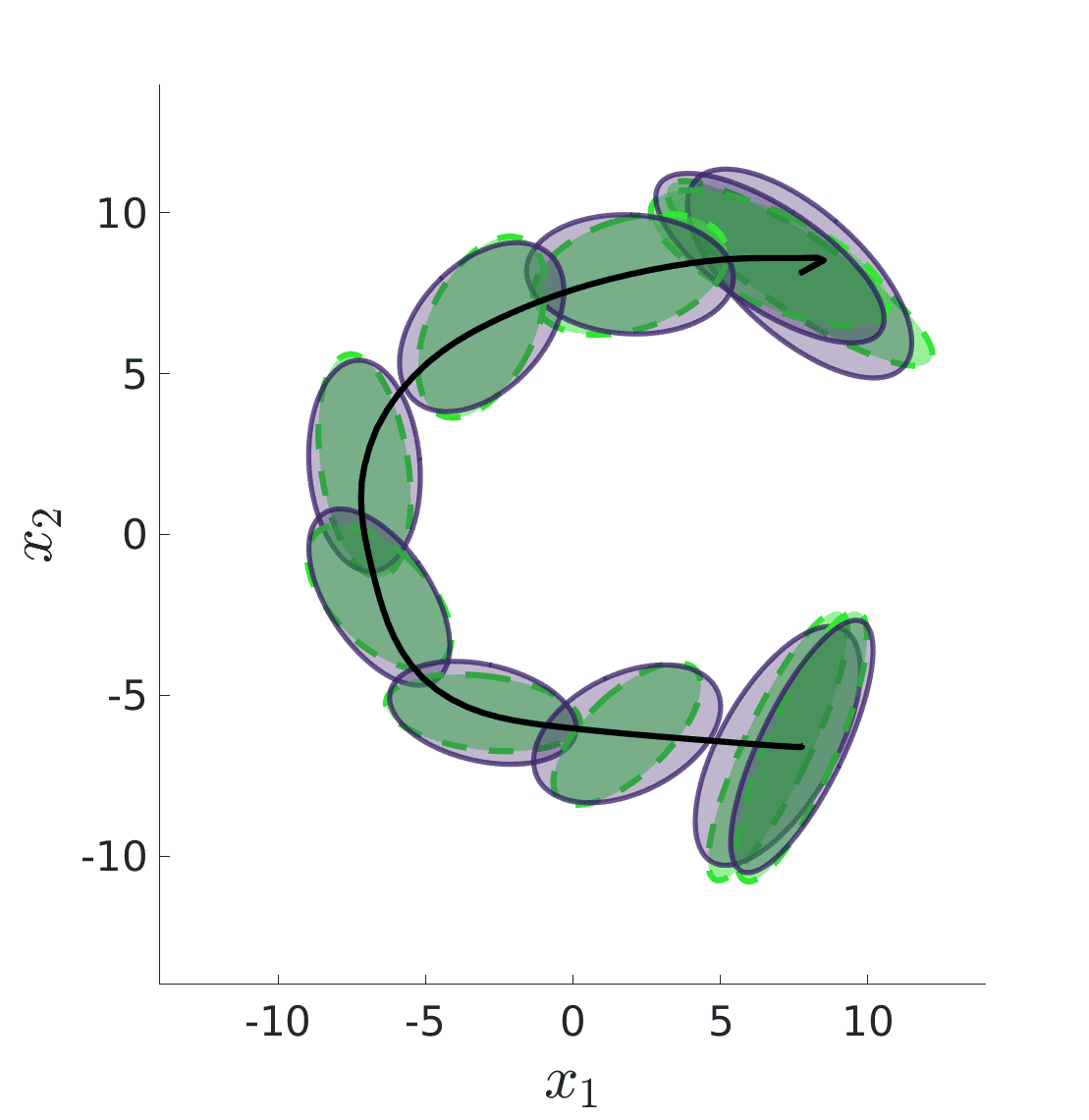}
		\caption{$\bm{K}_{\bm{M}}$ as a scalar}
		\label{subFig:ManipTranfer_Repro}
	\end{subfigure}	
	\begin{subfigure}[b]{0.24\textwidth}
		\includegraphics[width=\textwidth]{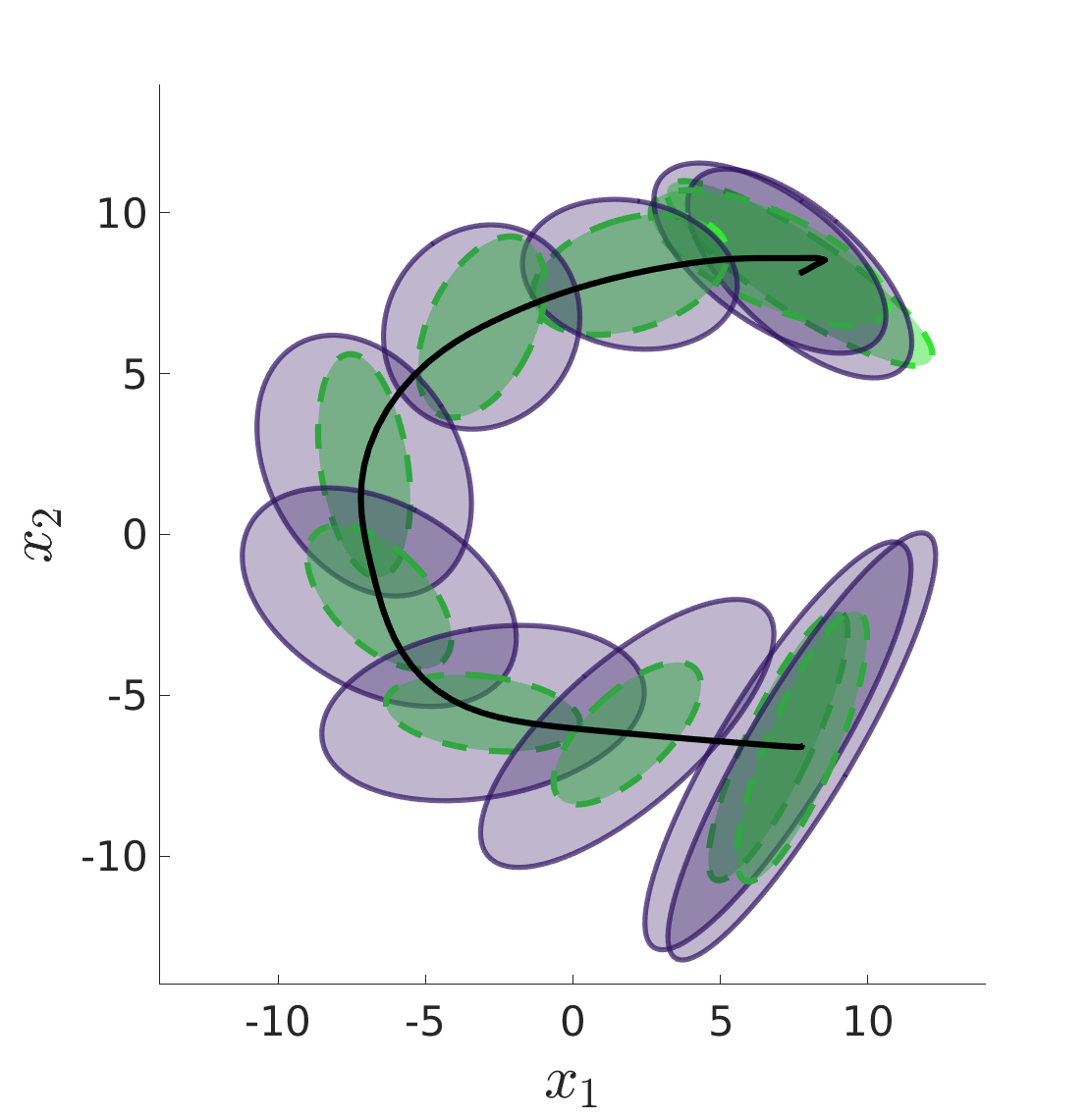}
		\caption{$\bm{K}_{\bm{M}}$ from GMR}
		\label{subFig:ManipTranfer_Repro_CovGain}
	\end{subfigure}
	\caption{Reproductions of a learned C-shape tracking task with desired manipulability ellipsoids. The end-effector trajectory is shown in black solid line, while the desired and reproduced manipulabilities are depicted in green and dark purple, respectively. (\emph{a}) $\bm{K}_{\bm{M}}$ is a scalar value, (\emph{b}) $\bm{K}_{\bm{M}}$ is the diagonal of the precision tensor retrieved by GMR. The required tracking precision is higher at the start and end of the task as a consequence of the low observed variability.}  
	\label{Fig:ManipTransferTracking}
\end{figure}

Therefore, our manipulability tracking approach may be readily combined with the manipulability learning framework introduced in Section~\ref{sec:Learning}.
In order to illustrate this, we show the reproduction phase of the experiment carried out in Section~\ref{subsec:GMR}. The 5-DoF \textit{student} robot was requested to track a desired Cartesian trajectory as main task, while varying its joint configuration for matching desired manipulability ellipsoids as secondary task. 
The student robot used the geometry-aware controller defined by~\eqref{Eq:ManipTrackSecTask}, where $\bm{K}_{\bm{M}}$ was defined either as a scalar value or as a diagonal matrix with the diagonal elements of ~\eqref{Eq:FullGainMat} with the precision tensor being equal to the inverse of the covariance tensor $\hat{\bm{\mathcal{S}}}_\ty{OO}^\ty{OO}$ retrieved by GMR~\eqref{Eq:SPD_GMRcov}. Our goal here was to exploit the learned variability information of the task to demand the robot a high precision tracking where low variability was observed in the demonstrations, and vice-versa. Successful reproductions of the demonstrated task using our manipulability-based redundancy resolution controller with scalar and variability-based matrix gains are shown in Figures~\ref{subFig:ManipTranfer_Repro} and~\ref{subFig:ManipTranfer_Repro_CovGain}, respectively. Note that the variability-based matrix gain changes the required tracking precision, where higher precision is enforced only at the beginning and the end of the task, which results in lower control efforts in between. These results validate that the proposed approach allows the robot to reproduce reference profiles of desired manipulability ellipsoids while adapting the tracking precision according to the demonstrated requirements of the task.

\subsection{Nullspace of the manipulability Jacobian}
\label{subsec:ManJacNullsp}
As traditionally done when designing redundancy resolution controllers, the nullspace of the manipulability Jacobian can also be exploited to fulfill secondary objectives when manipulability tracking is the main task. More specifically, a joint velocity $\bm{\dot{q}}_{N}$, aimed at fulfilling secondary objectives, can be projected into the nullspace of our manipulability tracking controller~\eqref{Eq:ManipTrackFirstTask} using the nullspace operator ${\Big( \bm{I}- (\bm{\mathcal{J}}_{(3)}^\dagger)^\trsp \bm{\mathcal{J}}_{(3)}^\trsp \Big)}$. Therefore, the resulting redundancy resolution controller is given by
\begin{multline}
\bm{\dot{q}}_t\!=\!(\bm{\mathcal{J}}_{(3)}^\dagger)^\trsp \, \bm{K}_{\bm{M}} \, \text{vec}\Big(\!\text{Log}_{\bm{M}_t}(\bm{\hat{M}}_t)\!\Big) \\
+ \Big(\! \bm{I}- (\bm{\mathcal{J}}_{(3)}^\dagger)^\trsp \bm{\mathcal{J}}_{(3)}^\trsp\! \Big) \bm{\dot{q}}_{N}.
\label{Eq:ManipTrackNullspace}
\end{multline}

\begin{figure}[!tbp]
	\centering
	\begin{subfigure}[b]{0.29\textwidth}
		\includegraphics[width=\textwidth]{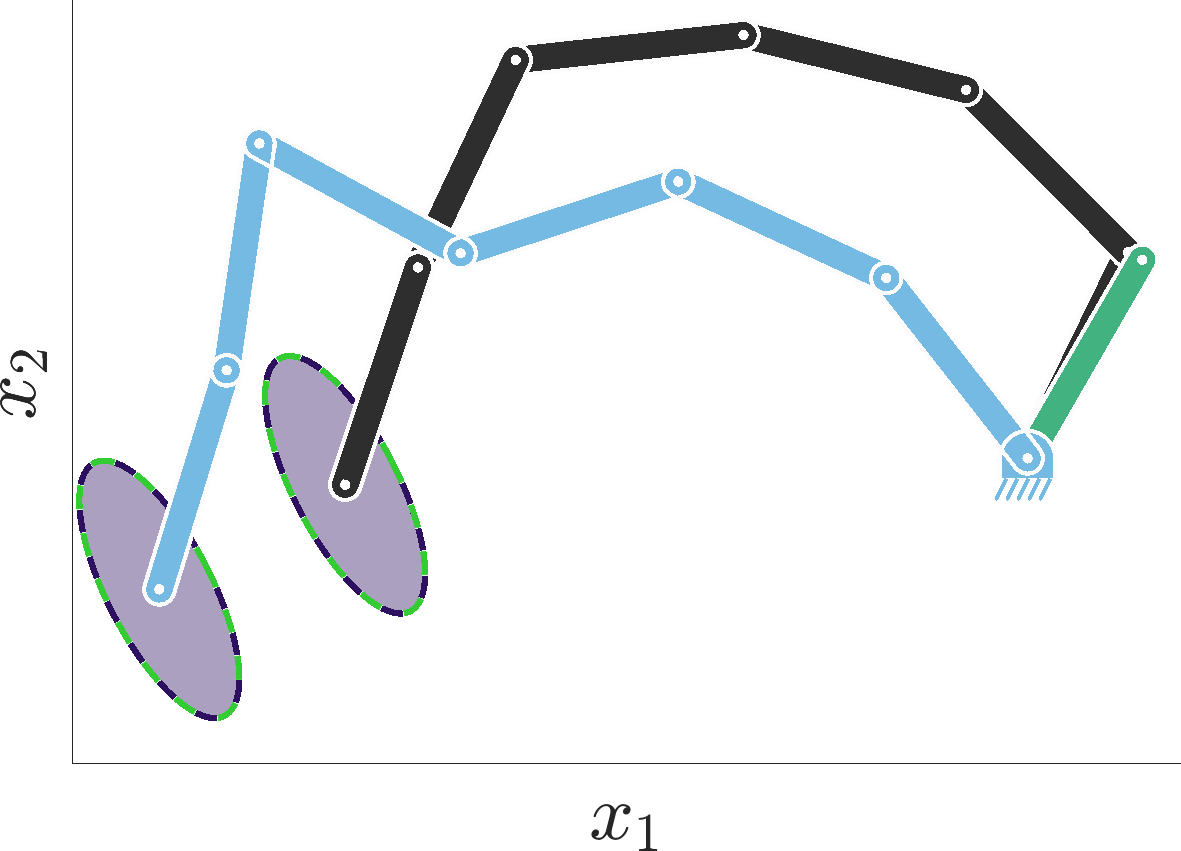}
	\end{subfigure}	
	\begin{subfigure}[b]{0.11\textwidth}
		\includegraphics[width=\textwidth]{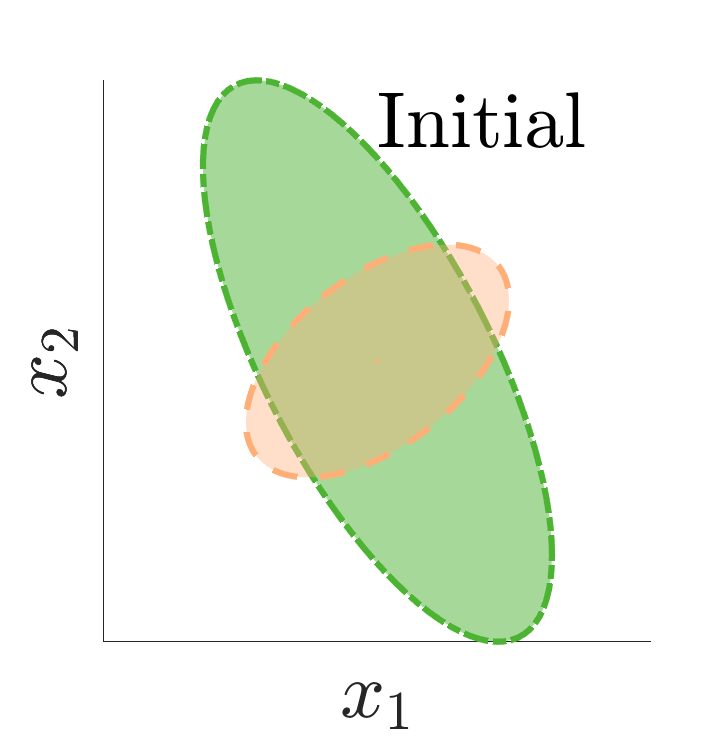}
		\includegraphics[width=\textwidth]{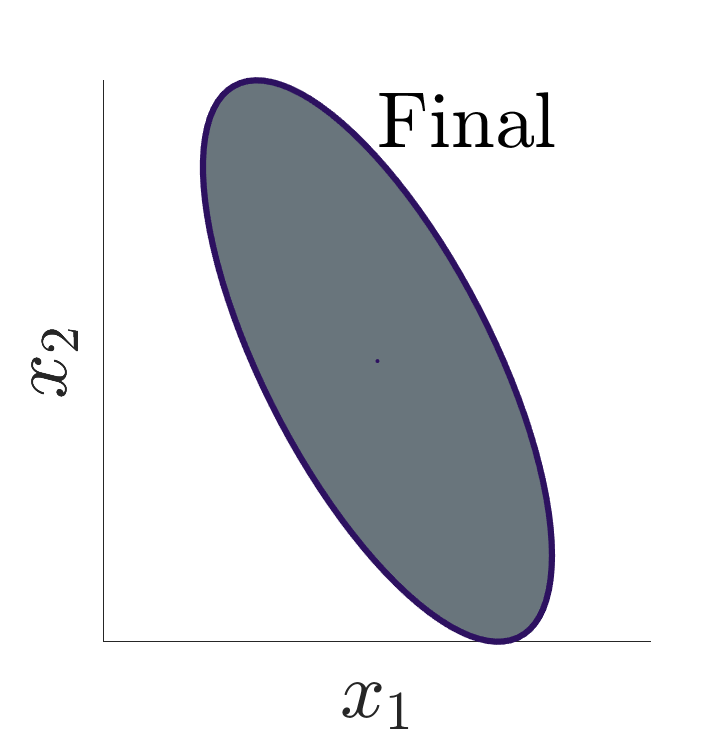}
	\end{subfigure}	
	\caption{Use of the nullspace of the manipulability Jacobian. Two 6-DoF planar robots are required to track a desired manipulability ellipsoid as main task. The black robot also keeps its first joint at a fixed position (depicted by the green link), which is a secondary objective projected into the nullspace of the manipulability Jacobian. The final manipulability ellipsoids (in purple) fully overlap the desired ones (in green), showing a precise manipulability tracking. The initial manipulability ellipsoid is depicted in yellow.} 
	\label{Fig:NullspaceManip}
\end{figure}

In order to show the functionality of this nullspace operator, we carried out experiments with a simulated 6-DoF planar robot. The main task of the robot is to track a desired manipulability ellipsoid while keeping a desired pose for its first joint $q_0$, which is considered as secondary task. Thus, the nullspace velocity is defined as a simple proportional controller $\bm{\dot{q}}_{N} = \bm{K}^\mathcal{P}_{\bm{q}} (\bm{\hat{q}} - \bm{q}_t)$
where $\bm{\hat{q}}$ is the desired joint configuration and $\bm{K}^\mathcal{P}_{\bm{q}}$ is a matrix gain defined so that only joint position errors in the first joint are compensated. Figure~\ref{Fig:NullspaceManip} shows that the black manipulator configuration is adjusted to track the desired manipulability ellipsoid and keep, as accurately as possible, the desired joint position for $q_0$. Note that the black robot is able to find an alternative joint configuration that permits not only to closely track the desired manipulability, but also fulfill secondary objectives projected into its nullspace, in contrast to the blue robot which exclusively implements a manipulability tracking task. These results show that the nullspace of the manipulability Jacobian is suitable to carry out a secondary task along with manipulability tracking as main objective.

\section{Importance of geometry-awareness}  
\label{sec:GeomImportance}
In the previous sections we introduced a geometry-aware manipulability transfer framework composed of \emph{(1)} a probabilistic model that encodes and retrieves manipulability ellipsoids, and \emph{(2)} manipulability tracking controllers. In this section, we show that the geometry-awareness of our formulations is crucial for successfully learning and tracking manipulability ellipsoids in addition to providing an appropriate mathematical treatment of both problems.

\subsection{Learning}
We first evaluate the proposed learning formulation compared to a framework that ignores that manipulability ellipsoids belong to the SPD manifold. To do so, we encode a distribution of manipulability ellipsoids with a GMM acting in the Euclidean space and we then retrieve desired manipulability ellipsoids via the corresponding GMR. To ensure the validity of the desired manipulability ellipsoids, GMM and GMR are performed on lower triangular matrices $\bm{L}$ obtained via Cholesky decomposition. Thus, the positive-definiteness of the desired manipulability ellipsoids computed as $\bm{\hat{M}} = \bm{\hat{L}}\bm{\hat{L}}^\trsp$ is guaranteed, where $\bm{\hat{L}}$ is the estimated GMR output. Note that this property is not guarantee in the case where GMM and GMR acting in the Euclidean space is applied directly to the manipulability ellipsoids $\bm{M}$. Therefore, we do not consider this approach in the comparison as the desired matrices $\bm{\hat{M}}$ may not be manipulability ellipsoids in some cases.

\begin{figure}[!tbp]
	\centering
	\begin{subfigure}[b]{0.22\textwidth}
		\includegraphics[width=\textwidth]{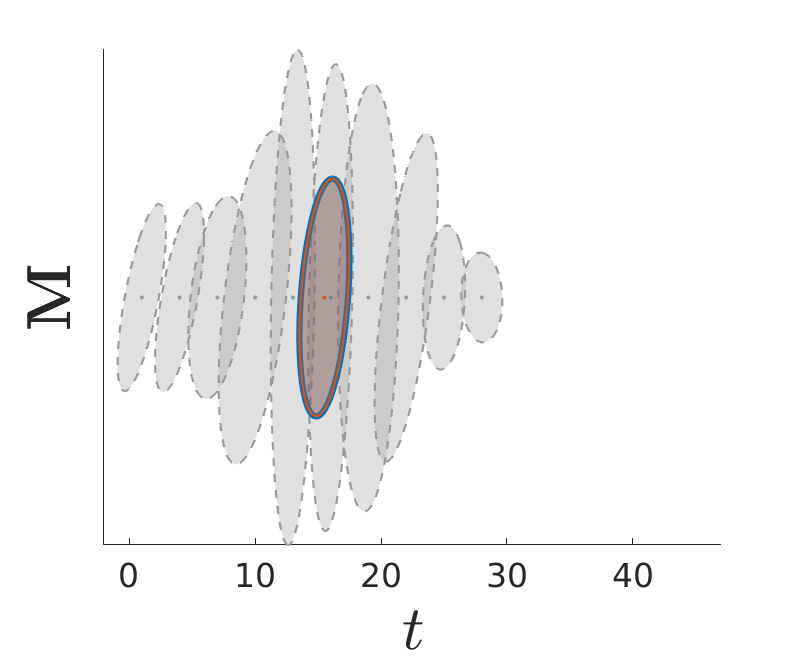}
		\caption{Data and GMM}
		\label{subFig:LearningComp_demos}
	\end{subfigure}
	\begin{subfigure}[b]{0.21\textwidth}
		\includegraphics[width=\textwidth]{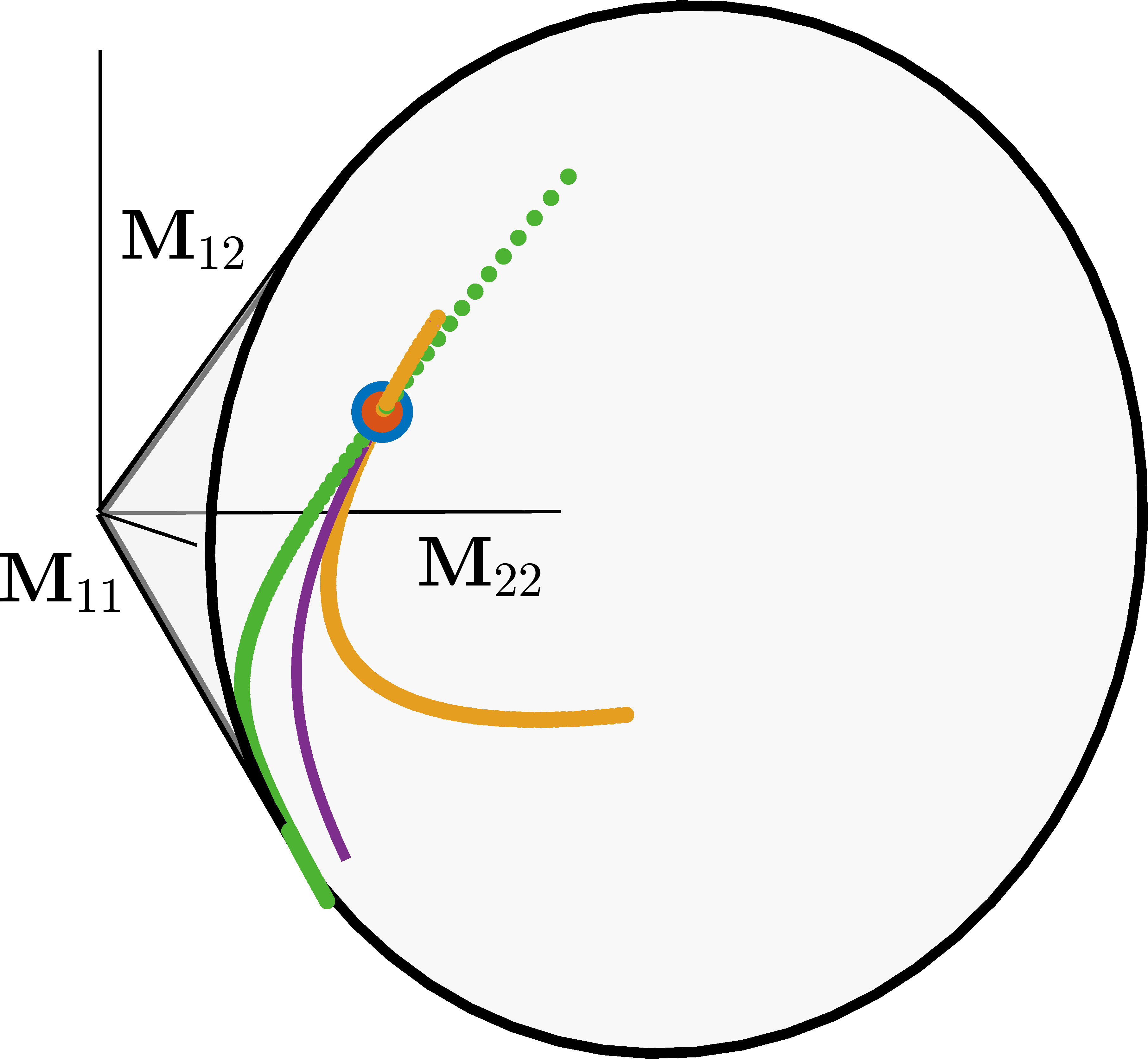}
		\caption{GMR profiles in $\mathcal{S}_{\ty{++}}^2$}
		\label{subFig:LearningComp_SPD}
	\end{subfigure}
	\begin{subfigure}[b]{0.49\textwidth}
		\includegraphics[width=\textwidth]{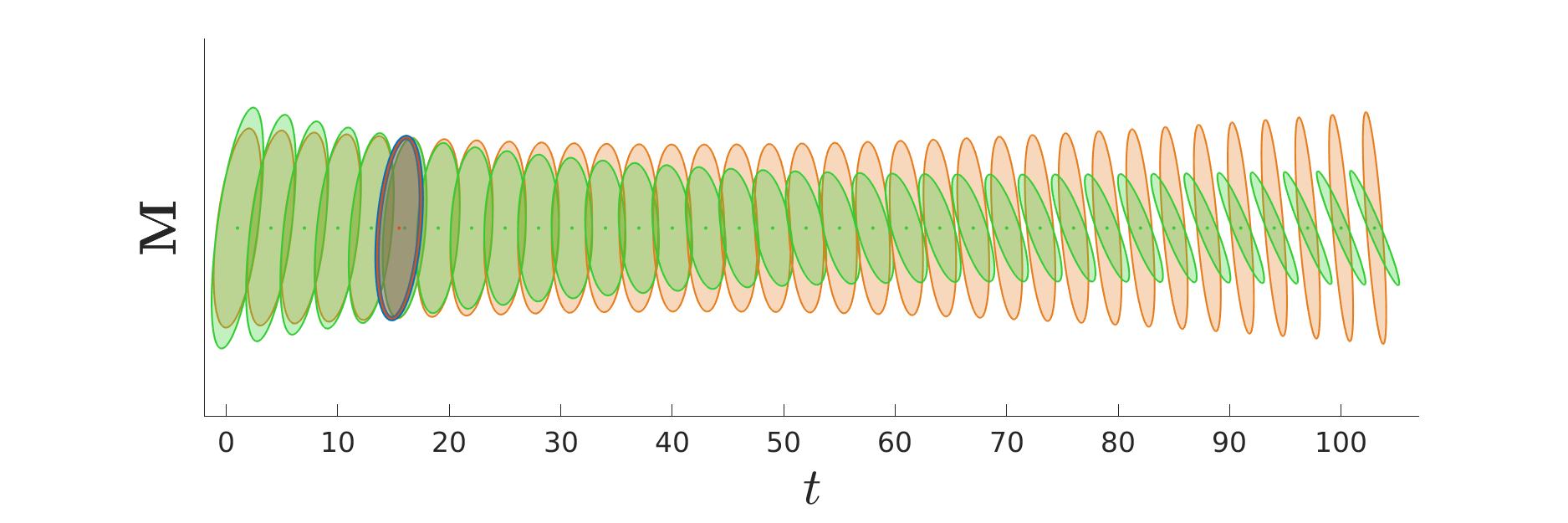}
		\caption{GMM and retrieved GMR profiles}
		\label{subFig:LearningComp_repros}
	\end{subfigure}
	\caption{Importance of geometry in manipulability learning formulations. (\emph{a}) Demonstrated data (depicted in light gray), and mean of the GMM component for the geometry-aware and Euclidean approaches (overlapping blue and red ellipsoids, respectively). (\emph{c}) Manipulability profiles retrieved by the geometry-aware and Euclidean GMR, shown as green and orange ellipses, respectively. The time axis is shared with (\emph{a}). (\emph{b}) Mean of the GMM component and estimated profiles in the cone of SPD matrices. The manipulability profile obtained by our approach, shown in green, follows a geodesic. The profile obtained by the Euclidean framework is depicted by the orange curve and does not follow a geodesic on the manifold.
	The geodesic containing the mean of the Euclidean GMM, being a geometrically valid trajectory (depicted in purple), does not correspond to the trajectory obtained with the Euclidean framework. Thus, the Euclidean approach is geometrically flawed.}
	\label{Fig:LearningComp}
\end{figure}

Figure~\ref{Fig:LearningComp} compares the proposed approach (Section~\ref{sec:Learning}) and the manipulability learning using GMM/GMR acting in Euclidean space. The demonstration consists of a time series of changing manipulability ellipsoids. For each approach, a 1-state GMM is trained and a reproduction is carried out for a longer time period than the demonstration using GMR. 
Both geometry-aware and Euclidean approaches obtain similar means of the GMM component (see Fig.~\ref{subFig:LearningComp_demos},~\ref{subFig:LearningComp_SPD}). This is due to the fact that the Euclidean mean computed using the Cholesky decomposition is a good approximation of the mean computed on $\mathcal{S}_{\ty{++}}^D$ if the SPD data are close enough to each other. However, the covariances of the GMM components of both approaches are not equivalent. Indeed, the covariance of our geometry-aware approach is computed using the SPD data projected in the tangent space of the mean, while that of the Euclidean GMM corresponds to the covariance of the elements of the vectorized Cholesky decomposition, which ignores the geometry of the SPD manifold.

The manipulability ellipsoids profiles retrieved by the geometry-aware and Euclidean GMR are similar around the mean of the GMM component, but diverge when moving away from it (see Fig.~\ref{subFig:LearningComp_repros}). This is because the estimated output in Euclidean space is only a valid approximation for input data lying close to the mean. In contrast, our approach is able to extrapolate the rotating behavior of the demonstrated manipulability ellipsoids as
the recovered trajectory follows a geodesic on the SPD manifold (see Fig.~\ref{subFig:LearningComp_SPD}). Note that this is the equivalent to following a straight line in Euclidean space, which is the expected result of a trajectory computed via Gaussian conditioning. This behavior is obtained by parallel transporting the GMM covariances to the tangent space of the mean of the estimated conditional distribution of GMR \eqref{Eq:SPD_GMRpt}. Therefore, the Euclidean GMR does not recover a trajectory following a geodesic on the manifold, leading to inconsistent extrapolated manipulability ellipsoids. 

The reported results show that our geometry-aware approach accurately reproduces the behavior of the demonstrated data, and therefore provides a mathematically sound method for learning and retrieving manipulability ellipsoids in the SPD manifold. Note that similar behaviors are observed for GMM with any number of states, the number $K=1$ was chosen here to facilitate the visualization of the results.

\subsection{Tracking}
\subsubsection{Comparisons with Euclidean tracking}
After showing the importance of geometry for learning manipulability ellipsoids, we compare the proposed tracking formulation against a controller ignoring the geometry of SPD matrices (i.e., treating the problem as Euclidean). Moreover, we evaluate our controller when the tracking of manipulability ellipsoids is assigned a secondary role. This evaluation compares our formulation against three Euclidean controllers, and the gradient-based approach in~\citep{Rozo17IROS:ManTransfer}.
For the case in which the manipulability tracking is the main objective, we consider a 4-DoF planar robot that is required to track a desired manipulability ellipsoid by minimizing the error between its current and desired manipulability ellipsoids $\bm{M}$ and $\bm{\hat{M}}$. We first compare the proposed approach \eqref{Eq:ManipTrackFirstTask} with the following Euclidean manipulability tracking controller
\begin{equation}
\bm{\dot{q}}_t = (\bm{\mathcal{J}}_{(3)}^\dagger)^\trsp \bm{K}_{\bm{M}} \text{vec}(\bm{\hat{M}}_t-\bm{M}_t) ,
\label{Eq:EuclFormulation}
\end{equation} 
where the difference between two manipulability ellipsoids is computed in Euclidean space, i.e., ignoring that manipulability ellipsoids belong to the set of SPD matrices. Secondly, we compare the proposed approach to the Cholesky-based Euclidean manipulability controller
\begin{equation}
\bm{\dot{q}}_t = (\bm{\mathcal{J}}_{(3)}^\dagger)^\trsp \bm{K}_{\bm{M}} \text{vec}(\bm{\Delta L}_t\bm{\Delta L}_t^\trsp) ,
\label{Eq:CholEuclFormulation}
\end{equation}
where $\bm{\Delta L} = \bm{\hat{L}}- \bm{L}$ and matrices $\bm{L}$ are obtained from the Cholesky decomposition such that $\bm{M} = \bm{L}\bm{L}^\trsp$. This controller ensures that the difference between two manipulability ellipsoids is positive definite, but ignores that they belong to the SPD manifold.
For completeness, we also compare our approach with the Cholesky-Jacobian-based Euclidean manipulability controller
\begin{equation}
\bm{\dot{q}}_t = (\mathcal{J}_{\text{chol}(3)}^\dagger)^\trsp \bm{K}_{\bm{M}}\; \text{vec}(\bm{\hat{L}}- \bm{L}),
\label{Eq:CholJacobianEuclFormulation}
\end{equation}
where $\mathcal{J}_{\text{chol}} = \frac{\delta \bm{L}}{\delta \bm{q}} = \frac{\delta \bm{L}}{\delta \bm{M}} \bm{\mathcal{J}}$ is the Cholesky-based manipulability Jacobian, so that $\bm{\dot{L}}=\mathcal{J}_{\text{chol}} \times_3 \bm{\dot{q}}^\trsp$. This approach tracks a desired manipulability solely through its Cholesky decomposition with an adapted manipulability Jacobian. Similarly to~\eqref{Eq:CholEuclFormulation}, it ensures the positive-definiteness of manipulability ellipsoids, but ignores that they belong to the SPD manifold.
For all the following comparisons, the gain matrices $\bm{K}_{\bm{M}}$ are identity matrices.

\begin{figure*}[!tbp]
	\centering
	\includegraphics[width=.9\textwidth]{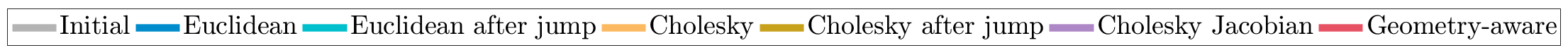}
	\begin{subfigure}[b]{0.9\textwidth}
		\includegraphics[width=.33\textwidth]{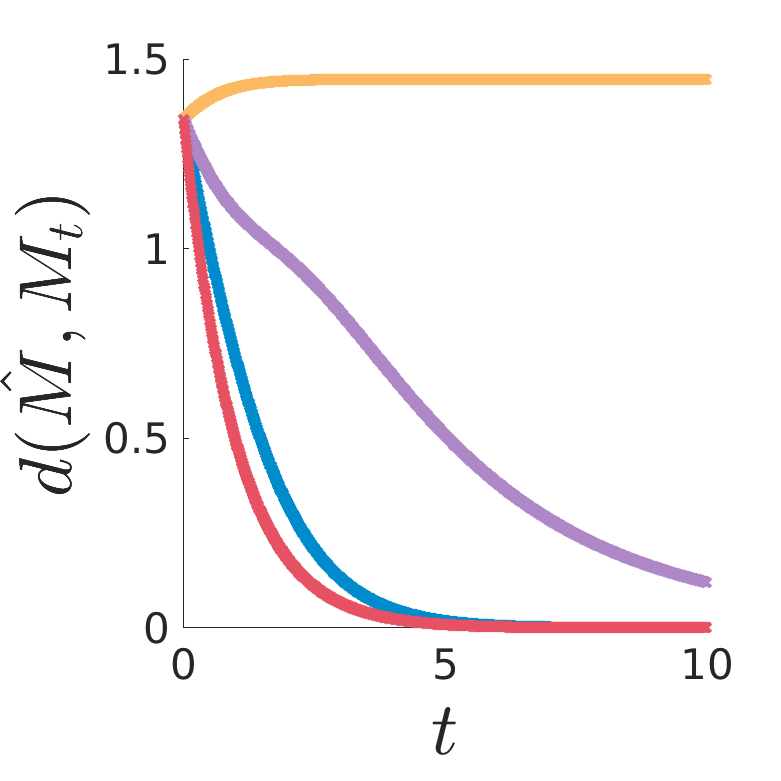}
		\includegraphics[width=.33\textwidth]{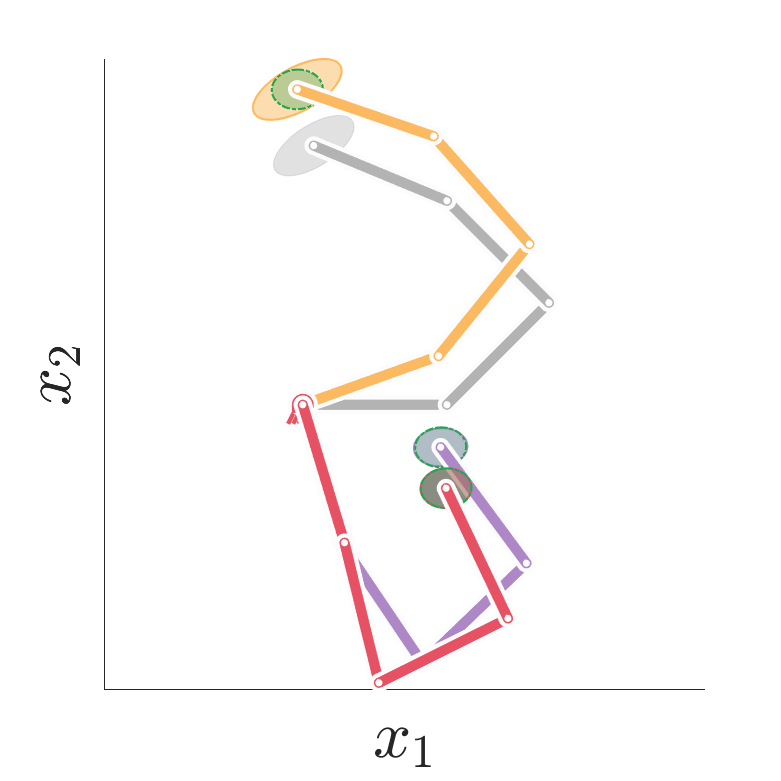}
		\includegraphics[width=.3\textwidth]{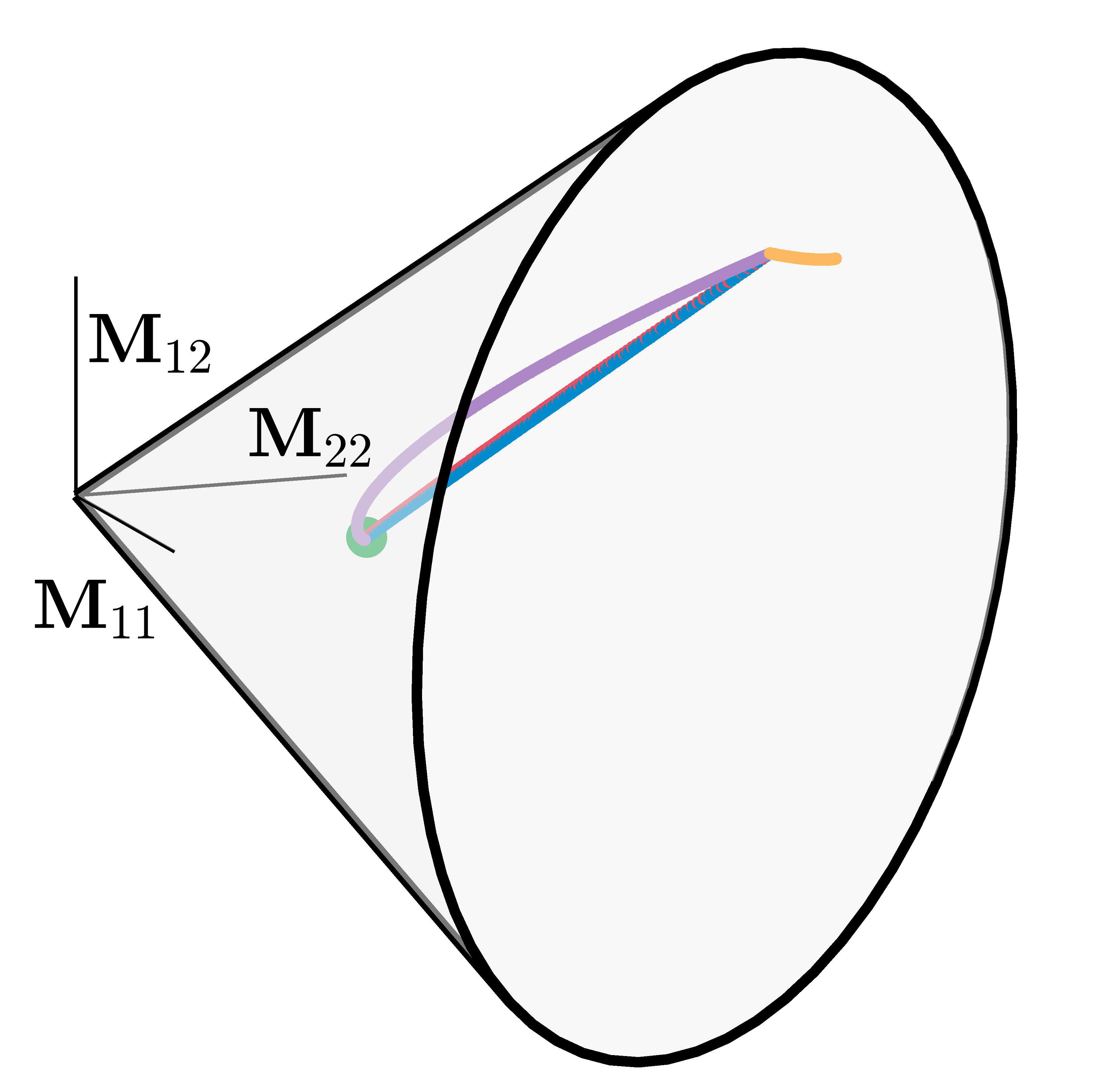}
		\caption{}
		\label{subFig:TrackingComp01}
	\end{subfigure}	
	\begin{subfigure}[b]{0.9\textwidth}
		\includegraphics[width=.33\textwidth]{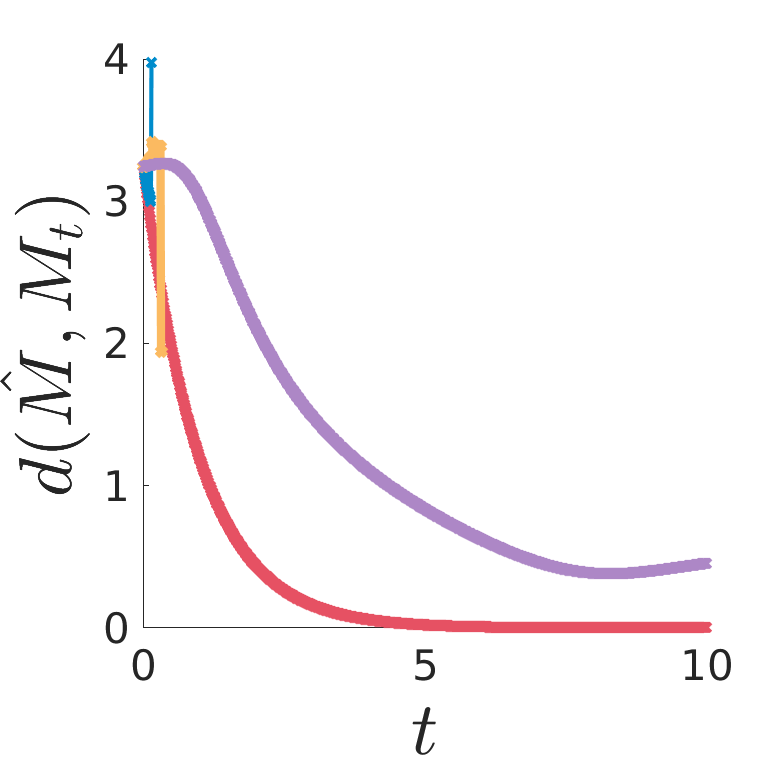}
		\includegraphics[width=.33\textwidth]{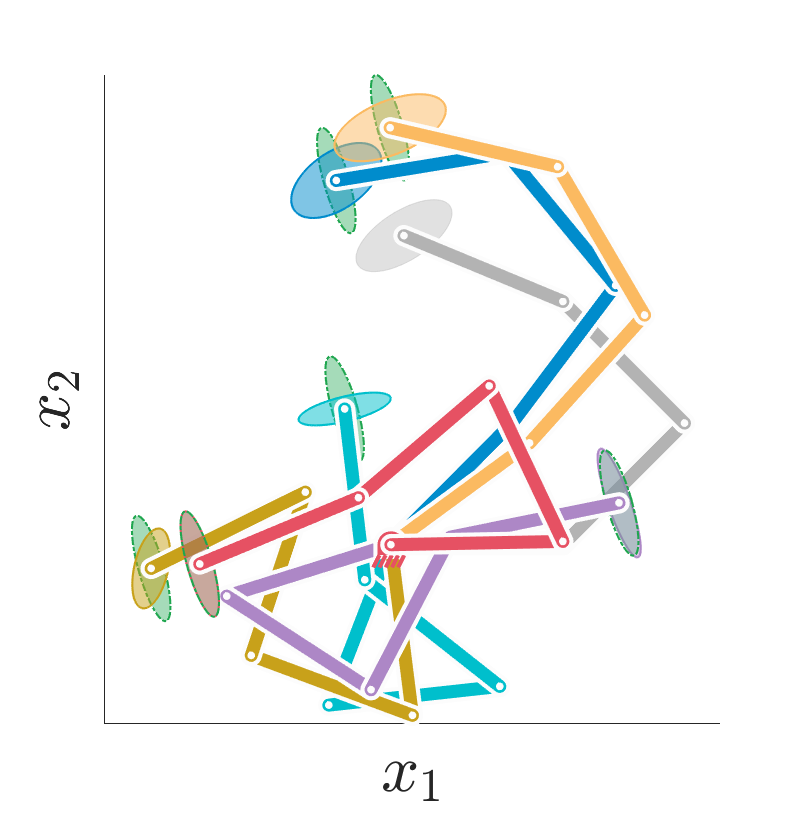}
		\includegraphics[width=.3\textwidth]{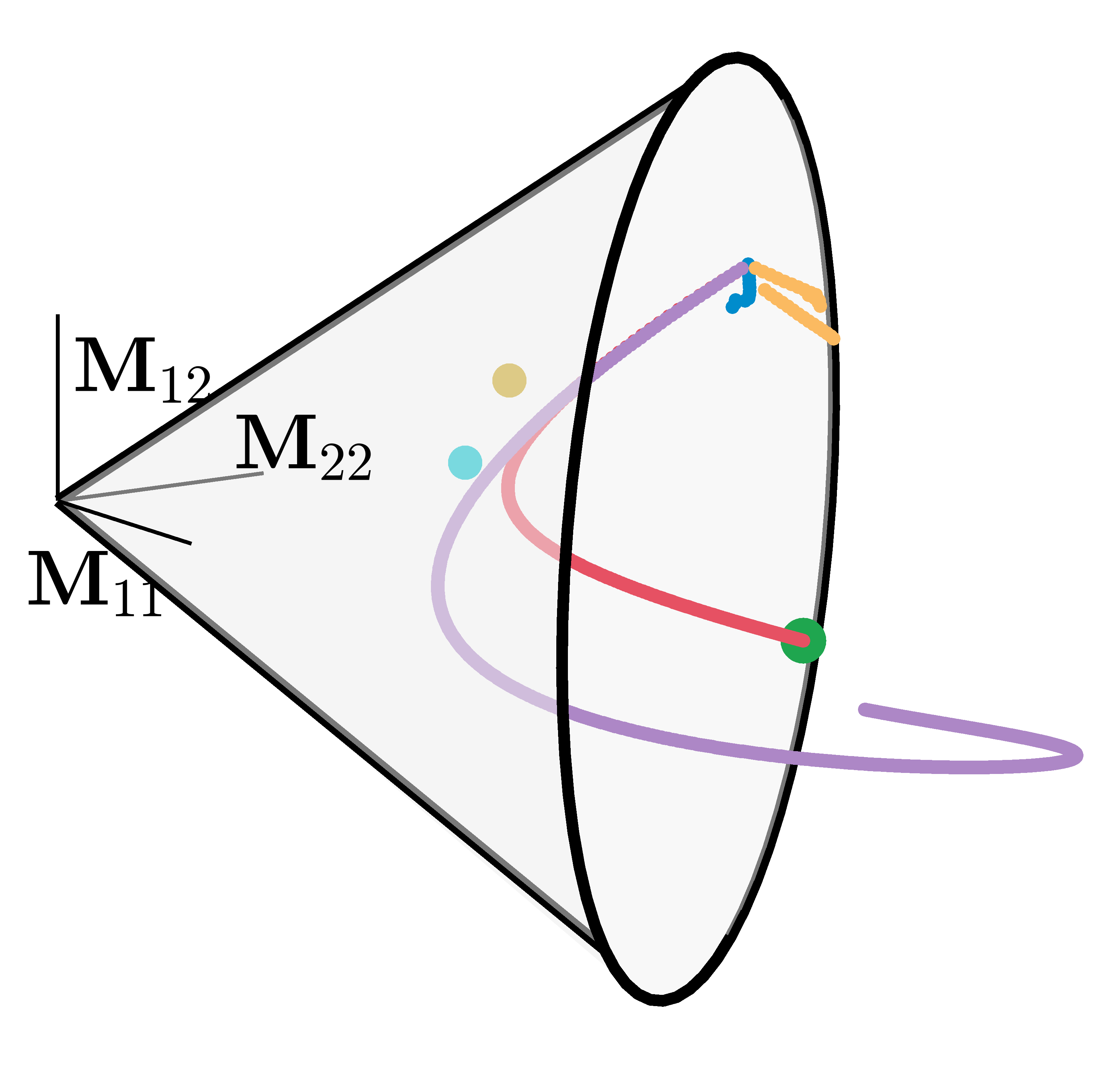}
		\caption{}
		\label{subFig:TrackingComp02}
	\end{subfigure}
	\caption{Performance of different manipulability tracking formulations. The \emph{left} graphs show the affine-invariant distance between the current and desired manipulability ellipsoids over time. The distances for the Euclidean, Cholesky-based Euclidean, Cholesky-Jacobian-based Euclidean and geometry-aware approaches are respectively depicted in blue, yellow, lila and red. The \emph{middle} graphs display the initial and final robot postures and the final manipulability ellipsoids. The initial posture is depicted in light gray, while the final posture and corresponding manipulability for the three methods are depicted in the same color as the distances. The desired manipulability is depicted in green. \emph{Middle-(b)} also shows the sudden change in the robot posture for both Euclidean methods \eqref{Eq:EuclFormulation} and \eqref{Eq:CholEuclFormulation}. The robot posture before and after the abrupt change is shown in blue and light blue, respectively for \eqref{Eq:EuclFormulation} and in yellow and olive, respectively for \eqref{Eq:CholEuclFormulation}. The \emph{right} graphs depict the evolution of the manipulability ellipsoids in the SPD manifold. The colors correspond to those of the previous graphs with the green dot representing the desired manipulability. The isolated light blue and olive dots in the \emph{bottom-(b)} graph represent the manipulability ellipsoids after the abrupt changes in the robot joint configuration.}
	\label{Fig:TrackingComp}
\end{figure*}
\begin{table*}[t]
	\footnotesize
	\renewcommand*{\arraystretch}{1.2}
	\caption{Final distances $d(\bm{\hat{M}},\bm{M}_t)$ between the current and desired manipulability ellipsoids for the performance comparison of the different manipulability tracking formulations.}
	\label{Tab:TrackingComp}
	\begin{center}
		\setlength\tabcolsep{2.5pt}
		\begin{tabular}{c|cc|cc|c|c|}
			\textbf{Approach} & Euclidean& (after jump) & Cholesky&(after jump)&Cholesky Jacobian &Geometry-aware\\
			\hline
			\textbf{Fig.~\ref{subFig:TrackingComp01}} & $1.3e^{-4}$ & - & $1.446$ & - & $0.1204$ & $6e^{-5}$\\ 
			\hline
			\textbf{Fig.~\ref{subFig:TrackingComp02}} & $2.997$ & $3.977$ & $3.385$ & $1.944$ & $0.455$ & $1.4e^{-4}$\\ 
			\hline
		\end{tabular}
	\end{center}
	\normalsize
\end{table*}

Figure~\ref{Fig:TrackingComp} shows the convergence rate for the proposed geometry-aware controller, the Euclidean-based approach, the Cholesky-based Euclidean and Cholesky-Jacobian-based Euclidean formulations. Two tests were carried out by varying the initial configuration of the robot and the desired manipulability ellipsoid. In the first case, the Euclidean and geometry-aware formulations converge to similar robot joint configurations with a distance between the current and desired manipulability close to zero (see Fig.~\ref{subFig:TrackingComp01}-\emph{left}, \emph{middle} and Table~\ref{Tab:TrackingComp}). However, in the second test, the Euclidean formulation induces a sudden change in the joint configuration, resulting in an abrupt increase on the error measured between the current and desired manipulability ellipsoids (see Fig.~\ref{subFig:TrackingComp02}-\emph{left}, \emph{middle}). In real scenarios, such unstable robot behavior would certainly be harmful and unsafe. This erroneous tracking performance can be explained by the fact that the Euclidean path between two SPD matrices is a valid approximation of the geodesic only if these are close enough to each other, as shown in Fig.~\ref{subFig:TrackingComp01}-\emph{right}. When this approximation is not valid (see Fig.~\ref{subFig:TrackingComp02}-\emph{right}), the Euclidean controller outputs inconsistent reference joint velocities that destabilize the robotic system, therefore failing to track the desired manipulability. Note that the Cholesky-based Euclidean formulation does not converge in both cases (see Table~\ref{Tab:TrackingComp}) and induces a sudden change in joint configuration of the robot in the second scenario, similarly to the Euclidean formulation. This can be explained by the fact that the path induced by this method is not close to geodesics on the SPD manifold as shown by Fig.~\ref{Fig:TrackingComp}-\emph{right}. As opposed to the two Euclidean formulations, the Cholesky-Jacobian-based Euclidean controller does not induce unstable robot behaviors and converges towards the desired manipulability ellipsoid for both cases. However, this method shows a poor convergence rate compared to our geometry-aware approach, as shown by Fig.~\ref{Fig:TrackingComp}-\emph{left}. This can be explained by the fact that, although this approach generates curved paths on the SPD manifold, these paths do not resemble geodesics and tend to induce detours to reach the desired manipulability ellipsoid (see Fig.~\ref{Fig:TrackingComp}-\emph{left}). This is particularly visible for the second test, where the resulting joint configuration is farther from the initial pose of the robot compared to the joint configuration obtained by the proposed geometry-aware controller (see Fig.~\ref{subFig:TrackingComp02}-\emph{middle}, \emph{right}).

\begin{figure*}[!tbp]
	\centering
	\begin{subfigure}[b]{0.24\textwidth}
		\includegraphics[width=.8\textwidth]{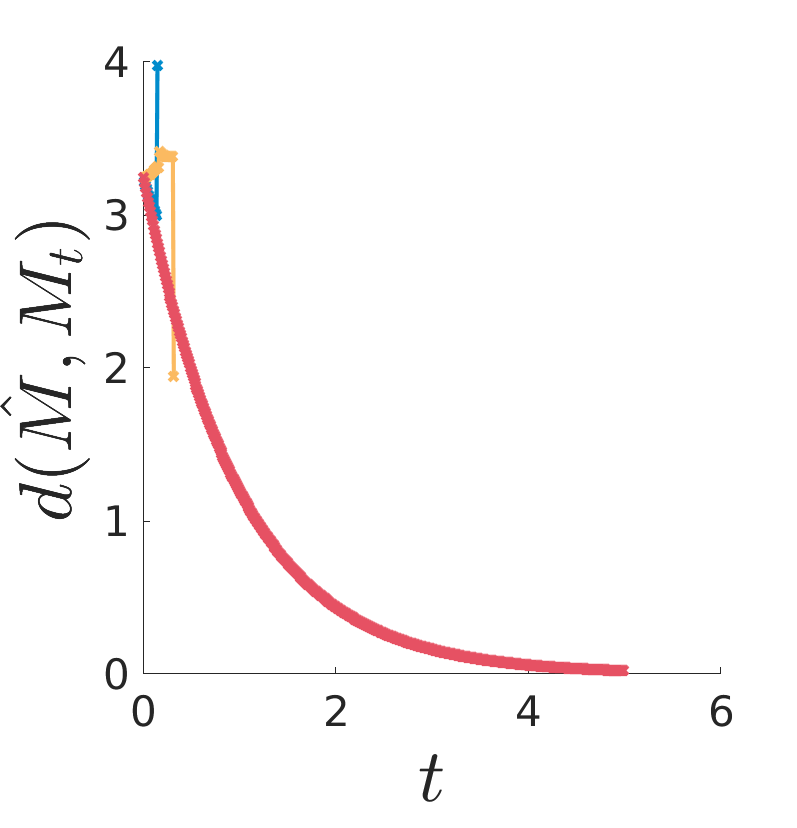}
		\includegraphics[width=.8\textwidth]{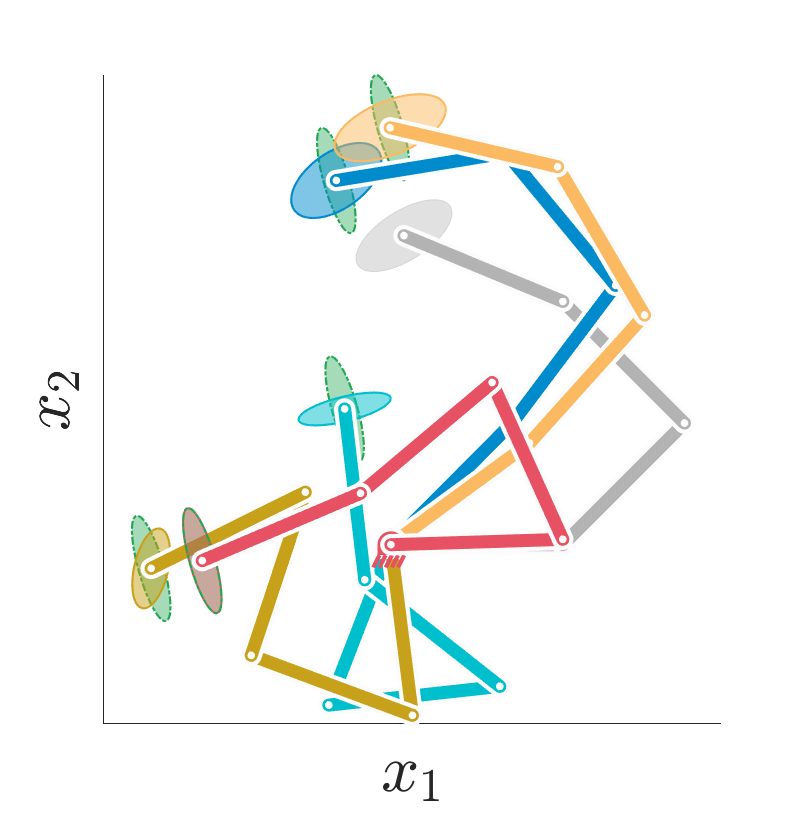}
		\includegraphics[width=.75\textwidth]{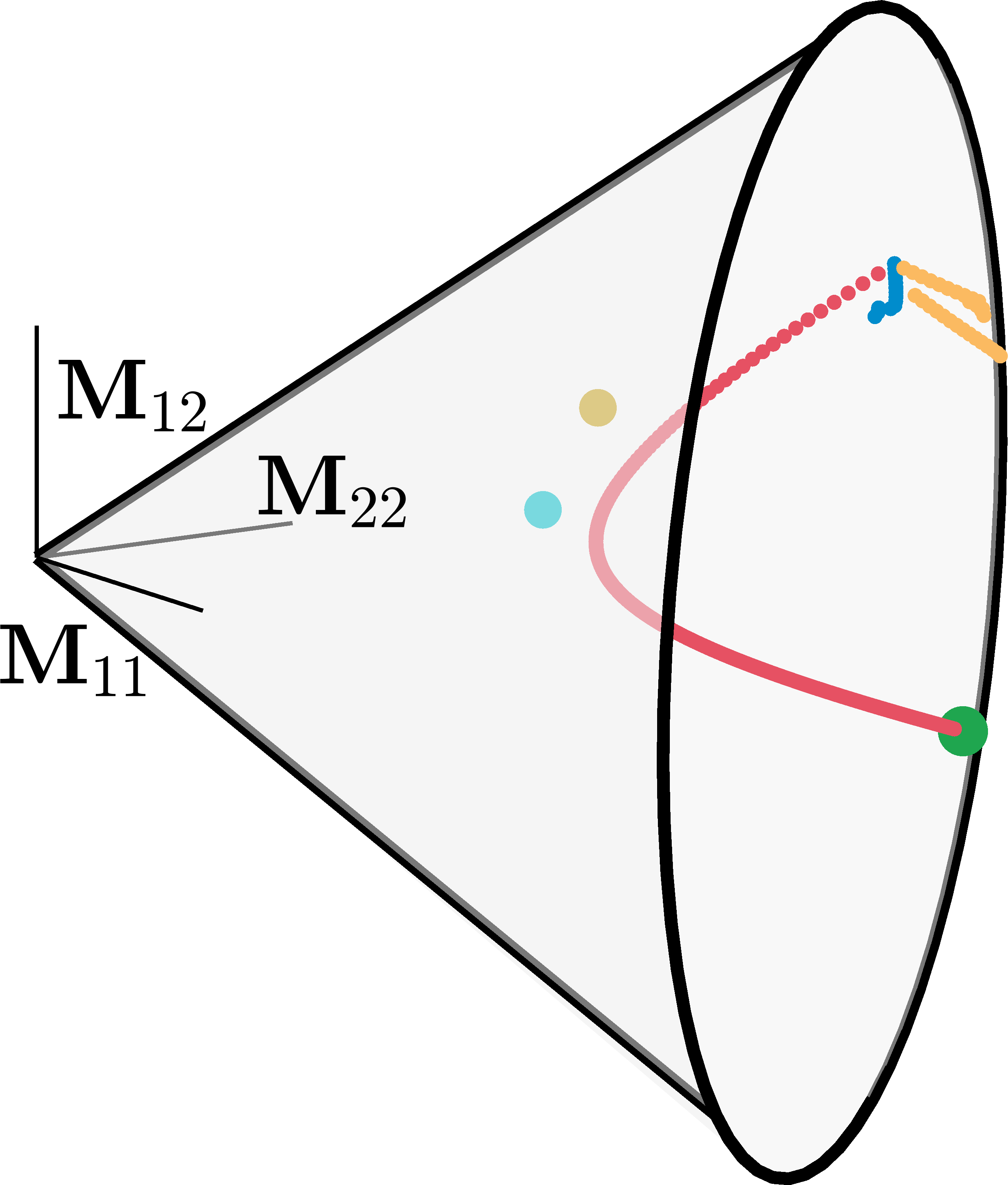}
		\caption{$\bm{K}_{\bm{M}}=\bm{I}$}
		\label{subFig:TrackingComp02_Km1}
	\end{subfigure}	
	\begin{subfigure}[b]{0.24\textwidth}
		\includegraphics[width=.8\textwidth]{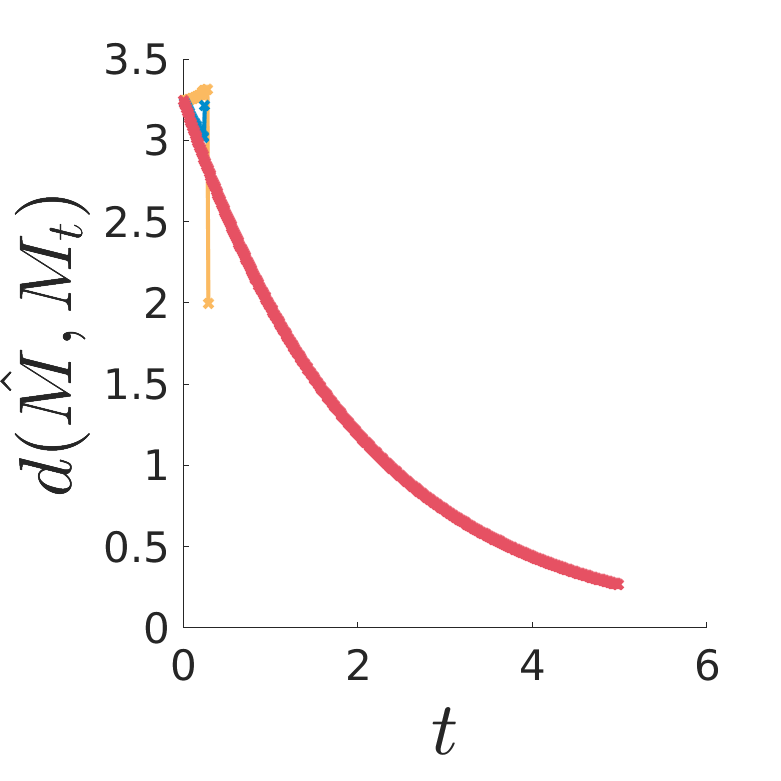}
		\includegraphics[width=.8\textwidth]{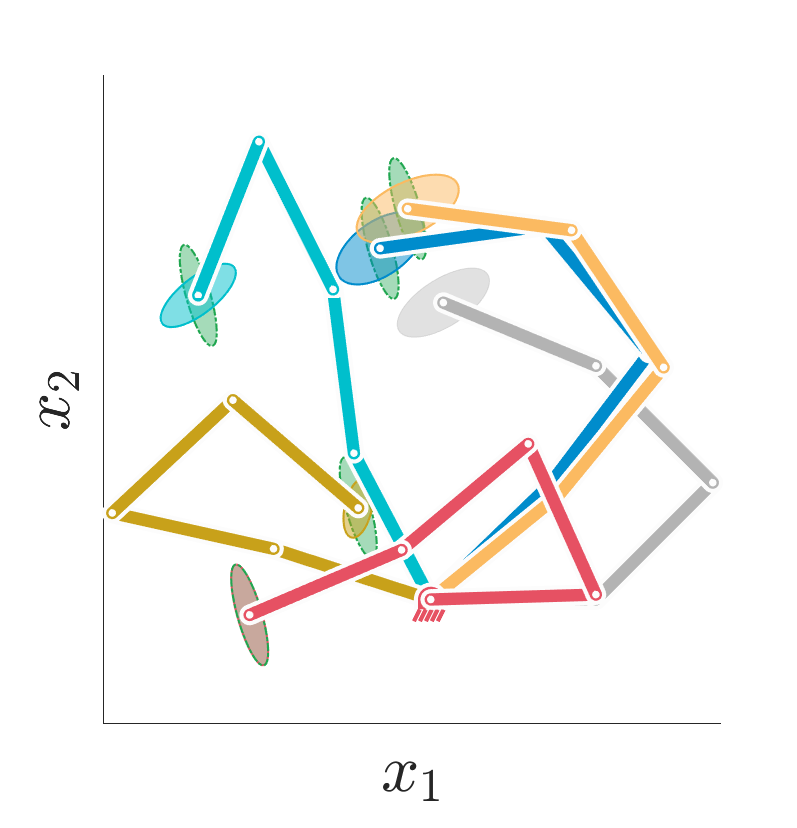}
		\includegraphics[width=.75\textwidth]{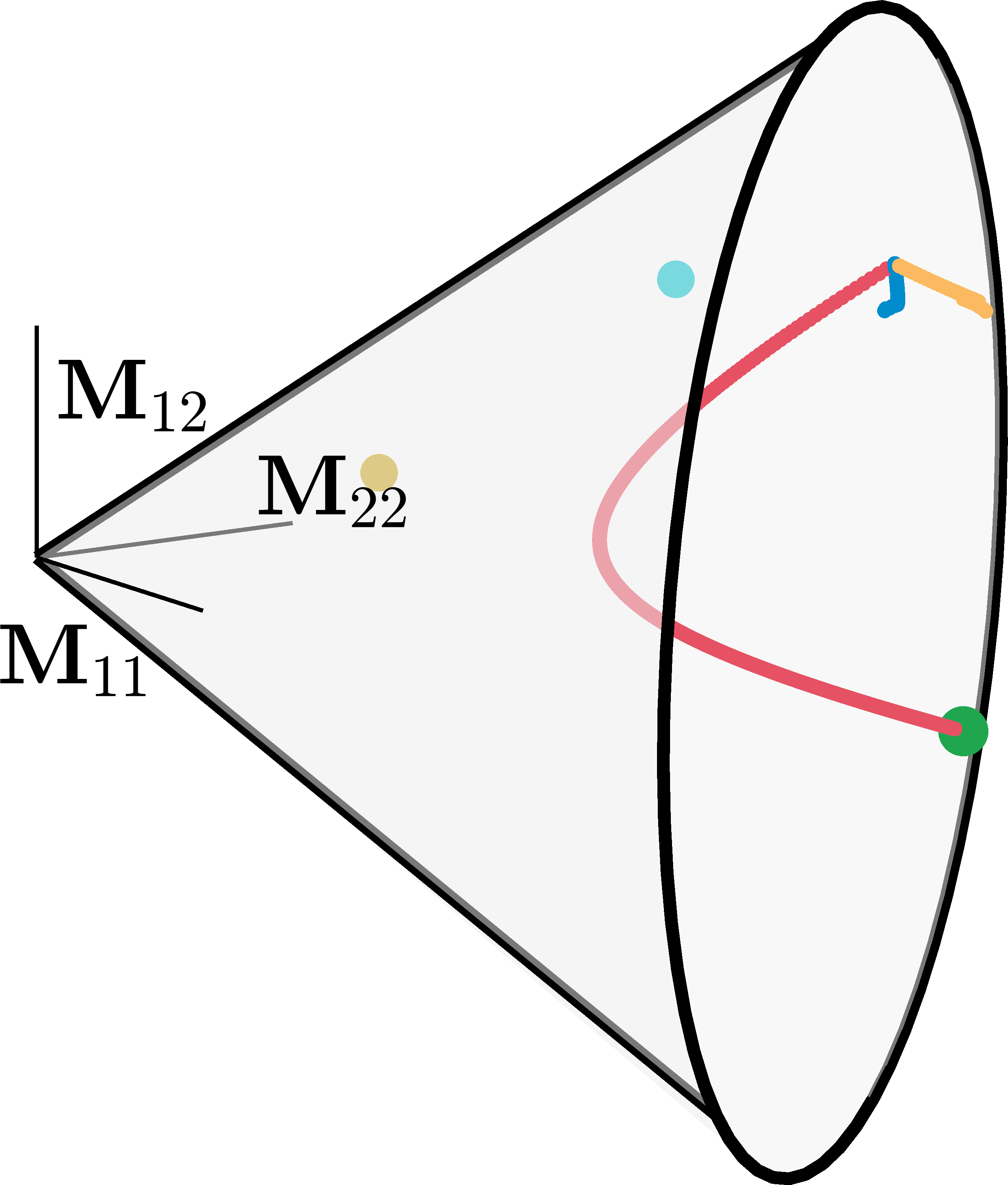}
		\caption{$\bm{K}_{\bm{M}}=0.5\bm{I}$}
		\label{subFig:TrackingComp02_Km05}
	\end{subfigure}
	\begin{subfigure}[b]{0.24\textwidth}
		\includegraphics[width=.8\textwidth]{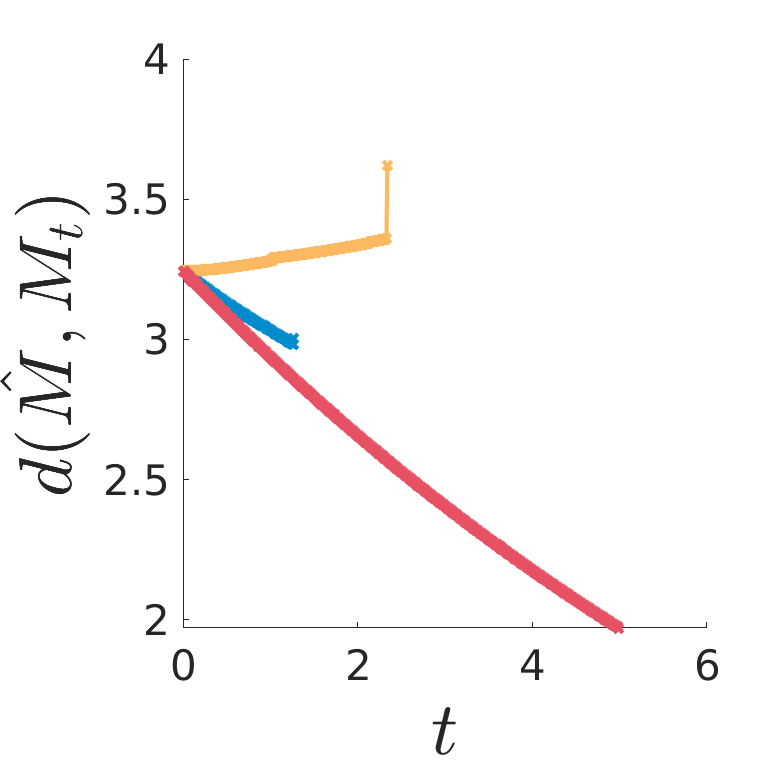}
		\includegraphics[width=.8\textwidth]{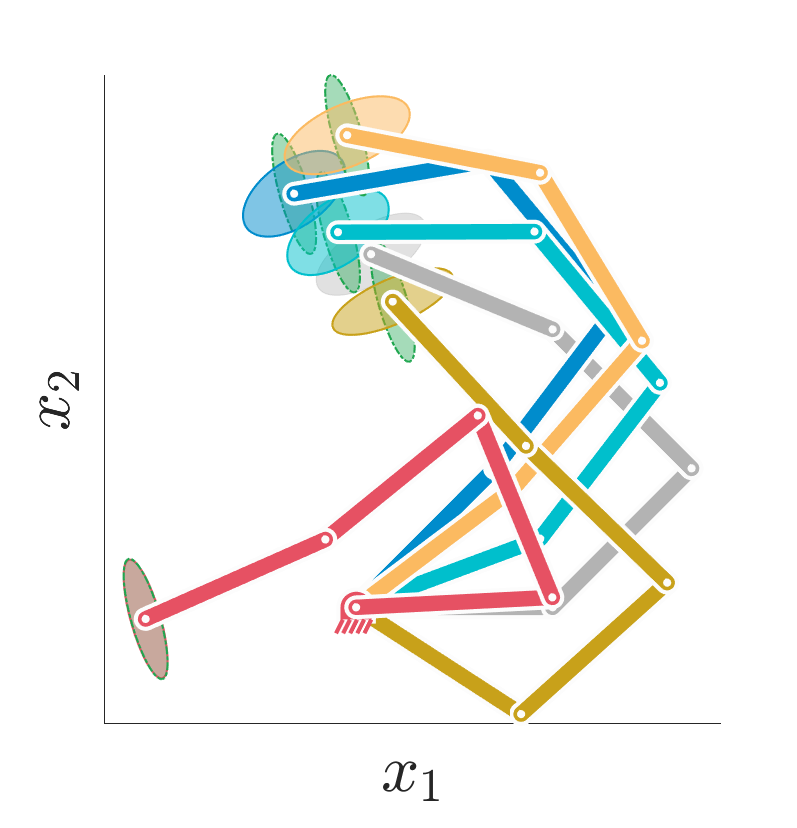}
		\includegraphics[width=.75\textwidth]{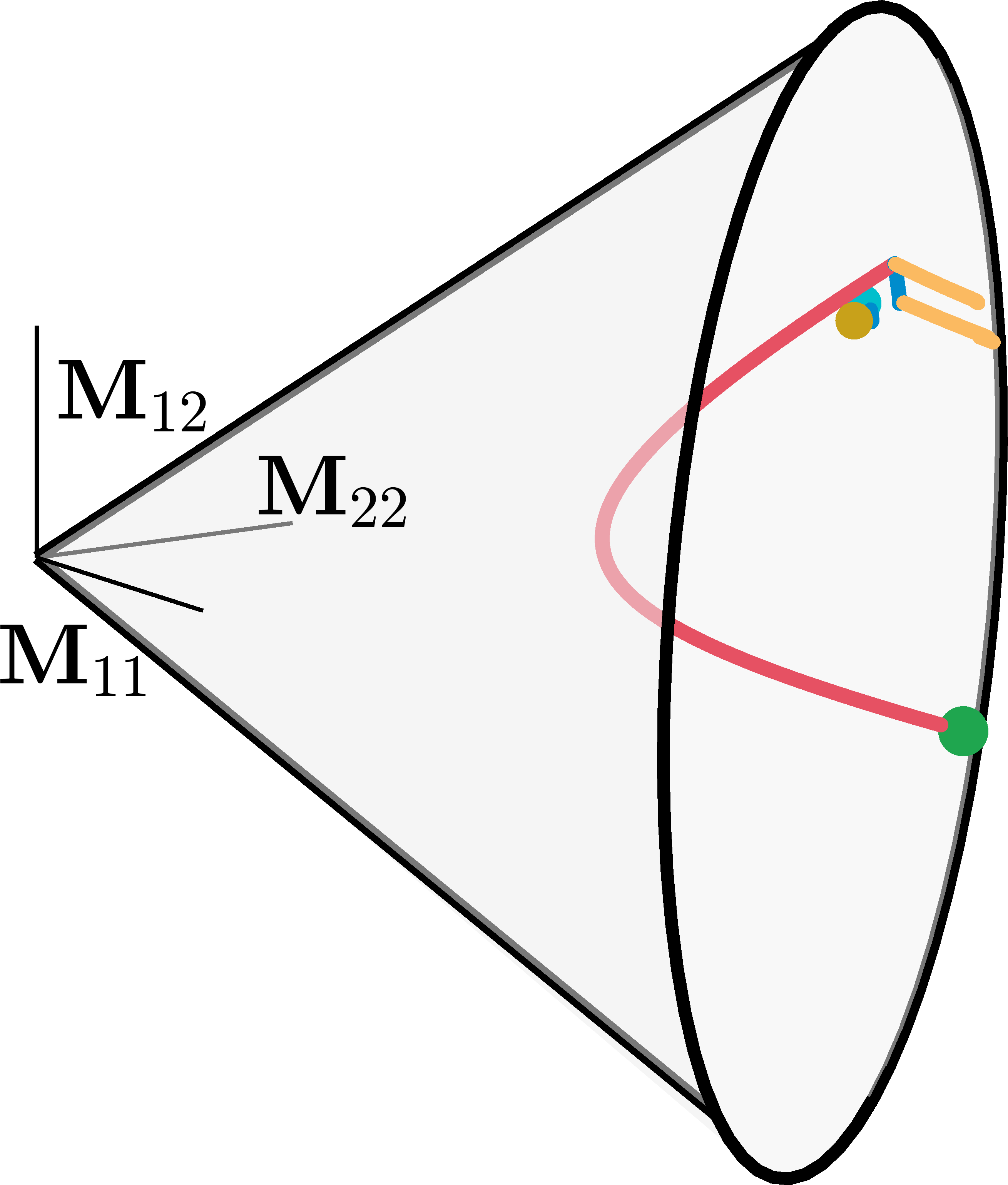}
		\caption{$\bm{K}_{\bm{M}}=0.1\bm{I}$}
		\label{subFig:TrackingComp02_Km01}
	\end{subfigure}
	\begin{subfigure}[b]{0.24\textwidth}
		\includegraphics[width=.8\textwidth]{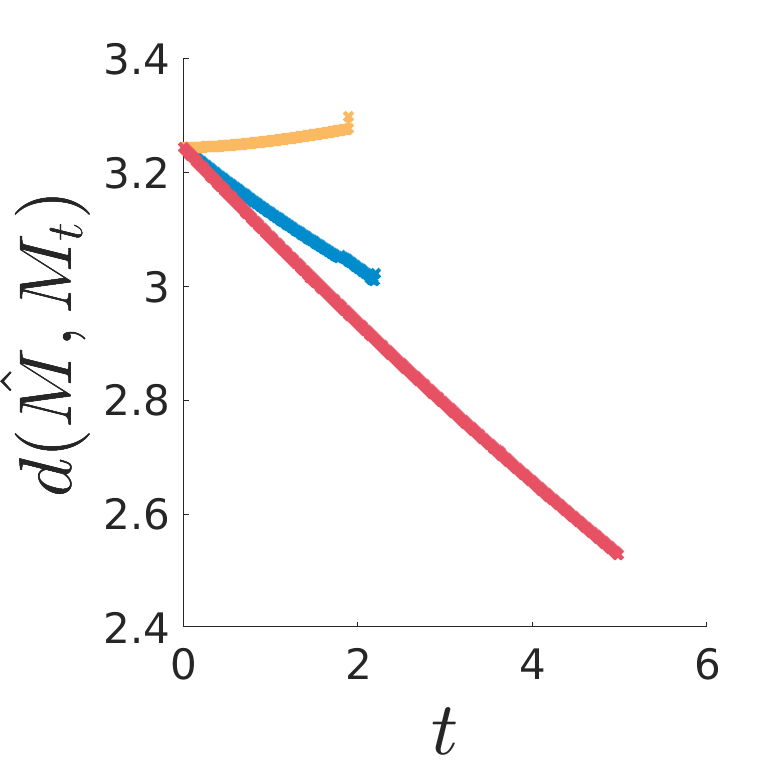}
		\includegraphics[width=.8\textwidth]{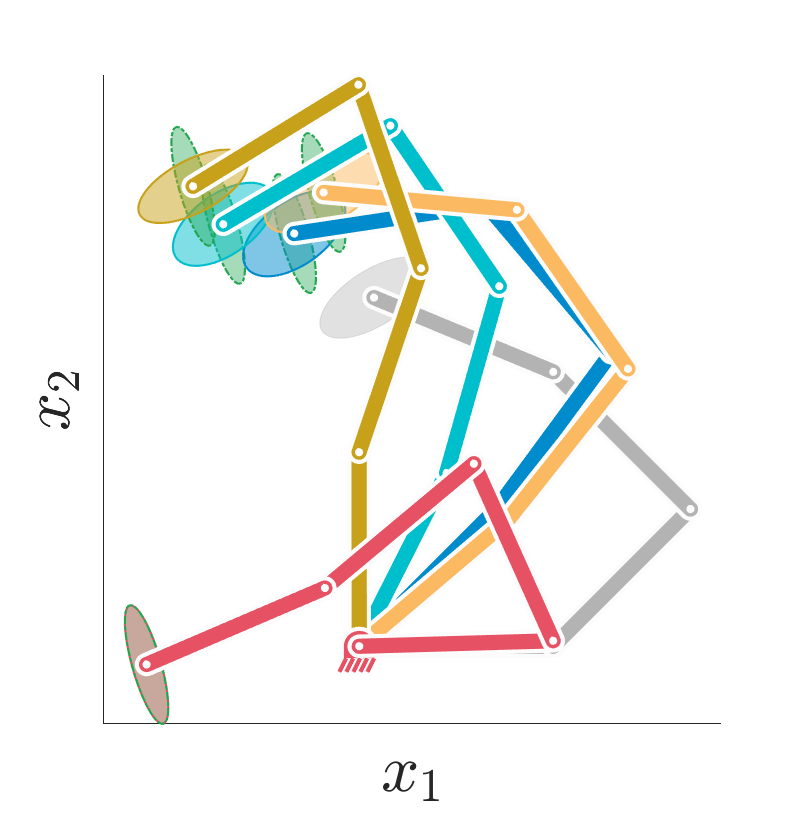}
		\includegraphics[width=.75\textwidth]{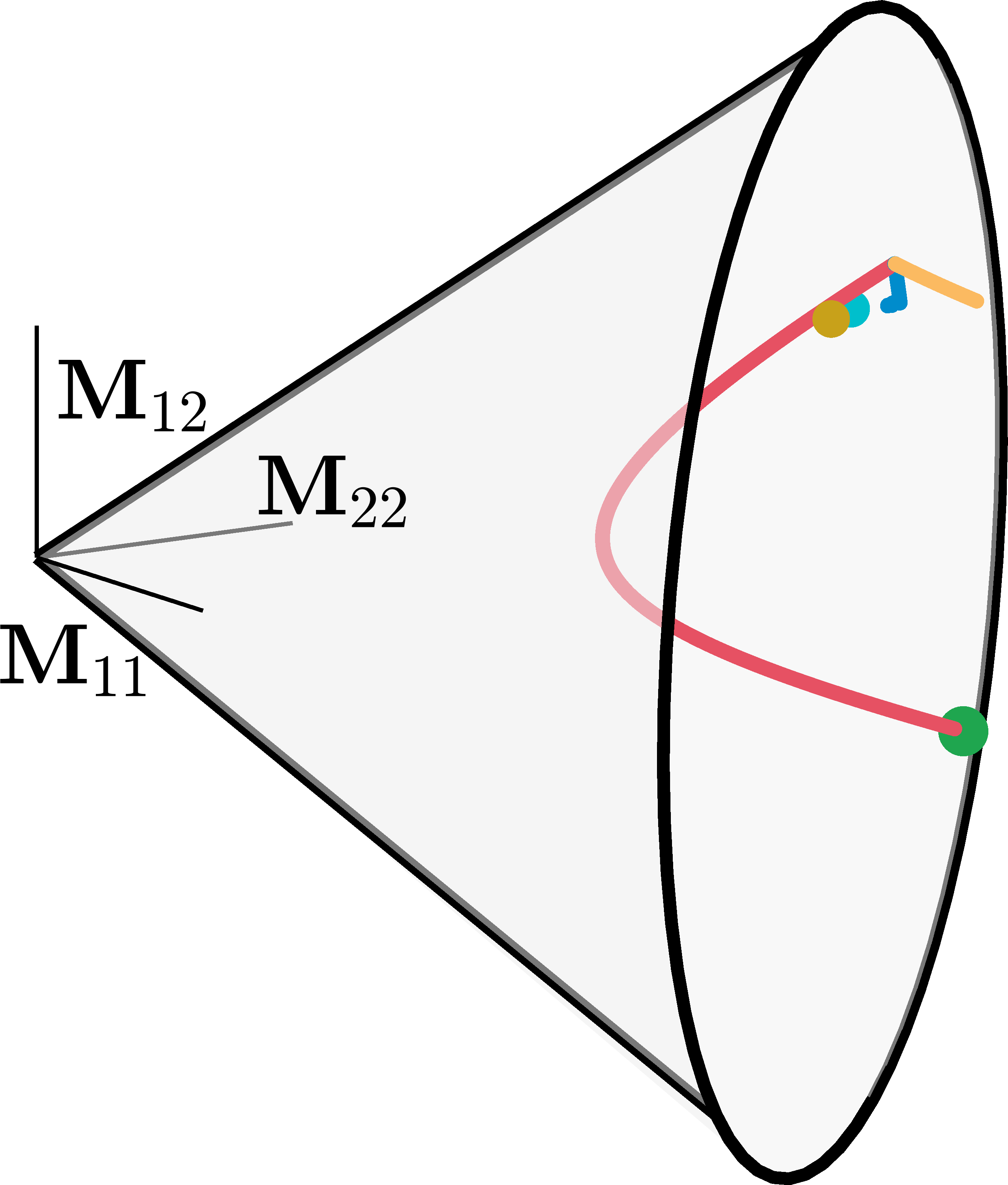}
		\caption{$\bm{K}_{\bm{M}}=0.05\bm{I}$}
		\label{subFig:TrackingComp02_Km005}
	\end{subfigure}
	\caption{Comparison of the performance of different manipulability tracking formulations for different gains $\bm{K}_{\bm{M}}$. 
	The organization of the graphs and the colors are identical to Fig.~\ref{Fig:TrackingComp}. The Cholesky-Jacobian-based Euclidean formulation is not shown.}
	\label{Fig:TrackingCompLowGains}
\end{figure*}

Previously, we hypothesized that the sudden changes in joint configuration when using the Euclidean and Cholesky-based Euclidean formulations in the second scenario are due to the path induced by the methods on the SPD manifold. In order to confirm this hypothesis, we reproduced the second test with lower gain values. Figure~\ref{Fig:TrackingCompLowGains} shows the convergence of the proposed geometry-aware controller, the Euclidean-based approach and the Cholesky-based Euclidean formulation for gain matrices equal to $\bm{I}$, $0.5\bm{I}$, $0.1\bm{I}$ and $0.05\bm{I}$.
We observe that, even for very low gains, both Euclidean and Cholesky-based Euclidean formulations lead to a sudden change in the joint configuration, resulting in an abrupt increase on the error measured between the current and desired manipulability ellipsoids (see Fig.~\ref{Fig:TrackingCompLowGains}-\emph{top}, \emph{middle}). Interestingly, the sudden changes occur at similar location along the path between the initial and desired manipulability ellipsoid independently of the gain value for both formulations (see Fig.~\ref{Fig:TrackingCompLowGains}-\emph{bottom}), therefore confirming our above statement. This can also be seen by looking at the yellow and dark blue robots of Fig.~\ref{Fig:TrackingCompLowGains}-\emph{middle} depicting the configurations before the jump, which are almost identical in all the graphs.

In the case in which the manipulability tracking task becomes a secondary objective, the 4-DoF planar robot is required to keep its end-effector at a fixed Cartesian position $\bm{\hat{x}}$ while minimizing the distance between its current and desired manipulability ellipsoids $\bm{M}$ and $\bm{\hat{M}}$. The four following approaches are considered for comparison with the proposed formulation \eqref{Eq:ManipTrackSecTask}. Firstly, we analyze the corresponding Euclidean manipulability-tracking controller
\begin{multline}
\dot{\bm{q}}_t = \bm{J}^\dagger \bm{K}_{\bm{x}}(\bm{\hat{x}}_t-\bm{x}_t) \\ + (\bm{I}-\bm{J}^\dagger\bm{J}) (\bm{\mathcal{J}}_{(3)}^\dagger)^\trsp \bm{K}_{\bm{M}} \text{vec}(\bm{\hat{M}}_t-\bm{M}_t) ,
\label{Eq:EuclFormulationNullspace}
\end{multline} 
where the difference between two manipulability ellipsoids is computed in Euclidean space, i.e., ignoring that manipulability ellipsoids belong to the set of SPD matrices. Secondly, we implement the corresponding Cholesky-based Euclidean manipulability controller
\begin{multline}
\dot{\bm{q}}_t = \bm{J}^\dagger \bm{K}_{\bm{x}}(\bm{\hat{x}}_t-\bm{x}_t) \\+ (\bm{I}-\bm{J}^\dagger\bm{J}) (\bm{\mathcal{J}}_{(3)}^\dagger)^\trsp \bm{K}_{\bm{M}} \text{vec}(\bm{\Delta L}_t\bm{\Delta L}_t^\trsp) ,
\label{Eq:CholEuclFormulationNullspace}
\end{multline} 
which ignores that manipulability ellipsoids lie on the SPD manifold but ensure a positive definite difference between two ellipsoids.
Thirdly, we analyze the Cholesky-Jacobian-based Euclidean manipulability controller
\begin{multline}
\dot{\bm{q}}_t = \bm{J}^\dagger \bm{K}_{\bm{x}}(\bm{\hat{x}}_t-\bm{x}_t) \\+ (\bm{I}-\bm{J}^\dagger\bm{J}) (\bm{\mathcal{J}}_{\text{chol}(3)}^\dagger)^\trsp \bm{K}_{\bm{M}} \text{vec}(\bm{\hat{L}}- \bm{L}),
\label{Eq:CholJacobianEuclFormulationNullspace}
\end{multline}
which tracks manipulability ellipsoids through their Cholesky decomposition.
Fourthly, we evaluate the gradient-based approach of~\citep{Rozo17IROS:ManTransfer} that implements the controller
\begin{equation}
\dot{\bm{q}}_t = \bm{J}^\dagger \bm{K}_{\bm{x}}(\bm{\hat{x}}_t-\bm{x}_t) - (\bm{I}-\bm{J}^\dagger\bm{J}) \alpha \nabla g_t(\bm{q}) ,
\label{Eq:GradientFormulation}
\end{equation}
where $\alpha$ is a scalar gain and
\begin{equation}
g_t(\bm{q}) = \log \det \Bigg( \frac{\bm{\hat{M}}_{t} + \bm{M}_{t}}{2}  \Bigg) - \frac{1}{2} \log \det \Big(\bm{\hat{M}}_{t}\bm{M}_{t}\Big) 
\label{Eq:SteinDiv}
\end{equation}
is a cost function based on Stein divergence (a distance-like function on the SPD manifold~\citep{Sra12}). The gain matrices $\bm{K}_{\bm{M}}$ are fixed as identity matrices and the scalar gain is set to $1$ for the comparison.

\begin{figure}[tbp]
	\centering
	\includegraphics[width=.42\textwidth]{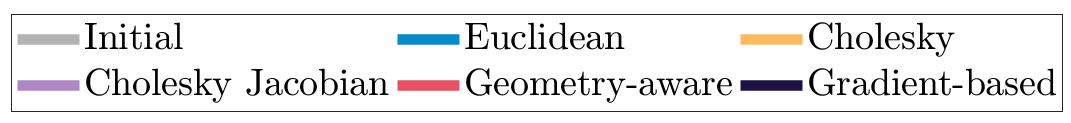}
	\begin{subfigure}[b]{0.48\textwidth}
		\centering
		\includegraphics[width=.45\textwidth]{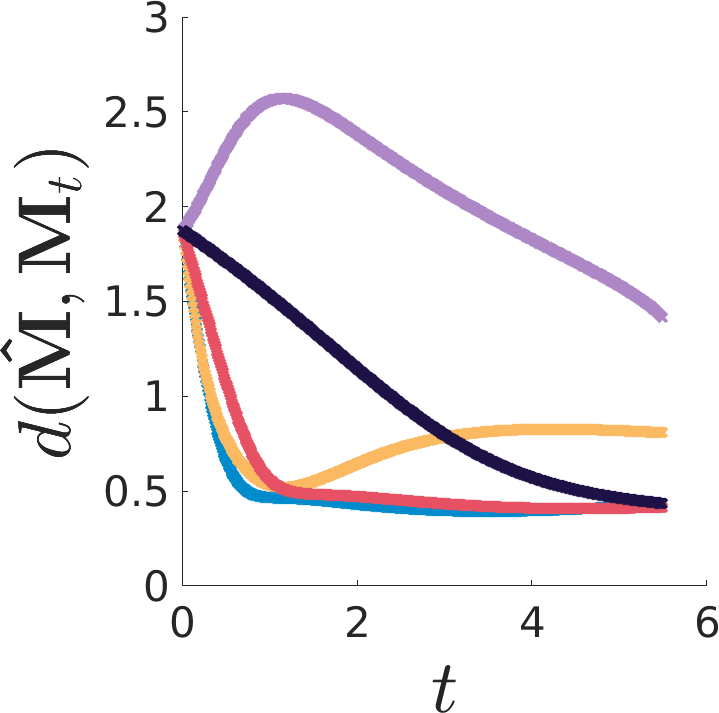}
		\includegraphics[width=.45\textwidth]{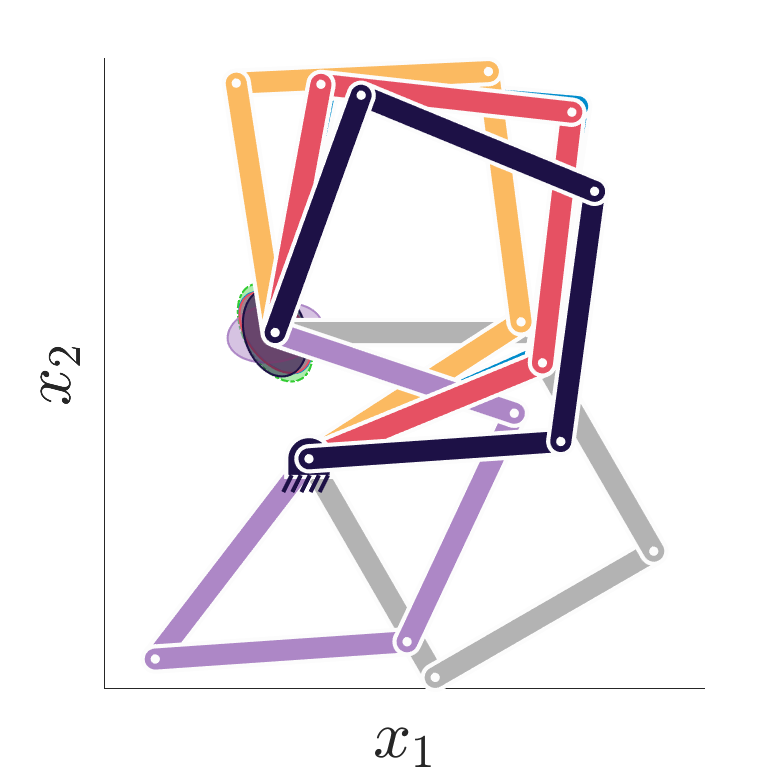}
		\caption{}
		\label{subFig:CompCost01}
	\end{subfigure}
	\begin{subfigure}[b]{0.48\textwidth}
		\centering
		\includegraphics[width=.45\textwidth]{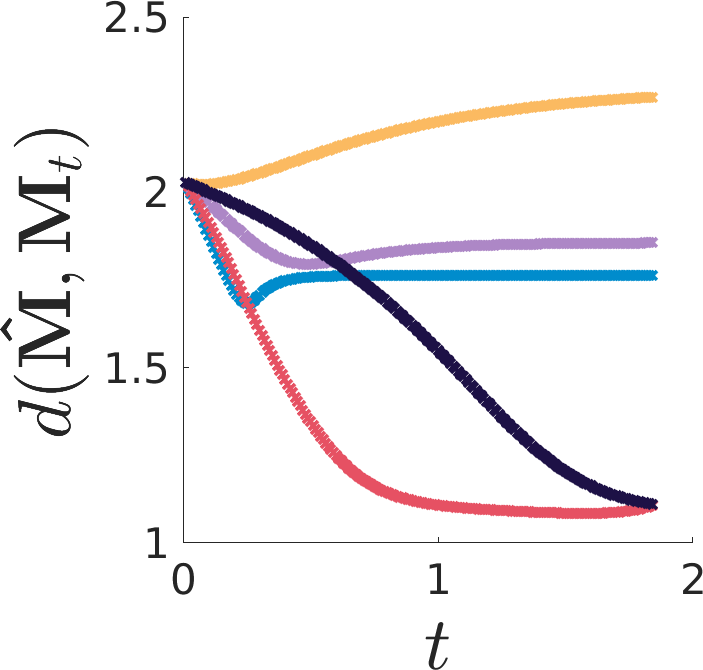}
		\includegraphics[width=.45\textwidth]{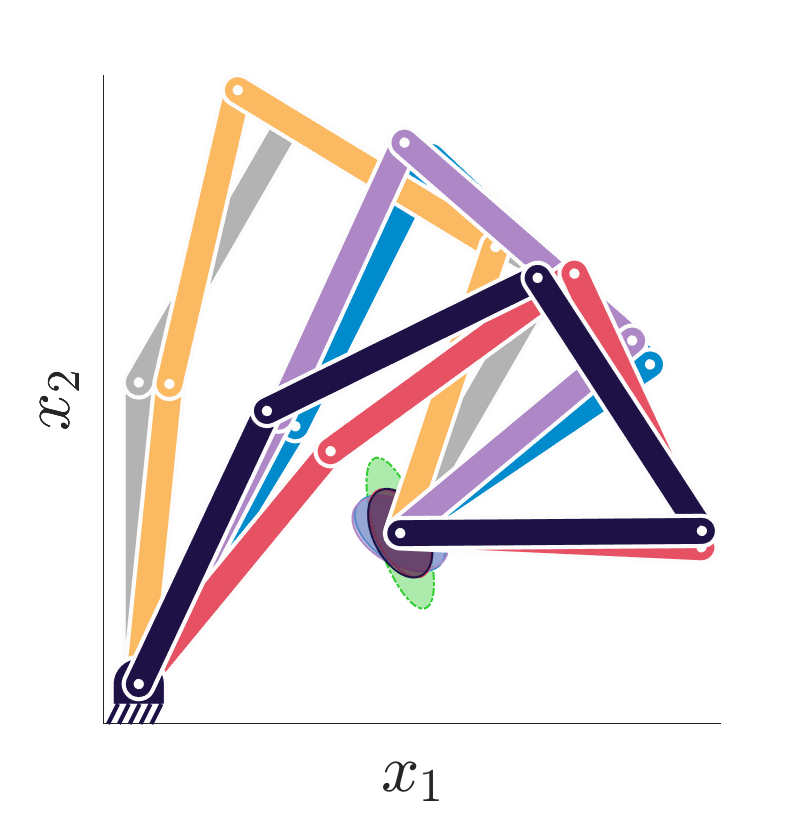}	
		\caption{}
		\label{subFig:CompCost02}
	\end{subfigure}
	\caption{Performance comparison of the different manipulability-based redundancy resolution formulations. Two cases are shown with varying initial robot configuration and desired manipulability. The \emph{left} graph shows the convergence of the affine invariant distance between the current and the desired manipulability ellipsoid over time. The distances for the Euclidean, Cholesky-based Euclidean, Cholesky-Jacobian-based Euclidean, geometry-aware and gradient-based approaches are respectively depicted in blue, yellow, lila, red, and purple. The \emph{right} graph shows the initial and final posture of the robot along with the final manipulability ellipsoids. The initial posture of the robot is depicted in light gray. The final postures and the corresponding manipulability ellipsoids for the different methods are depicted in the same color as the distances. The desired manipulability ellipsoid is depicted in green.}
	\label{Fig:CostComp}
	\vspace{-0.2cm}
\end{figure}
\begin{table}[t]
	\renewcommand*{\arraystretch}{1.2}
	\caption{Final distances $d(\bm{\hat{M}},\bm{M}_t)$ between the current and desired manipulability ellipsoids for the performance comparison of the different manipulability-based redundancy resolution formulations.}
	\label{Tab:CompCost}
	\scriptsize
	\begin{center}
		\setlength\tabcolsep{2.5pt}
		\begin{tabular}{c|c|c|c|c|c|}
			\textbf{Approach} & Euclidean & Cholesky & Cholesky Jac. & Geometry-aware & Gradient-based\\
			\hline
			\textbf{Fig.~\ref{subFig:CompCost01}} & $0.433$ & $0.808$ & $1.418$ & $0.416$ & $0.436$\\
			\hline
			\textbf{Fig.~\ref{subFig:CompCost02}} & $1.763$ & $2.271$ & $1.856$ & $1.101$ & $1.110$\\
			\hline
		\end{tabular}
	\end{center}
	\normalsize
\end{table}

Figure~\ref{Fig:CostComp} shows the convergence rate for the manipulability-based redundancy resolution of the aforementioned approaches. Two tests were carried out by varying the initial configuration of the robot and the desired manipulability ellipsoid. In both cases, both geometry-aware and gradient-based approaches converge to a similar final robot configuration (see Fig.~\ref{subFig:CompCost01},~\ref{subFig:CompCost02}-\emph{right}), with similar values of the affine-invariant distance between the final and desired manipulability ellipsoids (see Fig.~\ref{subFig:CompCost01},~\ref{subFig:CompCost02}-\emph{left} and Table~\ref{Tab:CompCost}). More importantly, the proposed geometry-aware manipulability tracking approach shows a faster convergence than the gradient-based method, with a lower computational cost ($3.5$ ms and $4.2$ ms per time step, with non-optimized Matlab code on a laptop with 2.7GHz CPU and 32 GB of RAM). This notable difference may be attributed to the fact that despite both methods take into account the geometry of manipulability ellipsoids, our approach is more informative about the kinematics of the robot through the use of the manipulability Jacobian $\bm{\mathcal{J}}(\bm{q})$. 

Note that for some specific initial robot configurations and desired manipulability ellipsoids, the Euclidean manipulability-tracking controller \eqref{Eq:EuclFormulationNullspace} shows a slightly faster convergence rate than our method (see Fig.~\ref{subFig:CompCost01}). However, this Euclidean formulation again leads to unstable behaviors in some configurations (see Fig.~\ref{subFig:CompCost02}), where the distance between the final and desired manipulability ellipsoids remains high compared to the two geometry-aware approaches. This poor tracking performance can be attributed to the fact that the Euclidean difference between two SPD matrices is an approximation that is only valid if the matrices are close enough to each other. Thus, similarly to Euclidean controller aimed at tracking manipulability ellipsoids as first task \eqref{Eq:EuclFormulation}, the Euclidean manipulability-based redundancy resolution is only effective if the current and desired ellipsoids are very similar. Moreover, the distance between the final and desired manipulability ellipsoids remains higher than for the geometry-aware methods and the Euclidean controller by using the Cholesky-based Euclidean manipulability-based redundancy resolution. This tendency is similar to the observations made for the tracking of manipulability ellipsoids as main objective and is due to the fact that the controller \eqref{Eq:CholEuclFormulationNullspace} induces paths on the manifold that are not close to geodesics.
Furthermore, the Cholesky-Jacobian-based Euclidean controller shows a poor tracking performance for the two considered scenarios. Notably, the distance between the current and desired ellipsoids is largely increased before decreasing slowly in the first case (see Fig.~\ref{subFig:CompCost01}). Moreover, in some configurations, the final distance remains high compared to the geometry-aware approaches as shown by Fig.~\ref{subFig:CompCost02}. These behaviors are due to the fact that the controller~\eqref{Eq:CholJacobianEuclFormulationNullspace} does not follow geodesic paths on the SPD manifold.

The reported results supported our hypothesis that geometry-aware manipulability controllers result in good tracking performance while providing stable convergence regardless of the manipulability tracking error. This was observed when manipulability tracking was the main task and a secondary objective of the robot. Moreover, our manipulability-based redundancy resolution approach outperforms the gradient-based method. Furthermore, our controller permits to directly exploit the variability information of a task, given in the form of a 4th-order covariance tensor, through the gain matrix of the controller. This allows the robot to exploit the precision required while tracking a manipulability ellipsoid either as main or secondary objective. This operation is not available in the gradient-based method used for comparison, since the corresponding controller gain is a scalar.

\subsubsection{Comparisons with manipulability-based optimization}
We compare our tracking approach against two state-of-the-art manipulability-based optimization methods widely used to improve robots posture for task execution. We first evaluate our geometry-aware controller against manipulability volume maximization. Then, we compare our controller to the compatibility index maximization~\citep{Chiu87:RedundantRbtCtrl}, where the distance from the ellipsoid center to its surface is maximized along a specified direction. To do so, we consider two 8-DoF planar robots that are required to track a desired Cartesian velocity trajectory that leads to an L-shape path in the Cartesian space. In order to achieve high dexterity in motion, the first robot is requested to track a desired manipulability ellipsoid whose main axis is elongated along the direction of motion. The second robot varies its posture in order to maximize either the manipulability volume or the compatibility index along the direction of motion.

\begin{figure}[tbp]
	\begin{subfigure}[b]{0.22\textwidth}
		\centering
		\includegraphics[width=\textwidth]{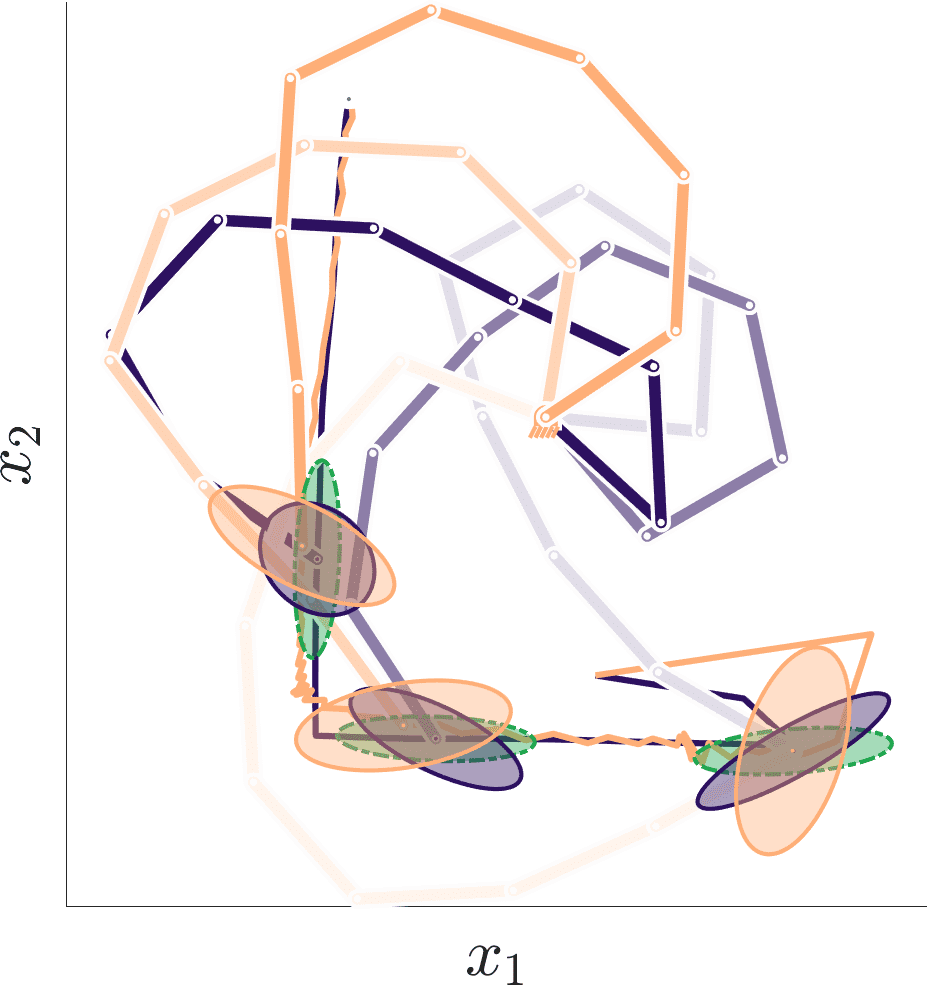}
		\caption{}
		\label{subFig:ComparisonVolume}
	\end{subfigure}
	\begin{subfigure}[b]{0.075\textwidth}
		\centering
		\includegraphics[width=\textwidth]{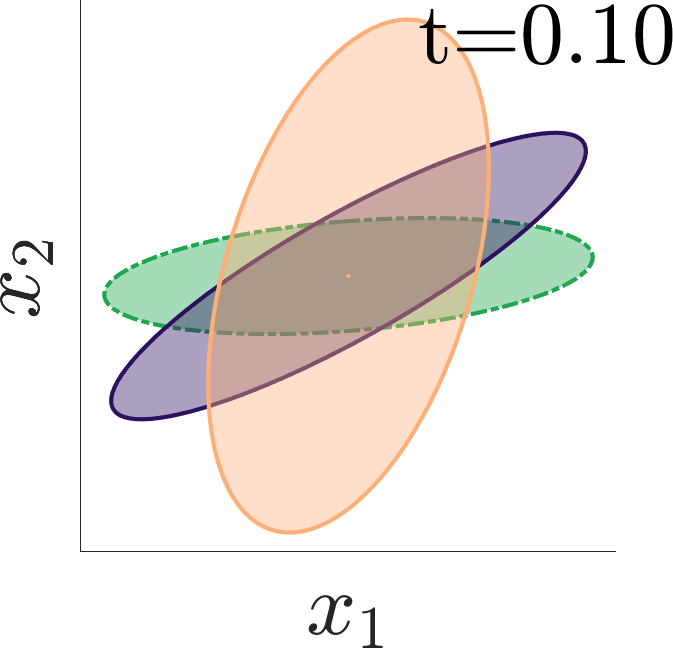}
		\includegraphics[width=\textwidth]{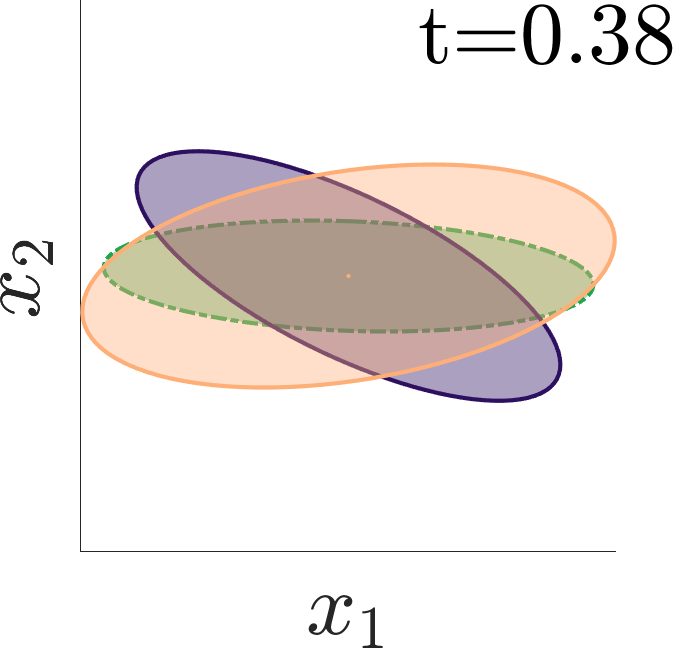}
		\includegraphics[width=\textwidth]{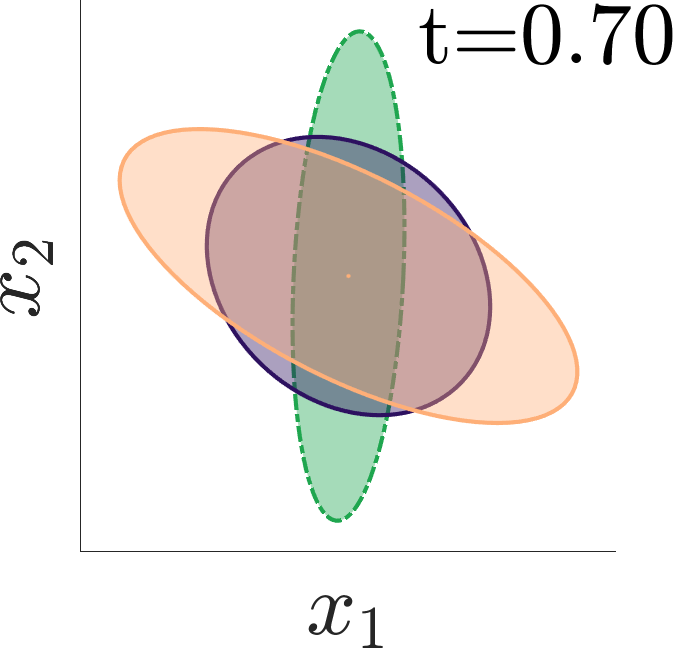}
		\caption{}
		\label{subFig:ComparisonVolumeEllipsoids}
	\end{subfigure}
	\begin{subfigure}[b]{0.16\textwidth}
		\centering
		\includegraphics[width=\textwidth]{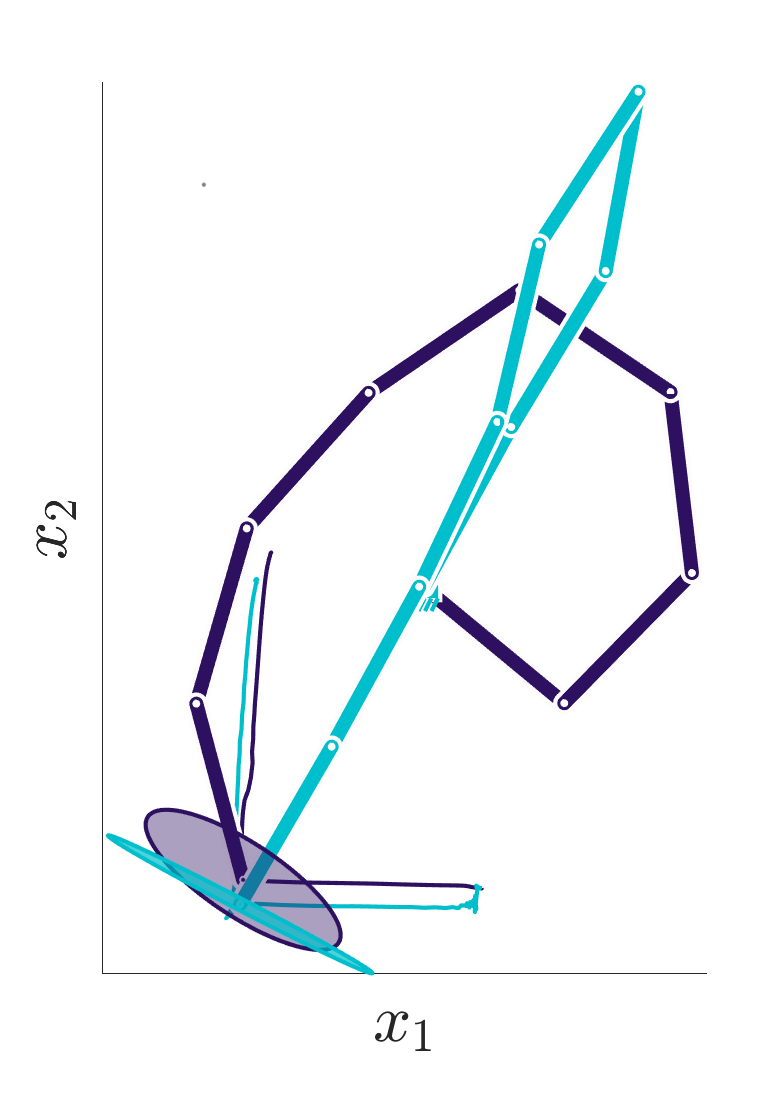}
		\caption{}
		\label{subFig:ComparisonCompatibilityIndex}
	\end{subfigure}
	\caption{(\emph{a}) Comparison of our manipulability tracking controller (in purple) with the manipulability volume maximization (in yellow). The main axis of the desired manipulability ellipsoids (in green) are aligned with the direction of motion in order to allow high velocities during the task execution. The robot colors become darker with the evolution of the movement. (\emph{b}) Close-up plots of the manipulabilities represented in (\emph{a}). (\emph{c}) Comparison of our manipulability tracking controller (in purple) with the compatibility index maximization (in light blue).}
	\label{Fig:ComparisonVolume}
\end{figure}

\begin{figure}[tbp]
	\centering
	\includegraphics[width=.23\textwidth]{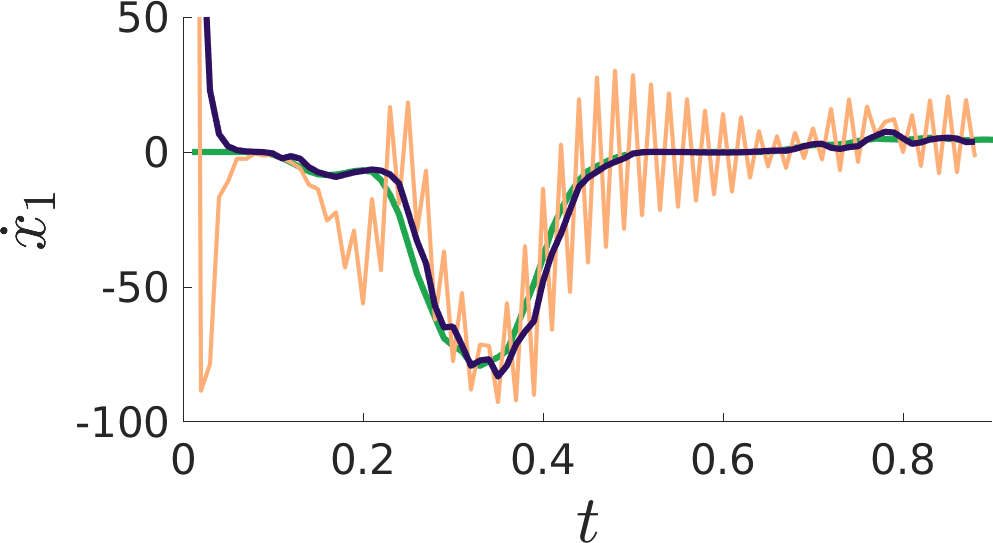}
	\includegraphics[width=.23\textwidth]{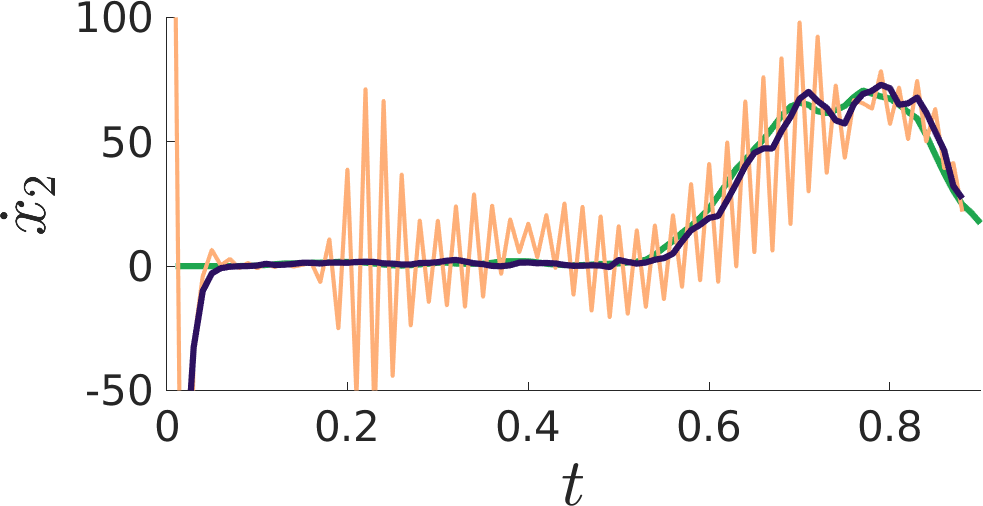}
	\caption{Cartesian velocities achieved with our manipulability tracking controller (purple) and the volume maximization approach (yellow) as secondary objective for a Cartesian velocity controller. The gain of the velocity controller are equal for both approaches. The desired velocities are shown in green.}
	\label{Fig:ComparisonVolumeVelocities}
\end{figure}

Fig.~\ref{subFig:ComparisonVolume} shows the resulting joint configurations and manipulability ellipsoids of the two robots at different stages of the task where the second robot maximizes the manipulability volume as secondary objective. We observe that the main axis of the manipulability ellipsoid obtained with the volume maximization approach is often perpendicular to the direction of motion, which often occurs as this method does not consider any geometric information about the desired manipulability ellipsoid. Also, since the resulting posture leads to ellipsoids that are not consistent with the task requirement (task velocity control directions) and degrade the robot capabilities, this becomes unstable when the gain of the velocity tracking controller is increased to achieve higher Cartesian velocities, as shown in Fig.~\ref{Fig:ComparisonVolumeVelocities}. Conversely, the robot tracking a desired manipulability ellipsoid successfully completes the task when higher velocities are required.

The main advantage of maximizing the compatibility index over the volume is that the directions in which the ellipsoid should be elongated are specified. However, this approach favors robot configurations that may be close to singularities as the manipulability ellipsoids corresponding to these posture are flat ellipsoids that can be largely elongated (see Fig.~\ref{subFig:ComparisonCompatibilityIndex}). This effect exacerbates when the compatibility index maximization is the main task of the robot, as this is not required to match a specific position in Cartesian space. \citet{Chiu88:TaskCompatibility} extended the compatibility index optimization approach by defining the compatibility cost as a weighted sum, allowing the maximization or minimization of the ellipsoid along several directions. This method provides more flexibility on the resulting ellipsoid due to the weighted combination, at the cost of a laborious tuning. Moreover, the orientation and elongation of the main axes of the ellipsoid after the optimization are hard to infer from the cost weights.

In contrast to the considered manipulability-based optimization methods, the proposed geometry-aware controllers seeks to fit the full desired manipulability ellipsoid in all its directions. Singular configurations can therefore be easily avoided by defining appropriate desired manipulability ellipsoids. Moreover, our manipulability controller allows the tracking of any manipulability ellipsoid, including those providing a compromise between dexterity in motion and force exertion along any axis. This is not possible when using the compatibility index approach of~\citep{Chiu87:RedundantRbtCtrl} as it always favors the dexterity in motion over force or vice-versa. Although this compromise might be achievable using the compatibility index approach of~\citep{Chiu88:TaskCompatibility}, our method does not require a laborious tuning process. Manipulability tracking is also hard to achieve through manipulability volume maximization as there is no explicit control on the resulting ellipsoid main axes.

\section{Experiments}  
\label{sec:Experiment}
Previously, in our former work~\citep{Jaquier18}, we showed the benefits of including the manipulability redundancy resolution controller in the nullspace of a position controller for a pushing and an insertion task. Contrary to the result obtained by the position controller alone, the posture of the robot significantly varied during the execution of the tasks to be compatible with their respective force requirements as a consequence of the force manipulability tracking.

In this section, we extensively evaluate the proposed tracking formulation with different robotic platforms and different types of manipulability ellipsoids in simulation. The approach is evaluated to track a desired manipulability for grasping with an Allegro hand (four 4-DoFs fingers) and to track a desired center of mass manipulability with NAO and Centauro robots (25- and 39-DoFs, respectively).
We then illustrate and evaluate the proposed manipulability transfer approach in a bimanual task using a Baxter robot (two 7-DoFs arms) and a couple of Franka Emika Panda robots (7-DoFs). 

\subsection{Manipulability tracking for a robotic hand}
In the context of robotic hands, manipulability ellipsoids have been used to analyze their performances in grasping tasks~\citep{Prattichizzo12}. In this experiment, we aim at modifying the posture of a robotic hand to match a desired manipulability ellipsoid while grasping an object.

For the case of multiple arm systems, the set of joint velocities of constant unit norm $\|\bm{\dot{q}}_a\| = \|(\bm{\dot{q}}_1^\trsp, \ldots, \bm{\dot{q}}_C^\trsp)^\trsp\| = 1$ is mapped to the Cartesian velocity space $\bm{\dot{x}}_a = (\bm{\dot{x}}_1^\trsp, \ldots, \bm{\dot{x}}_C^\trsp)^\trsp$ through
\begin{equation} 
\| \bm{\dot{q}}_a \|^2  = \bm{\dot{q}}_a^\trsp \bm{\dot{q}}_a  = \bm{\dot{x}}_a^\trsp(\bm{G}_a^{\dagger\trsp}\bm{J}_a\bm{J}_a^\trsp\bm{G}_a^{\dagger})^{-1}\bm{\dot{x}}_a,
\label{Eq:MultipleArmsVelocityMapping}
\end{equation}
with the Jacobian $\bm{J}_a = \text{diag}(\bm{J}_1, \ldots, \bm{J}_C)$, the grasp matrix $\bm{G}_a = (\bm{G}_1, \ldots, \bm{G}_C)$ and $C$ the number of arms. Therefore, the velocity manipulability ellipsoid of the $C$-arms system is given by $\bm{M}^{\bm{\dot{x}}_a} = \bm{G}_a^{\dagger\trsp}\bm{J}_a\bm{J}_a^\trsp\bm{G}_a^{\dagger}$ ~\citep{Chiacchio91}. Note that the system is modeled under assumptions that the arms are holding a rigid object with a tight grasp.

In this first experiment, the Allegro hand was required to track a desired manipulability, while maintaining relative positions between the different fingers. This experiment aims at emulating how humans adapt their finger configuration to the task at hand while grasping an object.
In this experiment, the desired velocity manipulability ellipsoid was designed by the experimenter to be a medium-size isotropic ellipsoid. The purpose of this design is to provide the hand with the capability to perform a displacement of the object while being resistant to external perturbations in all the directions. For example, in the case where the hand is holding a pen, it is desirable that the pen can be moved with dexterity, while the hand should resist to perturbations due to the pen-surface contacts.

The fingers were controlled according to a leader-follower strategy~\citep{Luh87}. Therefore the thumb joints were moved to track the desired manipulability ellipsoid using the controller \eqref{Eq:ManipTrackFirstTask} and the other fingers were required to maintain constant relative end-effector positions with respect to the thumb end-effector, while tracking the manipulability as secondary objective with the redundancy controller \eqref{Eq:ManipTrackSecTask}.
The center of the object was considered as the central position between the four fingers of the hand and the contact points were assumed to be at the finger tips. 

\begin{figure}[tpb]
	\centering
	\begin{subfigure}[b]{0.14\textwidth}
		\centering
		\includegraphics[width=\textwidth]{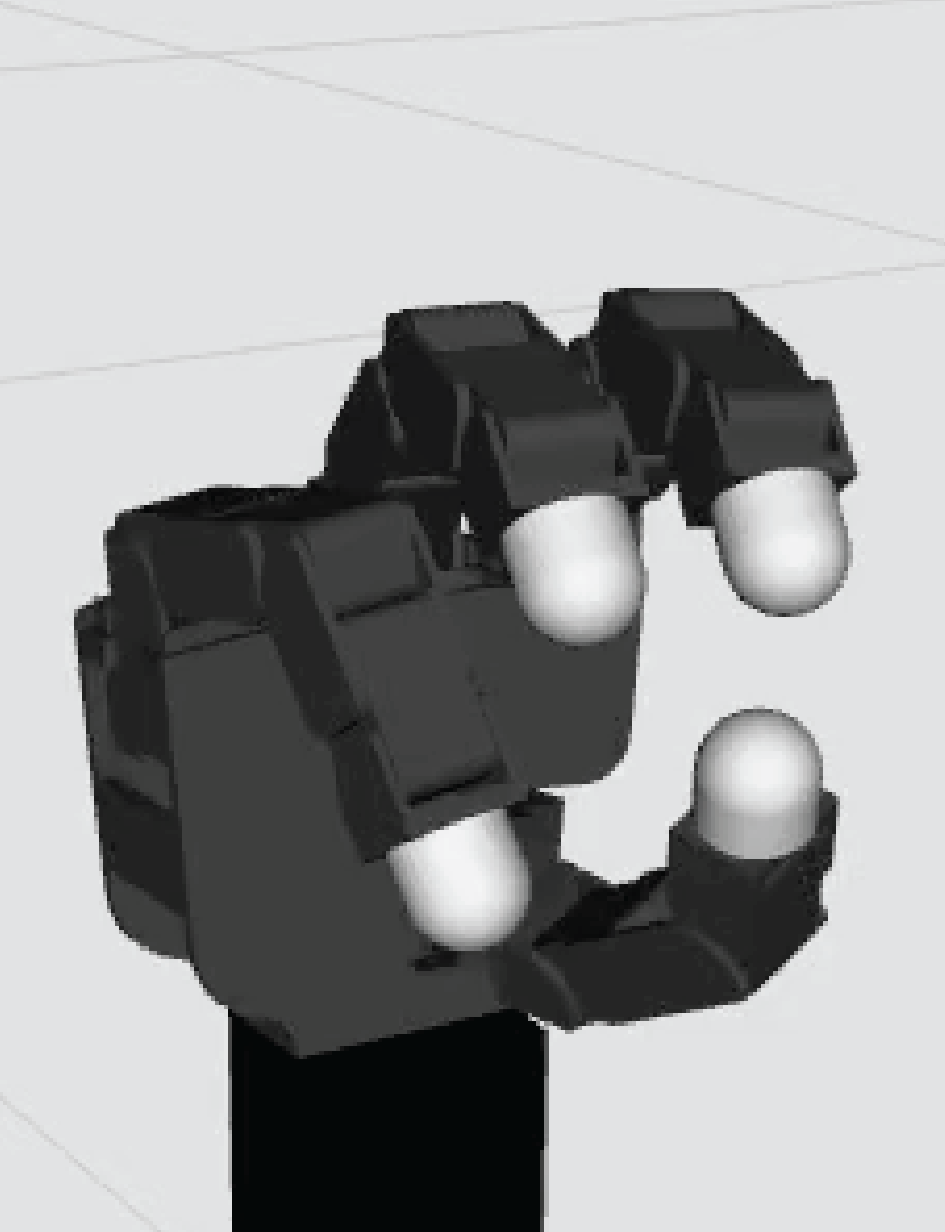}
		\caption{Initial}
		\label{subFig:AllegroJointsConfInit}
	\end{subfigure}
	\begin{subfigure}[b]{0.14\textwidth}
		\centering
		\includegraphics[width=\textwidth]{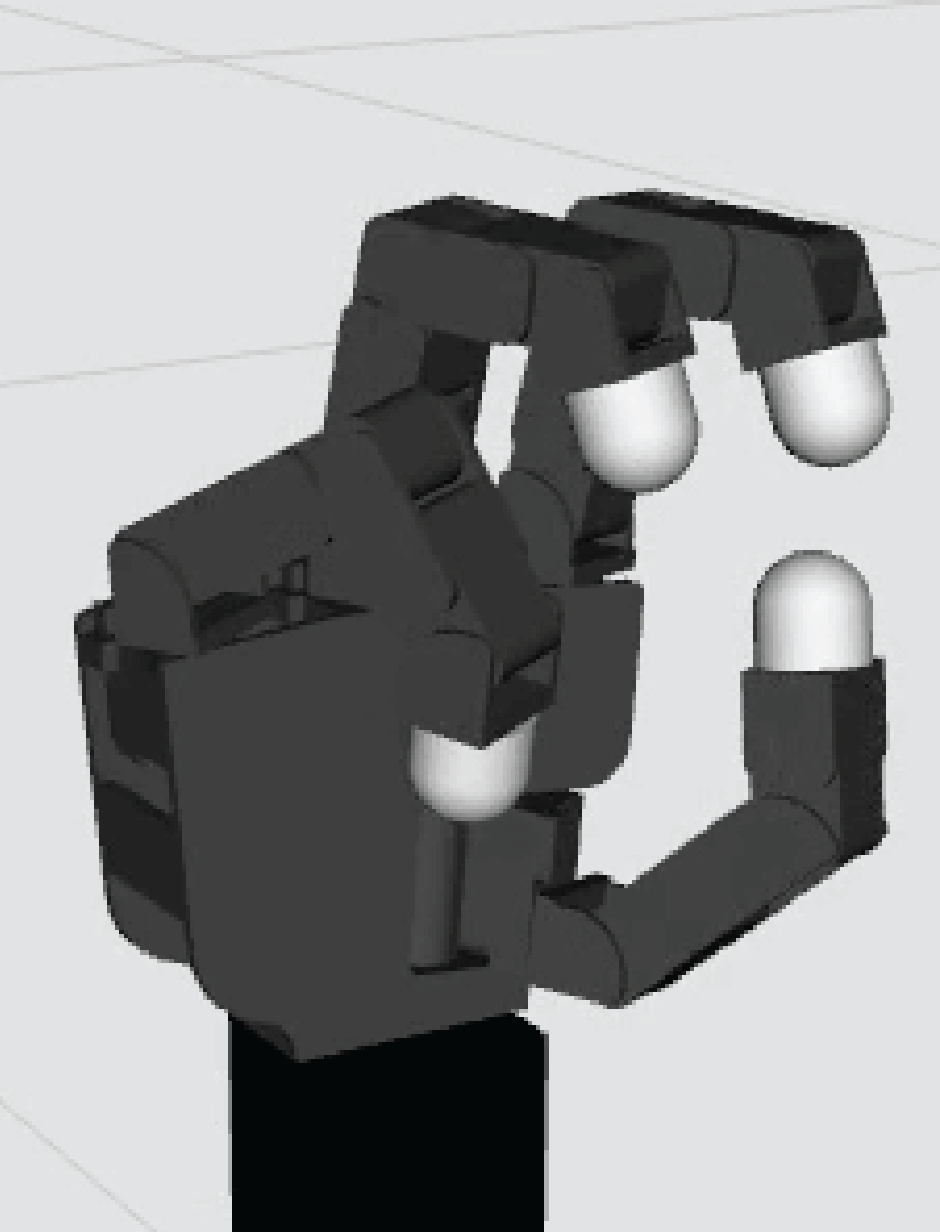}
		\caption{Final}
		\label{subFig:AllegroJointsConfFinal}
	\end{subfigure}
	\begin{subfigure}[b]{0.19\textwidth}
		\centering
		\includegraphics[width=0.45\textwidth]{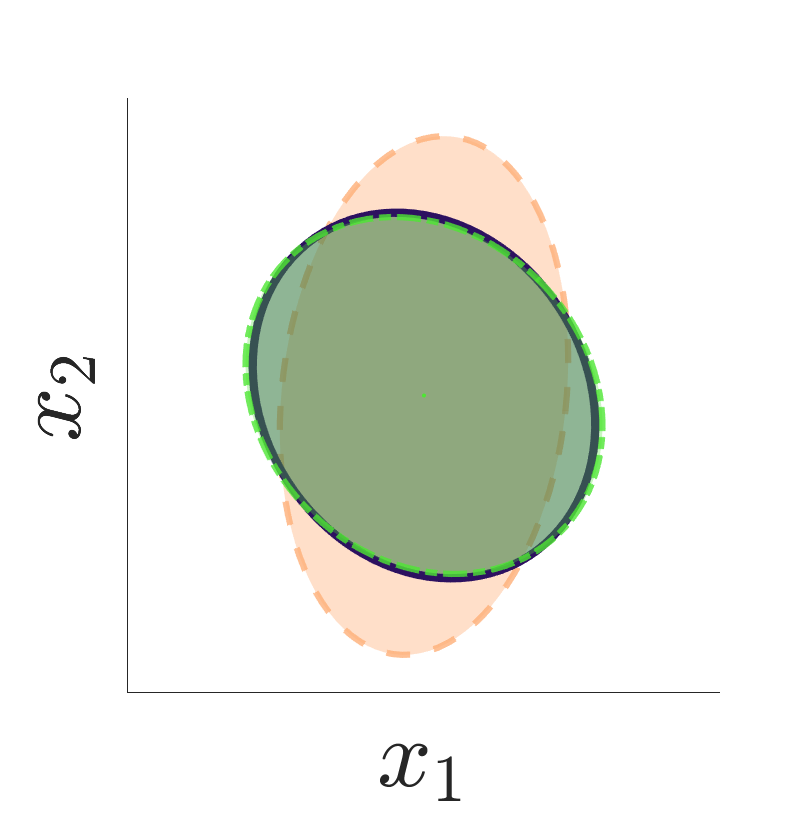}
		\includegraphics[width=0.45\textwidth]{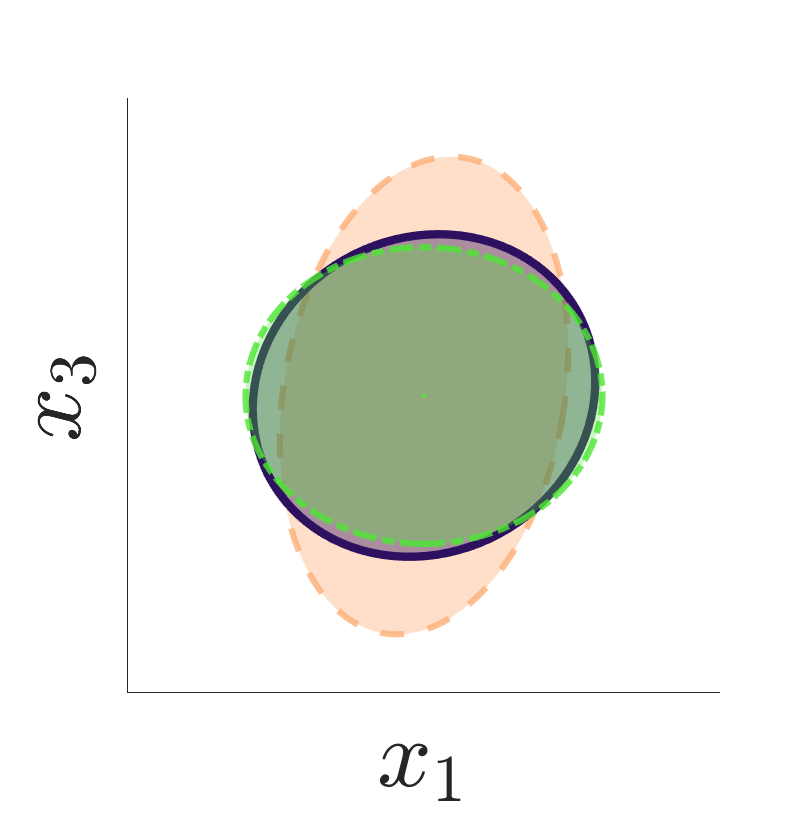}
		\includegraphics[width=0.45\textwidth]{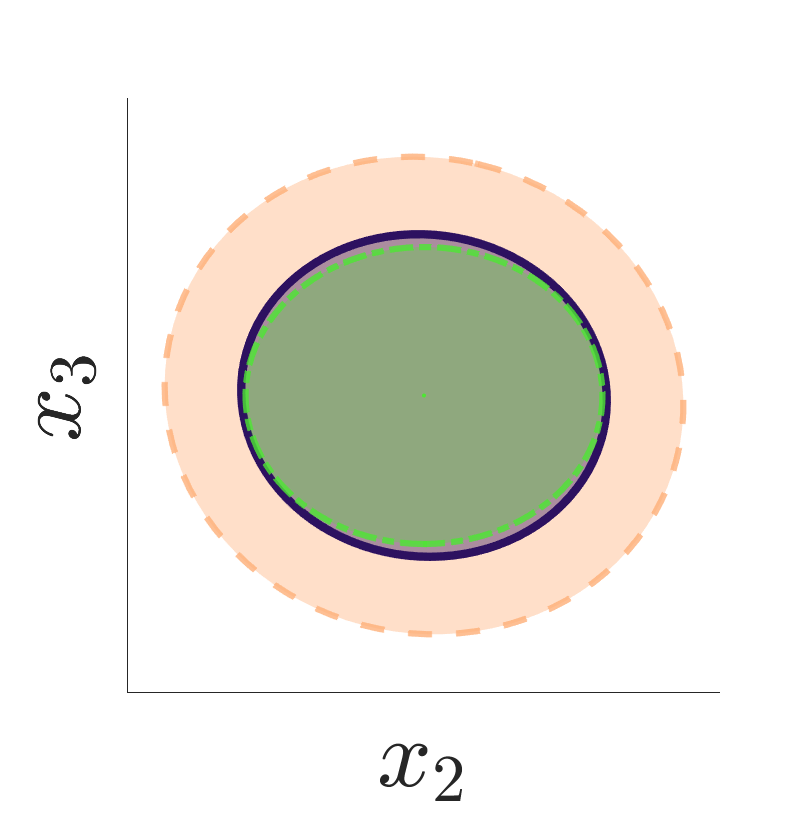}
		\includegraphics[width=0.45\textwidth]{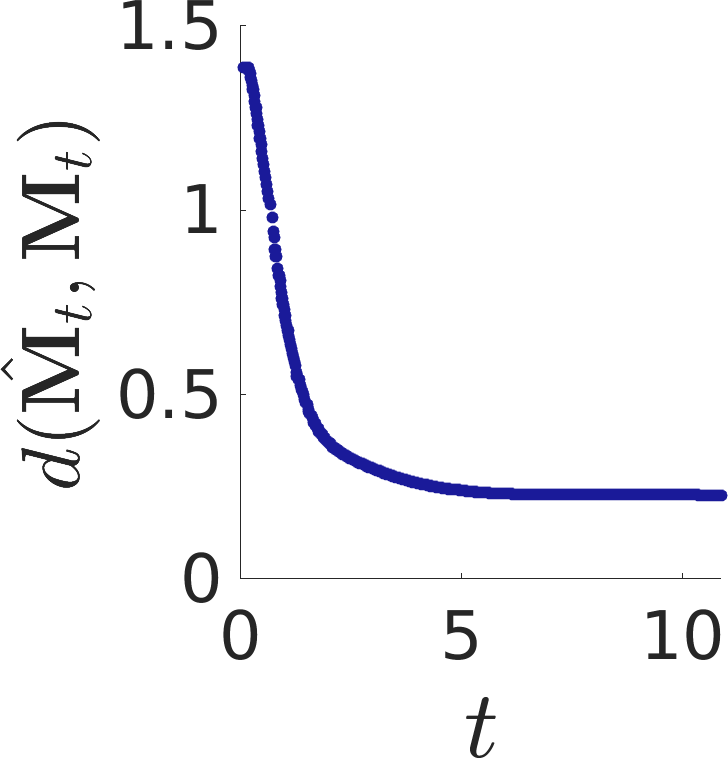}
		\caption{}
		\label{subFig:AllegroManip}
	\end{subfigure}
	\caption{Manipulability tracking for grasping tasks with the Allegro hand in simulation. \emph{(a)} and \emph{(b)} show the initial and final pose of the robot, respectively. \emph{(c)} Initial, final and desired manipulability ellipsoids respectively depicted in yellow, dark purple and green. The \emph{bottom-right} graph shows the evolution of the distance between the current and desired manipulability ellipsoid over time (in seconds).}
	\label{Fig:Allegro}
	\vspace{-0.2cm}
\end{figure}

Figures~\ref{subFig:AllegroJointsConfInit} and~\ref{subFig:AllegroJointsConfFinal} show an example of adaptation of the posture of the hand to track a desired velocity manipulability ellipsoid for a grasp defined by the user. As expected, the robot modified its joint configuration in order to match, as accurately as possible, the desired velocity manipulability (see Fig.~\ref{subFig:AllegroManip}). Note that the manipulability tracking in this experiment can only be achieved partially, because the robotic hand is also required to maintain the initial grasp. Nevertheless, this tracking may be further improved if the dimensionality of the nullspace of the main task is higher (e.g. not all the finger tips are position-constrained), or using a higher DoF robotic hand. 

\subsection{Manipulability tracking for a humanoid center of mass}
\label{subsec:Exper_COM}
An interesting use of manipulability ellipsoids arises when these are defined at the center of mass (CoM) of humanoid robots, which permits to analyze their capabilities to accelerate the CoM in locomotion~\citep{Azad17:CoMmanipulability, Gu15:CoMmanipulability}, or to evaluate how resistant they can be to external perturbations using the force manipulability at a specific humanoid posture. With the goal of getting some insights on the role of CoM manipulability ellipsoids in legged robots, we designed manipulability tracking experiments using two different floating-base robots in simulation, namely, the humanoid NAO and the Centauro robot~\citep{Baccelliere2017} within the Pyrobolearn framework~\citep{Delhaisse2019:Pyrobolearn}.

Specifically, we required the robots to track a desired manipulability ellipsoid defined at its CoM while keeping balance. We assumed a strict hierarchy of tasks that gave the highest priority to the task of maintaining the CoM position over the support polygon and zero velocity at all contact points with the floor, while the manipulability tracking was considered a secondary task. Under the aforementioned assumptions, we implemented the inverse kinematics-based controller for floating-base robots proposed in~\citep{Mistry08}, which we briefly introduce here. First, let us define the Jacobian for the primary task as 
\begin{equation}
\bm{J}_b = \left[\begin{matrix}
\bm{J}_{\text{feet}} \\
\bm{J}_{\text{CoM,xy}} \\
\end{matrix}\right] ,
\label{Eq:JacobianBalancing}
\end{equation} 
where $\bm{J}_{\text{feet}}$ represents the Jacobians for the position/orientation of the robot feet while $\bm{J}_{\text{CoM,xy}}$ is the Jacobian for the projection of the CoM onto the $(x,y)$ plane (assuming the gravity vector is in the $z$ direction). Next, we define the vector of primary desired velocities $\bm{x}_b$ (i.e. velocities of the robot feet and CoM), noting that all the robot feet velocities must equal zero in order to maintain constraints, therefore 
\begin{equation}
\bm{\dot{x}}_b = \left[\begin{matrix}
\bm{0} \\
\bm{\dot{x}}_{\text{CoM}} \\
\end{matrix}\right] ,
\end{equation}
where $\bm{\dot{x}}_{\text{CoM}}$ is the velocity at the robot CoM so that it lies in the support polygon. 

Regarding the secondary task, that is, the manipulability tracking at the robot CoM, we first compute the Jacobian at the CoM $\bm{J}_{\text{CoM}}$ for floating-base robots as in~\citep{Mistry08}, which allows us to calculate manipulability ellipsoids of the types introduced in Section~\ref{sec:ManTrack}. Depending on which type of manipulability we require the robot to track, we can use any of the manipulability Jacobians~\eqref{Eq:VelocityManipJacobian},~\eqref{Eq:ForceManipJacobian} or~\eqref{Eq:DynManipJacobian} to compute the desired joint velocities $\bm{\dot{q}}$ for the manipulability tracking task using~\eqref{Eq:ManipTrackFirstTask}. So, the full joint velocity controller for legged robots required to keep balance while tracking a desired manipulability ellipsoid at their CoM is defined as 
\small
\begin{equation}
\bm{\dot{q}} = \left[\begin{matrix}
\bm{I}_{n\times n} \\
\bm{0}_{6\times n} \\
\end{matrix}\right] ^\trsp\,\bigg( \bm{J}_b^{\dagger} \bm{\dot{x}}_b + \bm{N}_b \, (\bm{\mathcal{J}}_{(3)}^\dagger)^\trsp \, \bm{K}_{\bm{M}} \, \text{vec}\Big(\text{Log}_{\bm{M}_t}(\bm{\hat{M}}_t)\Big)\bigg),
\label{Eq:CoMtracking}
\end{equation} 
\normalsize
where the first term is included in order to account for the virtual joints of legged robots, $n$ is the number of DoF of the robot, and $\bm{N}_b$ is the nullspace of the Jacobian~\eqref{Eq:JacobianBalancing}. 

\begin{figure}[tpb]
	\centering
	\begin{subfigure}[b]{0.14\textwidth}
		\centering
		\includegraphics[width=\textwidth]{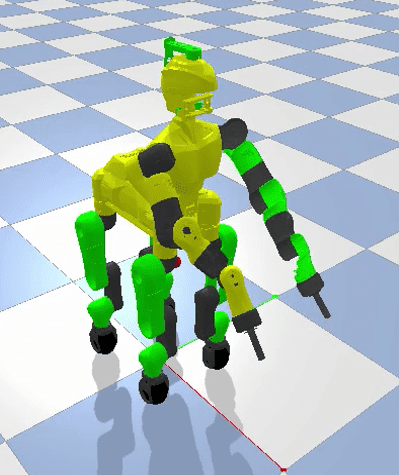}
		\caption{Initial}
		\label{subFig:CentauroPoseInit}
	\end{subfigure}
	\begin{subfigure}[b]{0.14\textwidth}
		\centering
		\includegraphics[width=\textwidth]{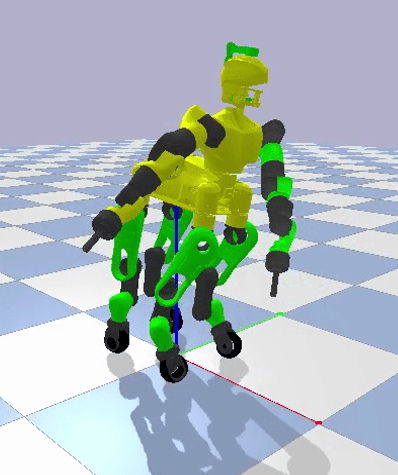}
		\caption{Final}
		\label{subFig:CentauroPoseFinal}
	\end{subfigure}
	\begin{subfigure}[b]{0.19\textwidth}
		\centering
		\includegraphics[width=0.45\textwidth]{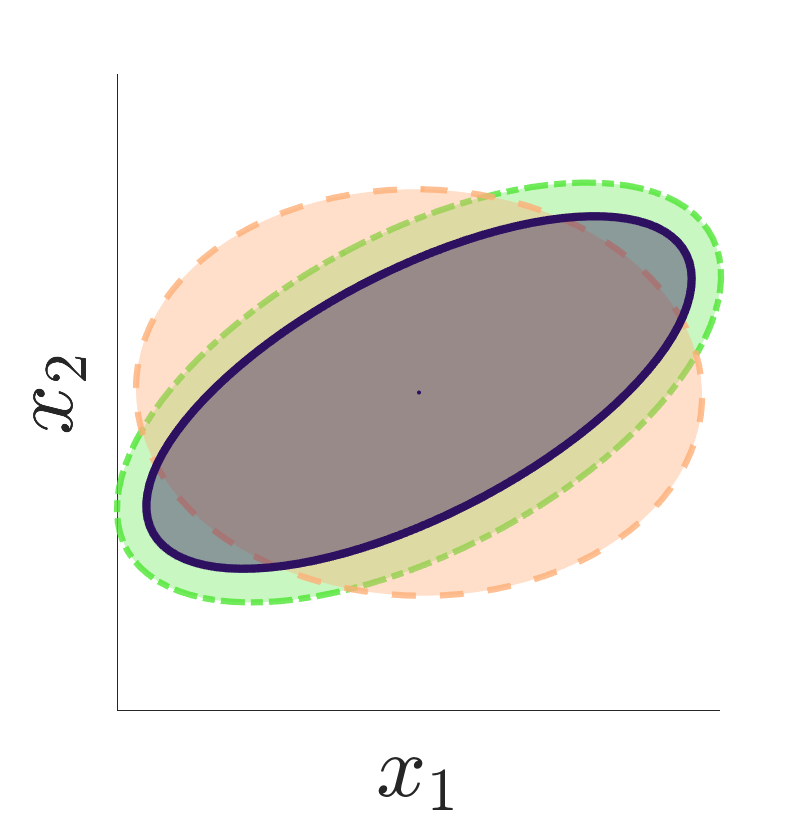}
		\includegraphics[width=0.45\textwidth]{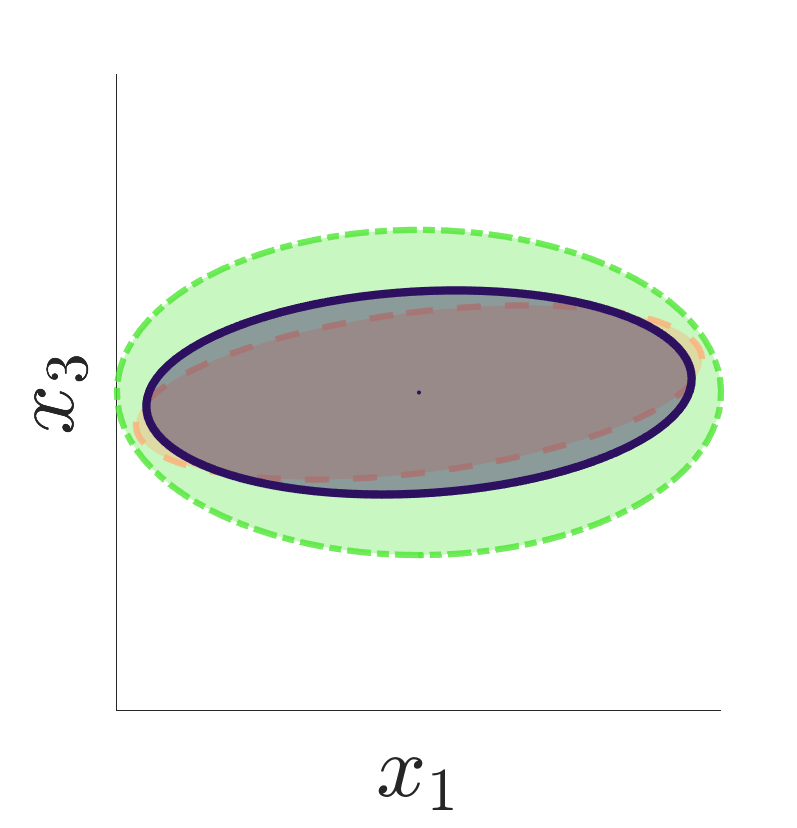}
		\includegraphics[width=0.45\textwidth]{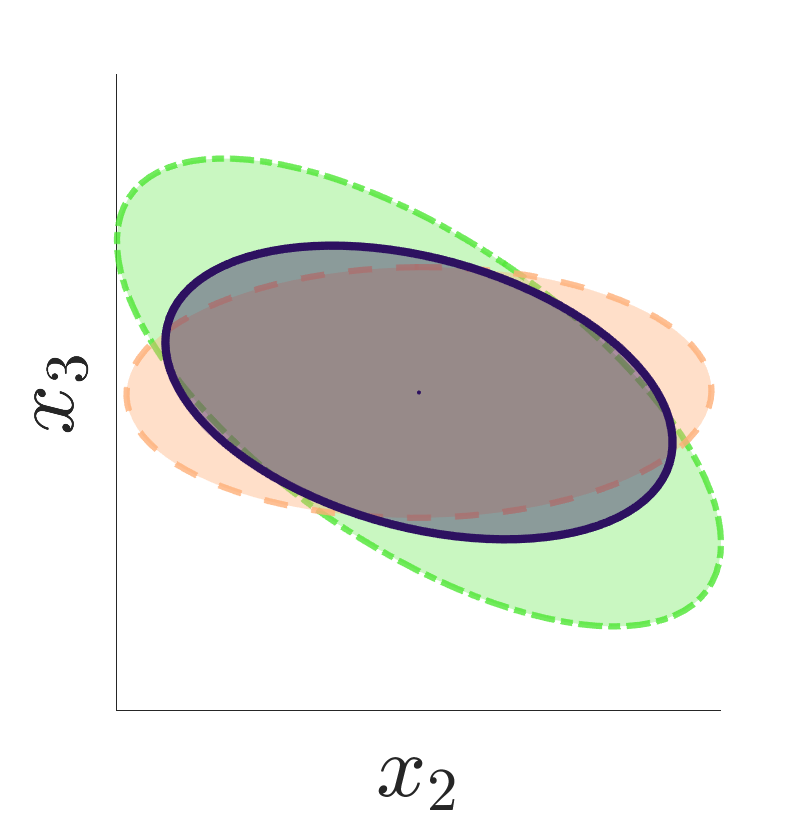}
		\includegraphics[width=0.45\textwidth]{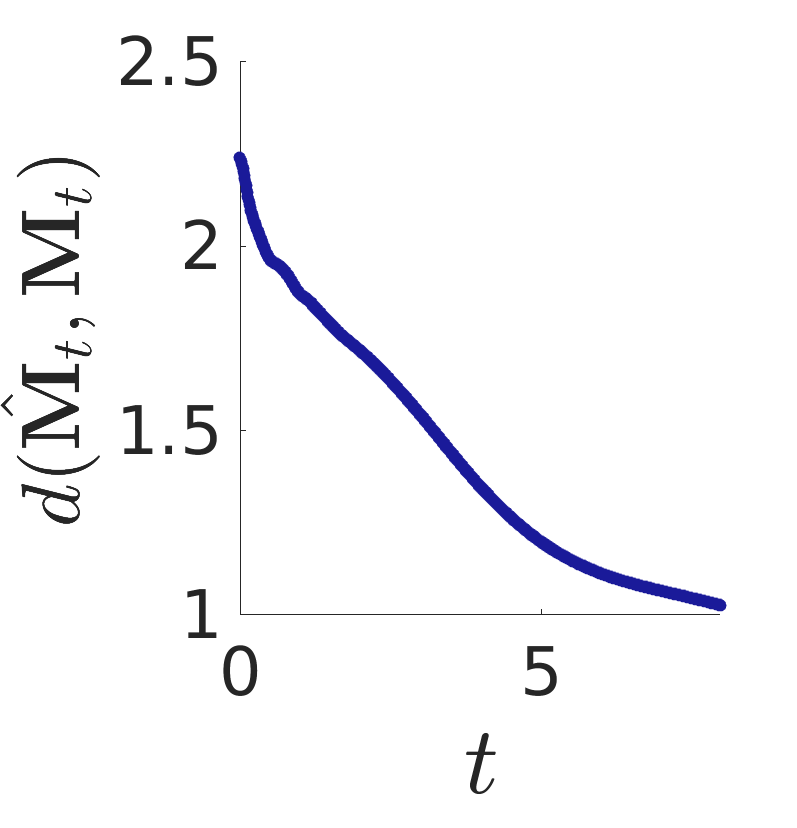}
		\caption{}
		\label{subFig:CentauroManip}
	\end{subfigure}
	\caption{Tracking of the COM manipulability with the Centauro robot in simulation. \emph{(a)} and \emph{(b)} show the initial and final pose of the robot, respectively. \emph{(c)} Initial, final and desired manipulability ellipsoids respectively depicted in yellow, dark purple and green. The \emph{bottom-right} graph shows the evolution of the distance between the current and desired manipulability ellipsoid over time, given in seconds.}
	\label{Fig:Centauro}
\end{figure}

\begin{figure}[tpb]
	\centering
	\begin{subfigure}[b]{0.14\textwidth}
		\centering
		\includegraphics[width=\textwidth]{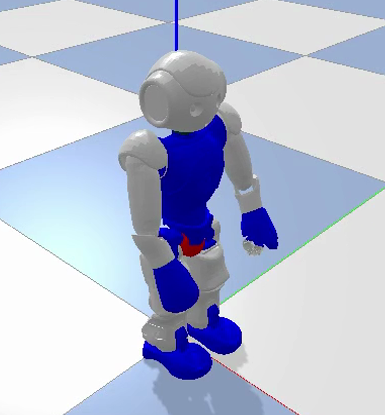}
		\caption{Initial}
		\label{subFig:NaoPoseInit}
	\end{subfigure}
	\begin{subfigure}[b]{0.14\textwidth}
		\centering
		\includegraphics[width=\textwidth]{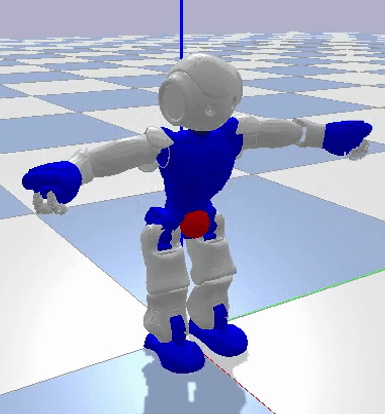}
		\caption{Final}
		\label{subFig:NaoPoseFinal}
	\end{subfigure}
	\begin{subfigure}[b]{0.19\textwidth}
		\centering
		\includegraphics[width=0.45\textwidth]{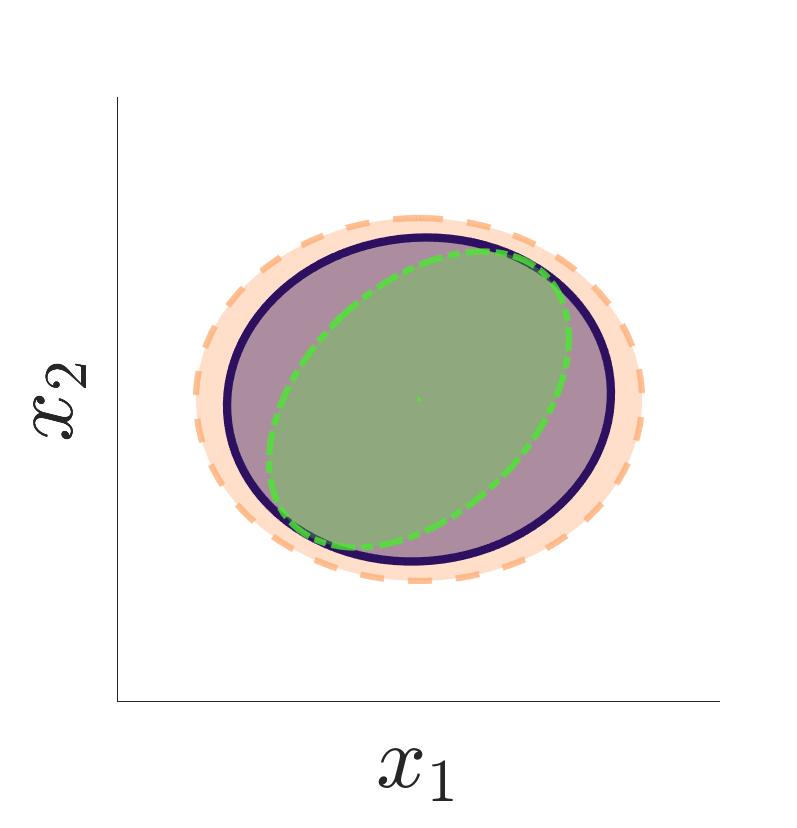}
		\includegraphics[width=0.45\textwidth]{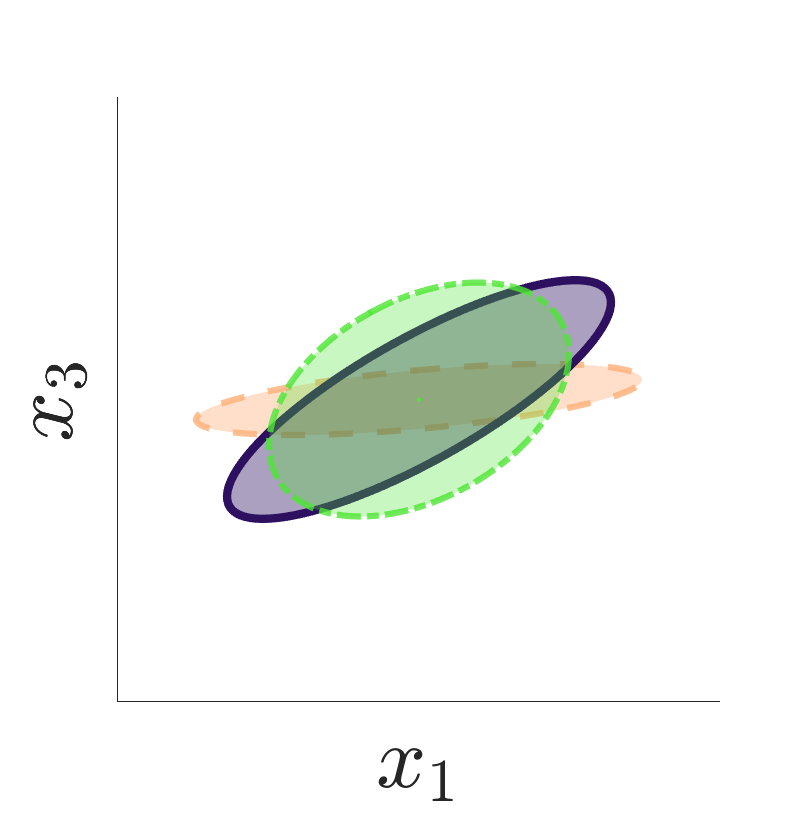}
		\includegraphics[width=0.45\textwidth]{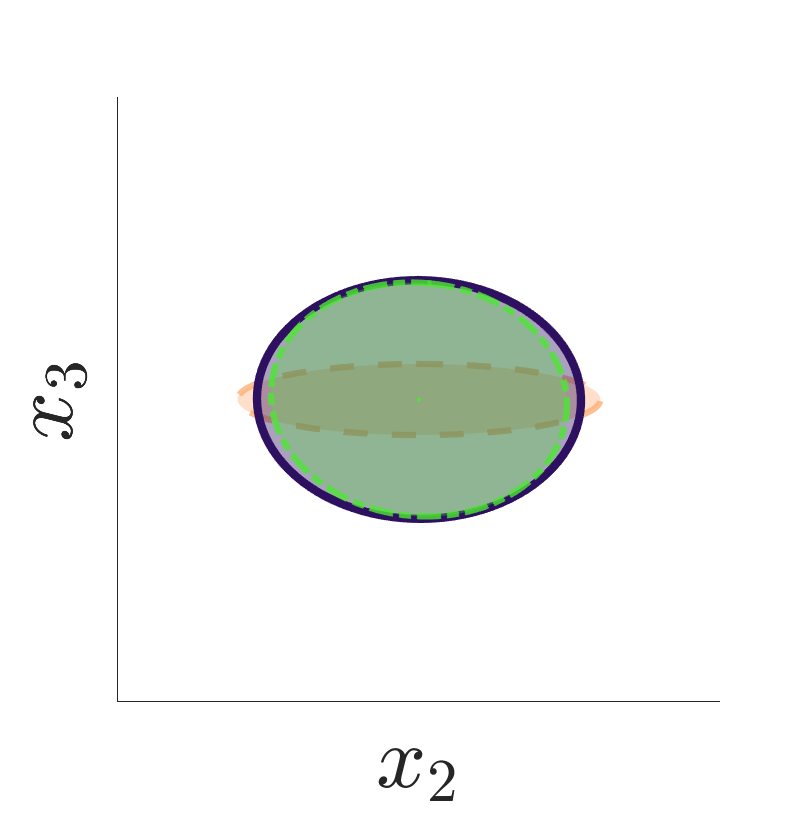}
		\includegraphics[width=0.45\textwidth]{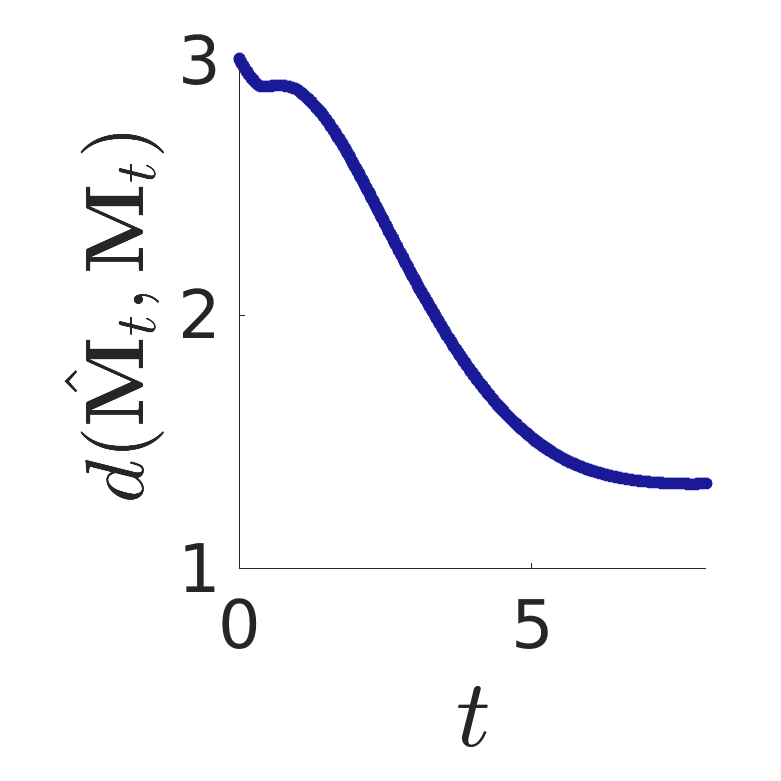}
		\caption{}
		\label{subFig:NaoManip}
	\end{subfigure}
	\caption{COM manipulability tracking with NAO in simulation. \emph{(a)} and \emph{(b)} show the initial and final pose of NAO, respectively. The CoM of the robot is depicted by a red sphere. \emph{(c)} Initial, final and desired manipulability ellipsoids respectively depicted in yellow, dark purple and green. The \emph{bottom-right} graph shows the distance between the current and desired manipulability over time (given in seconds).}
	\label{Fig:Nao}
\end{figure}

We ran several experiments for testing the manipulability tracking at the CoM of the Centauro (Fig.~\ref{Fig:Centauro}) and NAO (Fig.~\ref{Fig:Nao}) robots using the controller~\eqref{Eq:CoMtracking}. The tests consisted of manually setting a desired manipulability ellipsoid to be tracked at the CoM of the robot, and running a joint velocity controller given the reference provided by~\eqref{Eq:CoMtracking}. Notably, both Centauro and NAO tracked the desired manipulability as precisely as possible without compromising the balancing task. Figures~\ref{subFig:CentauroManip} and~\ref{subFig:NaoManip} show the distance between the desired and current CoM manipulability, which decreases over time as the robot adapts its posture to carry out a good tracking while keeping its balance. An interesting aspect about defining and tracking CoM manipulability ellipsoids is the final posture that the robots achieve. Figure~\ref{subFig:CentauroPoseFinal} shows the final posture achieved by Centauro when tracking a CoM manipulability whose projection on the $(x_1,x_2)$ plane is a tilted ellipse, which makes the robot adopt a posture where the front legs and torso rotate on the same plane (which corresponds to the floor in the virtual environment). The final posture of NAO displayed in Fig.~\ref{subFig:NaoPoseFinal} shows that both arms are completely extended along the humanoid frontal axis, in an attempt to align them with one of the main axis of the CoM manipulability ellipsoid. However, both the balancing task and the lower number of DoF constrain NAO to closely match the desired manipulability.

\subsection{Manipulability transfer between robots for a bimanual task}
\label{subsec:Exper_Transfer}
The performance of the proposed manipulability transfer framework was tested in a bimanual unplugging of an electric cable from a power socket. The central idea is to teach different dual-arm robots to execute a task requiring a specific manipulability profile via kinesthetic teaching provided only to one of the bimanual robots.

\begin{figure}[tbp]
	\centering
	\begin{subfigure}[b]{0.48\textwidth}
		\centering
		\includegraphics[width=.3\textwidth]{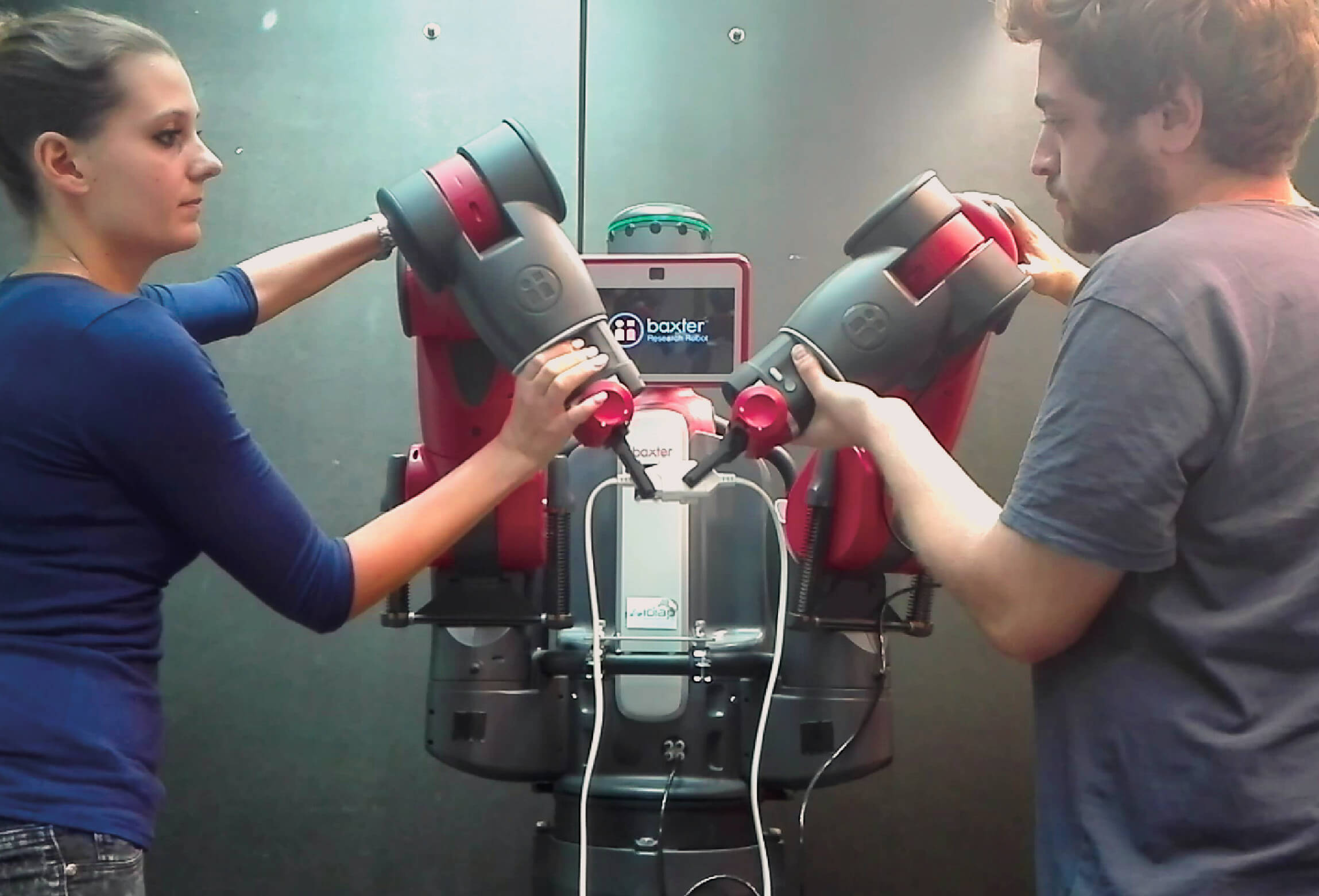}
		\includegraphics[width=.3\textwidth]{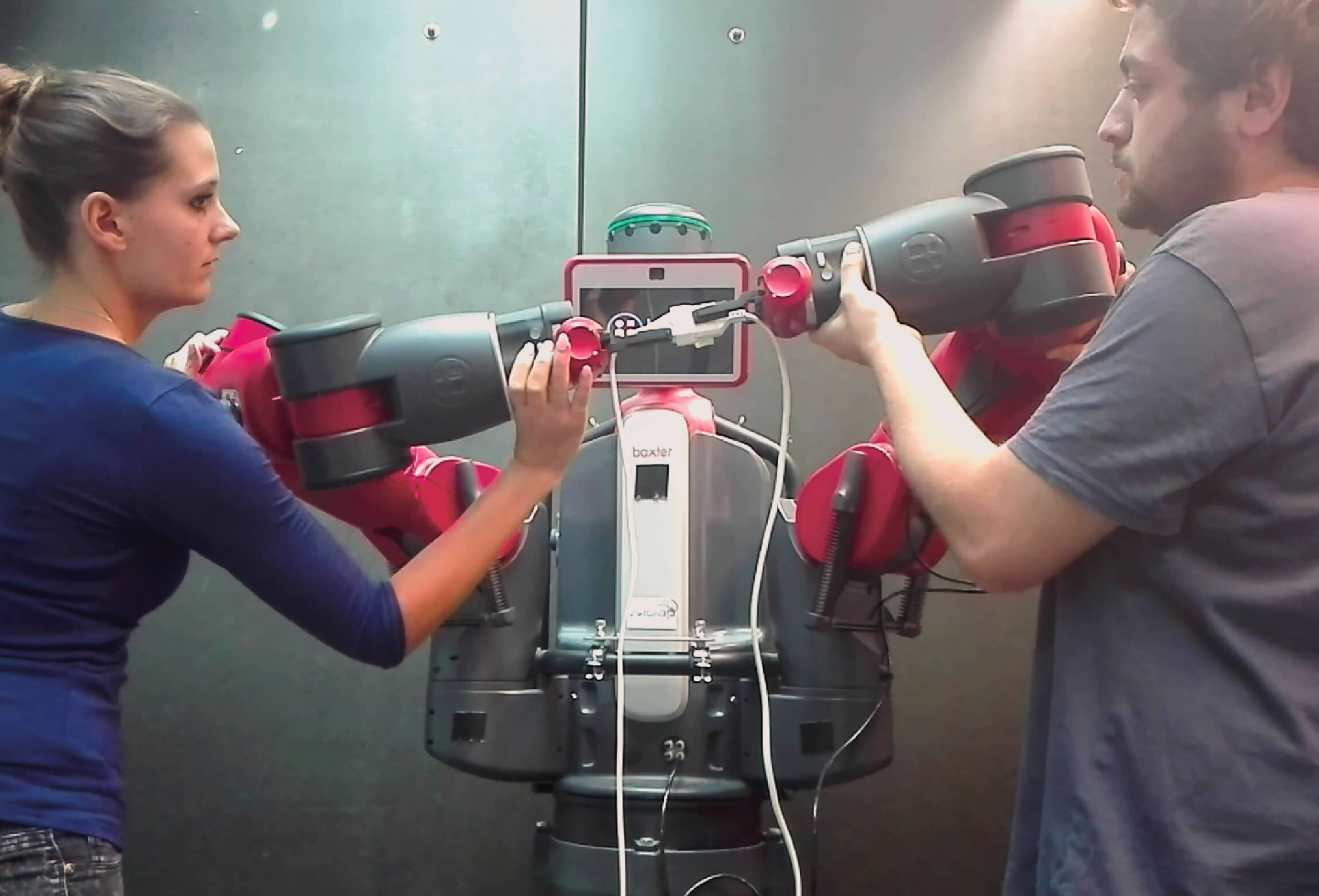}
		\includegraphics[width=.3\textwidth]{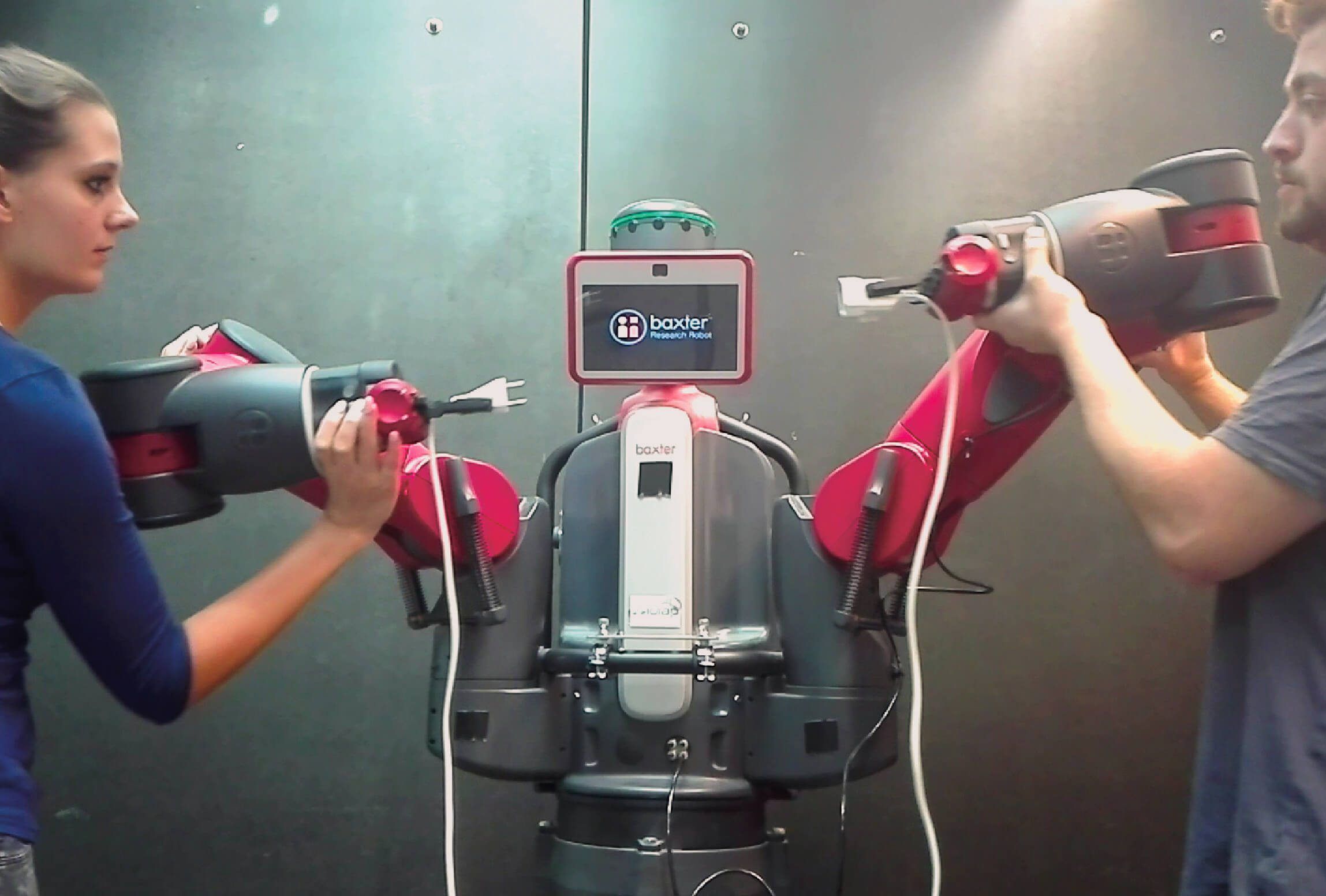}
		\caption{Demonstrations provided by the user on the Baxter robot}
		\label{subFig:ManipTransferPhotosDemos}
	\end{subfigure}
	\begin{subfigure}[b]{0.48\textwidth}
		\centering
		\includegraphics[width=.3\textwidth]{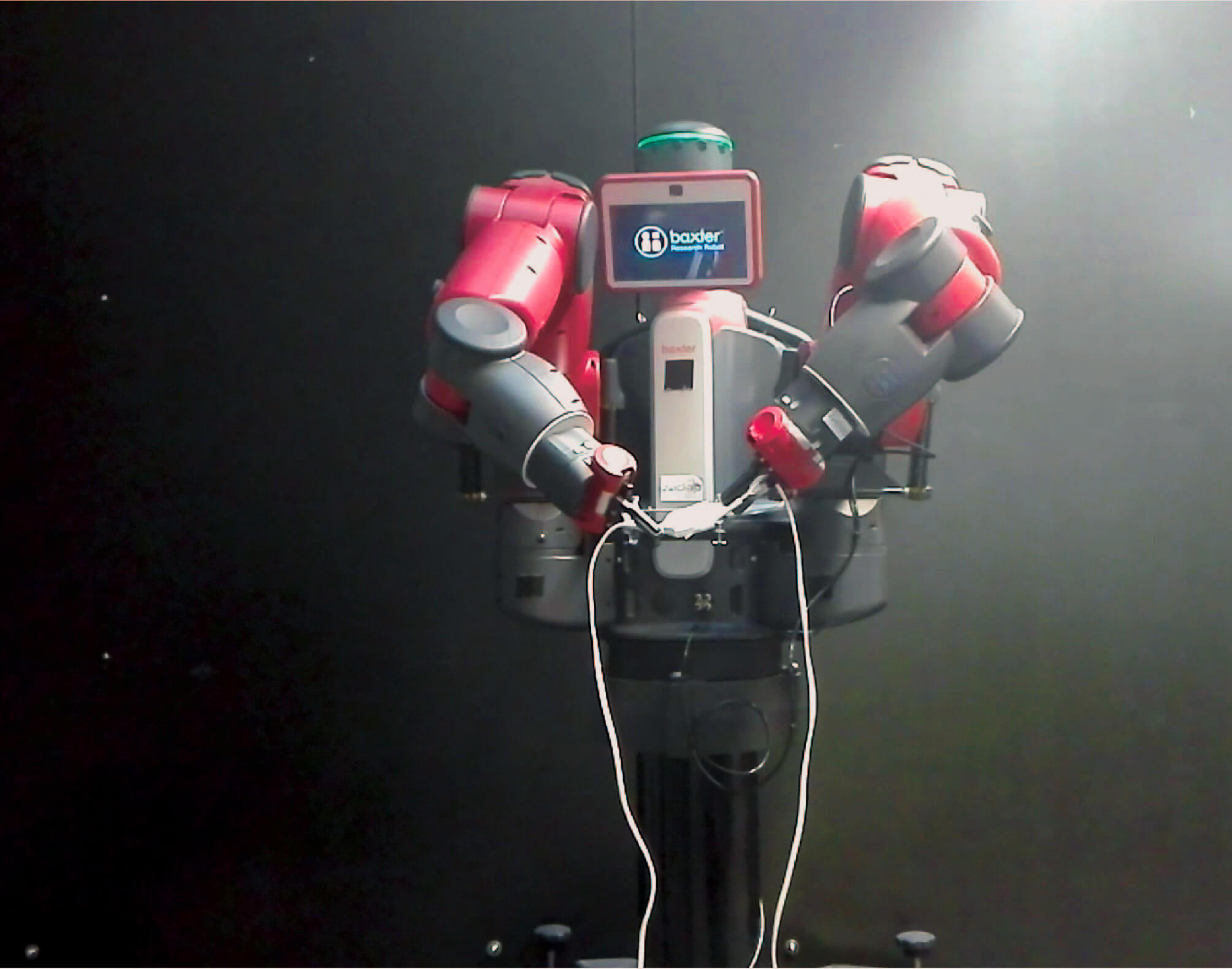}
		\includegraphics[width=.3\textwidth]{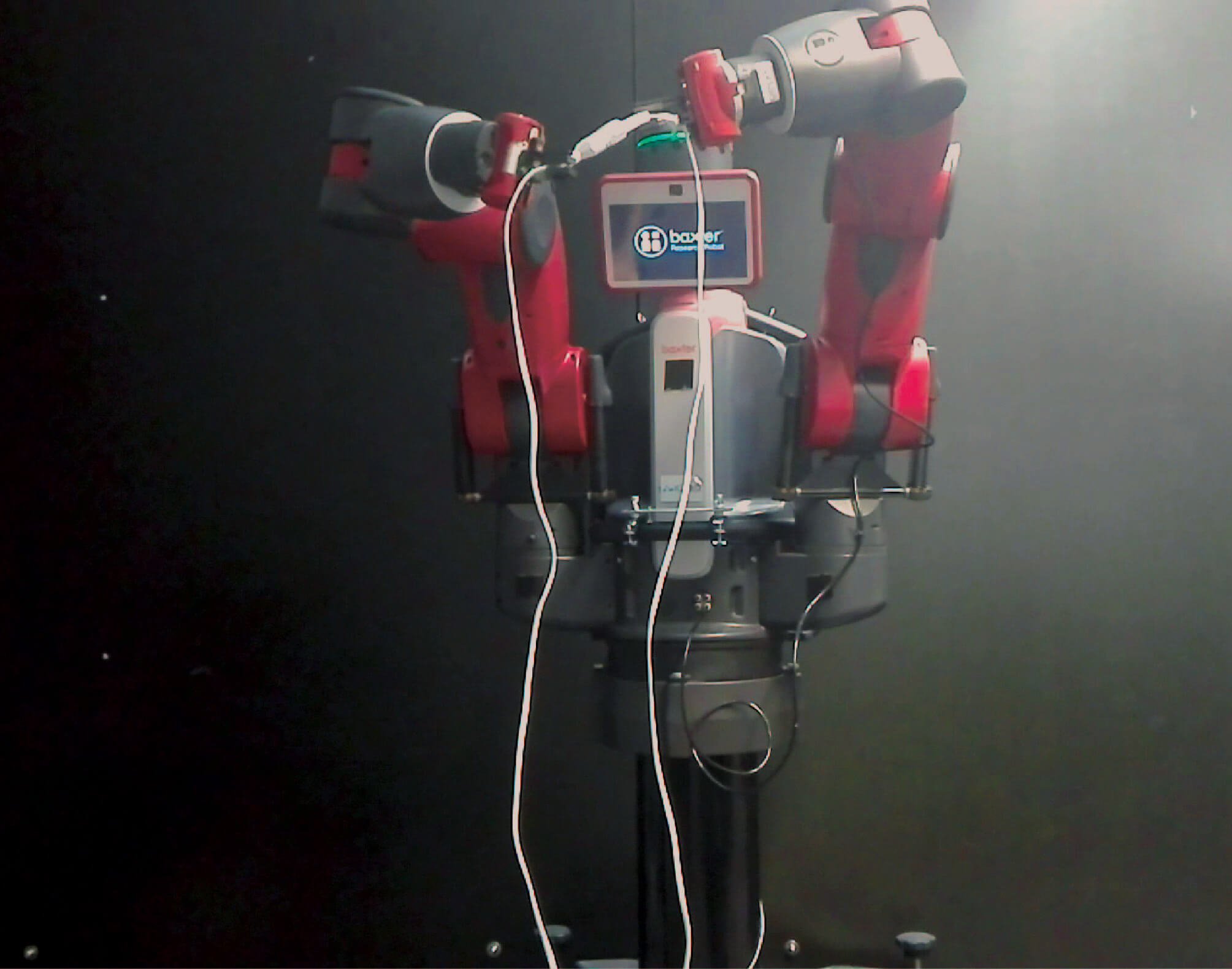}			\includegraphics[width=.3\textwidth]{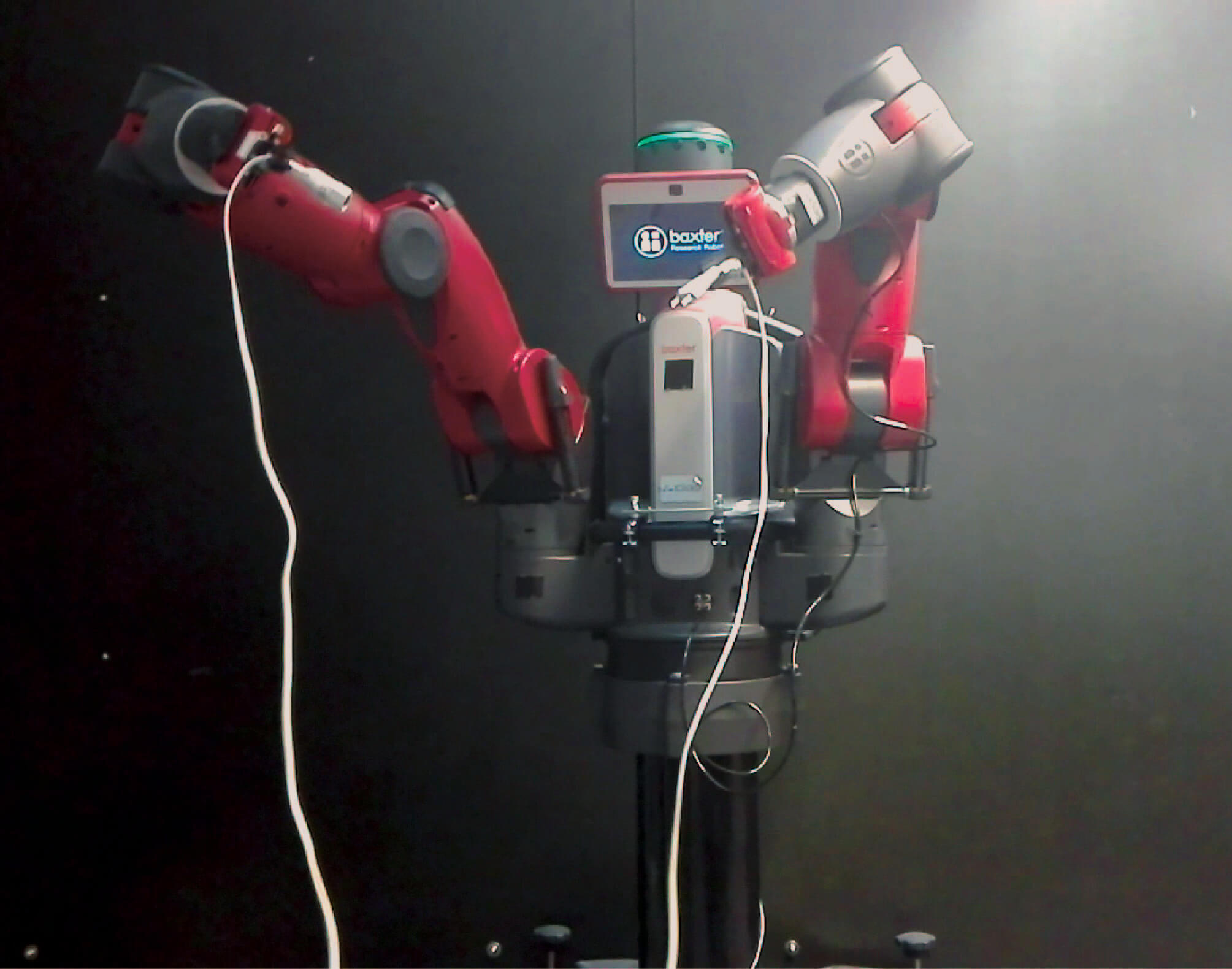}
		\caption{Reproduction by the Baxter robot}
		\label{subFig:ManipTransferPhotosReproBaxter}
	\end{subfigure}
	\begin{subfigure}[b]{0.48\textwidth}
		\centering
		\includegraphics[width=.3\textwidth]{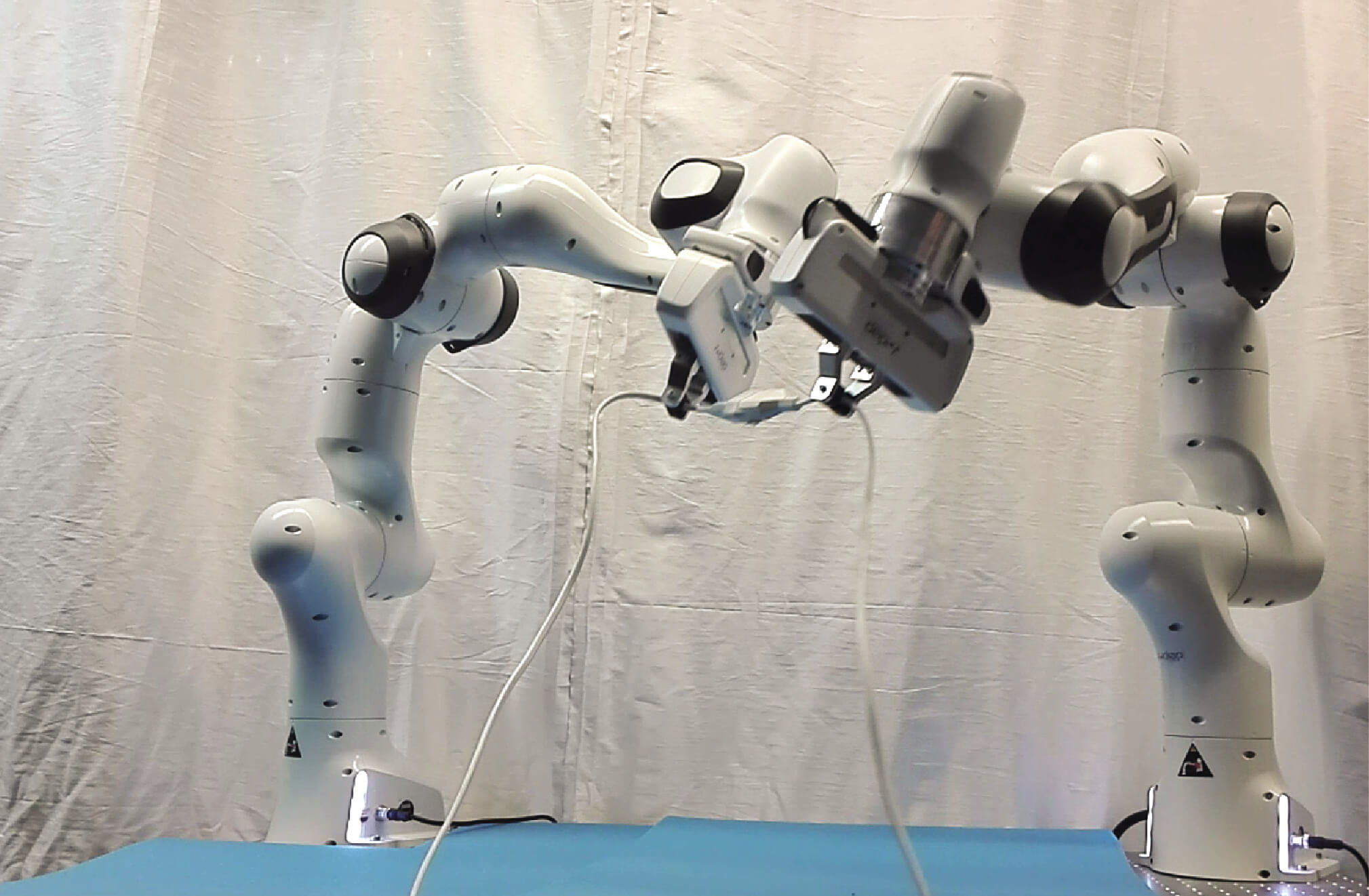}
		\includegraphics[width=.3\textwidth]{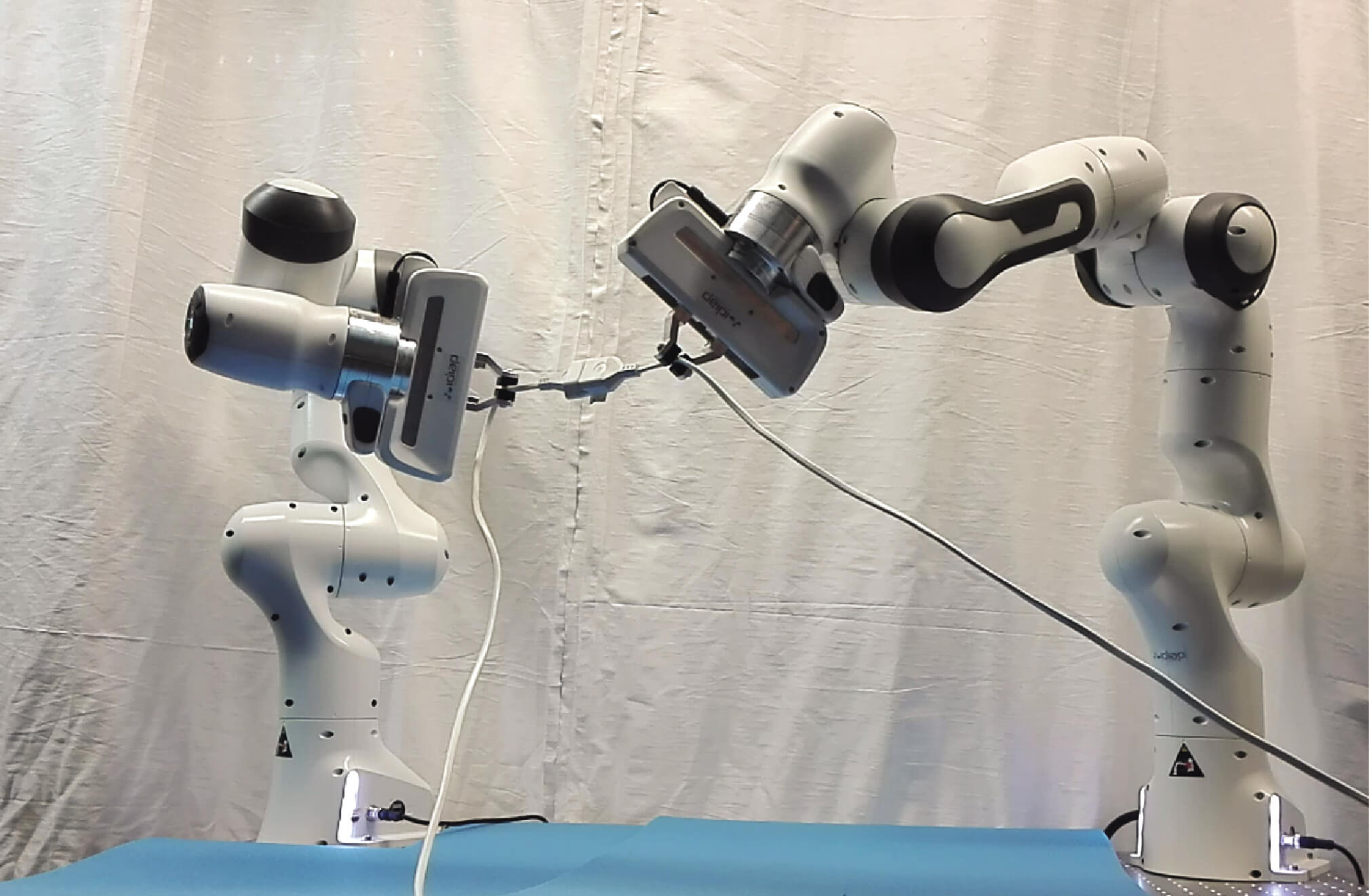}			\includegraphics[width=.3\textwidth]{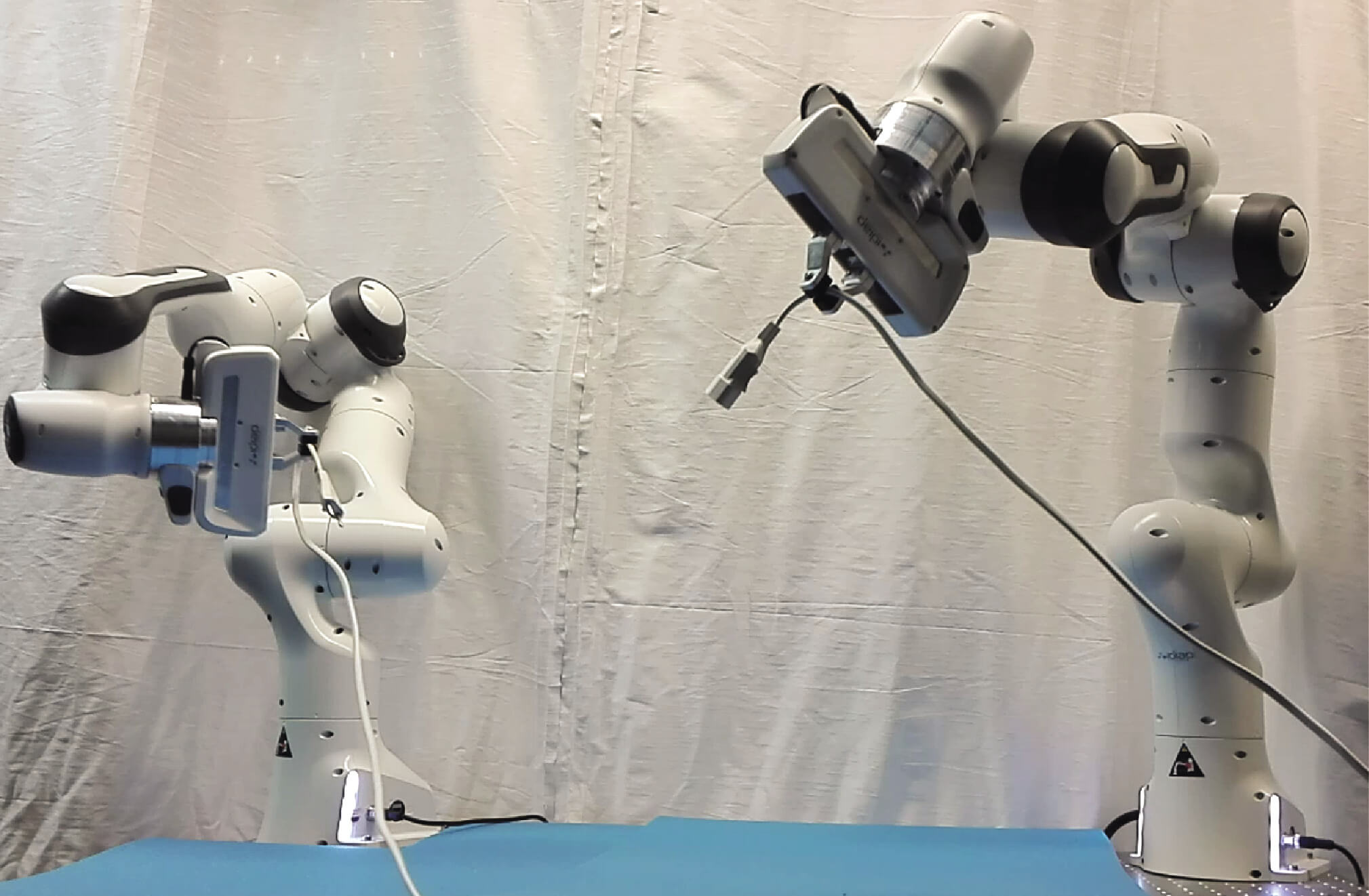}
		\caption{Reproduction by the two Franka Emika Panda robots}
		\label{subFig:ManipTransferPhotosReproFranka}
	\end{subfigure}
	\caption{Unplugging task. The robots pose at the beginning of the task, before and after the extraction of the cable from the socket are respectively shown in the \emph{left}, \emph{middle} and \emph{right} column.}
	\label{Fig:ManipTransferPhotos}
\end{figure}

In the first part of the experiment, the two 7-DoF arms of a Baxter robot are kinesthetically guided to provide demonstrations (see Fig.~\ref{subFig:ManipTransferPhotosDemos}). The posture of the arms is modified by the user so that the main axis of the dual force manipulability ellipsoid of the system $\bm{M}^{\bm{F}_a} = (\bm{G}_a^{\dagger\trsp}\bm{J}_a\bm{J}_a^\trsp\bm{G}_a^{\dagger})^{-1}$ is aligned with the direction of extraction. Then, the arms are moved in opposite directions to unplug the electric cable from the socket. We extracted both the relative position $\bm{\Delta x}_t$ between the end-effectors of both arms and the force manipulability ellipsoid of the system $\bm{M}^{\bm{F}}_{a,t}$. The collected data were time-aligned and split in two datasets of time-driven trajectories, namely relative Cartesian positions and manipulability. We trained a classical GMM over the time-driven relative positions and a geometry-aware GMM over the time-driven manipulability ellipsoids. The number of components of each model ($K=4$) was selected by the experimenter. 

In the second part of the experiment, the unplugging task is reproduced by both the Baxter robot and a pair of Franka Emika Panda robots (see Fig.~\ref{subFig:ManipTransferPhotosReproBaxter}, ~\ref{subFig:ManipTransferPhotosReproFranka}). For both reproductions, the relative position between the end-effectors and the desired manipulability of the system were computed at each time step by a classical GMR as $\bm{\hat{\Delta x}}_t \sim p(\bm{\Delta x}|t)$ and a geometry-aware GMR as $\bm{\hat{M}}^{\bm{F}}_{a,t} \sim P(\bm{M}_a^{\bm{F}}|t)$. In both cases, the left robotic arm was required to move its joints to track the desired manipulability ellipsoid \eqref{Eq:ManipTrackFirstTask}, while the right arm was required to maintain the desired relative Cartesian position with respect to the left arm, while tracking the desired manipulability as secondary objective \eqref{Eq:ManipTrackSecTask}. Note that the actuation contribution of each robot was taken into account to compute the manipulability ellipsoids through the whole experiment.

\begin{figure*}[tbp]
	\centering
	\begin{subfigure}[b]{0.32\textwidth}
		\centering
		\includegraphics[width=\textwidth]{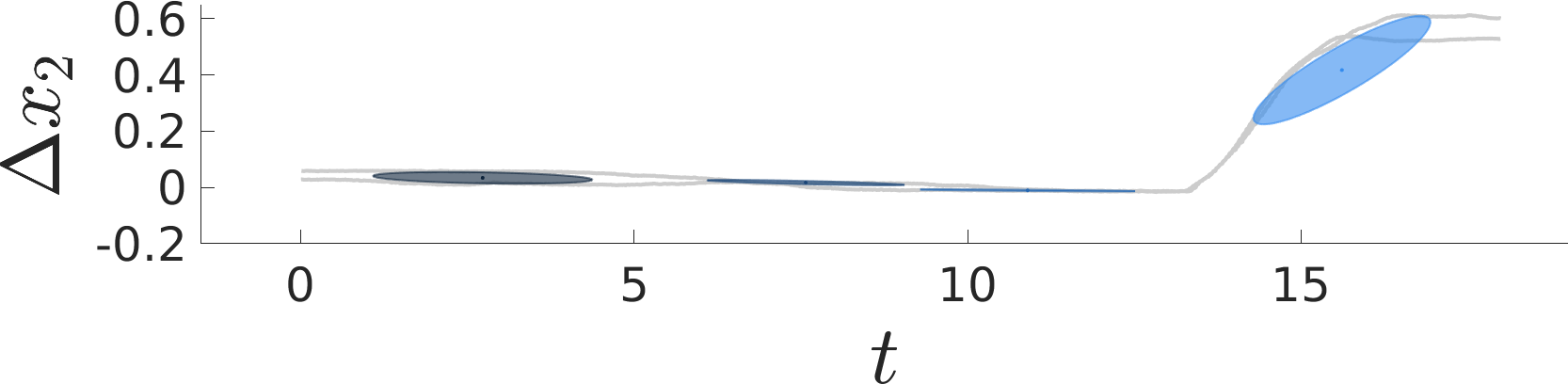}
		\includegraphics[width=\textwidth]{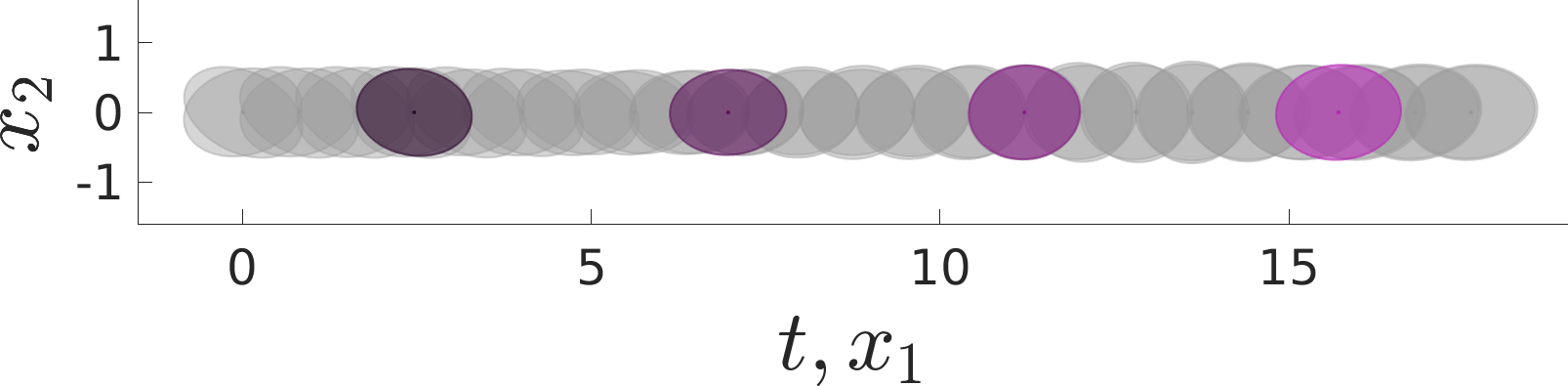}
		\includegraphics[width=\textwidth]{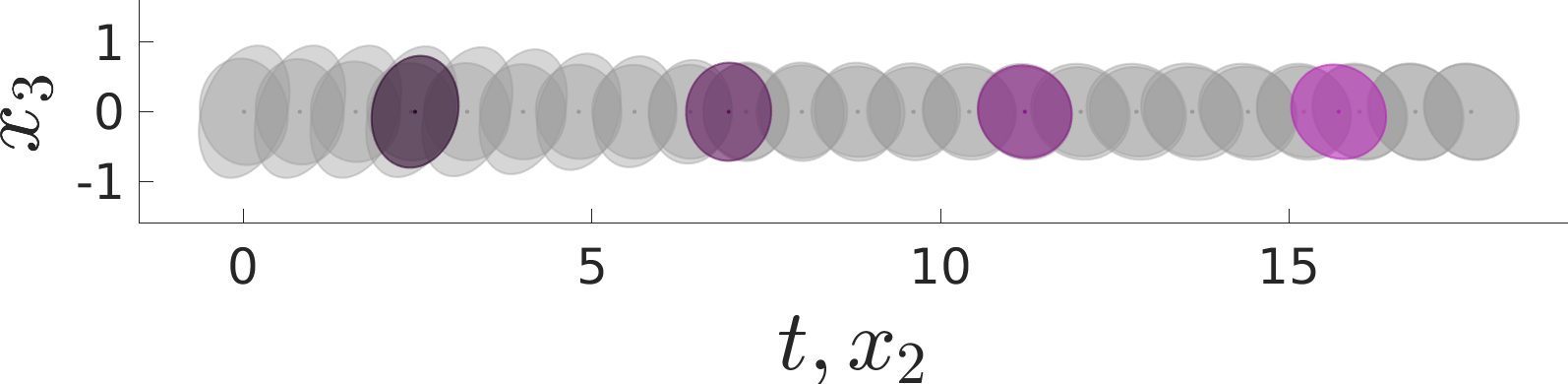}
		\caption{Demonstrations and GMM}
		\label{Fig:ManipTransferExpGMM}
	\end{subfigure}
	\begin{subfigure}[b]{0.32\textwidth}
		\centering
		\includegraphics[width=\textwidth]{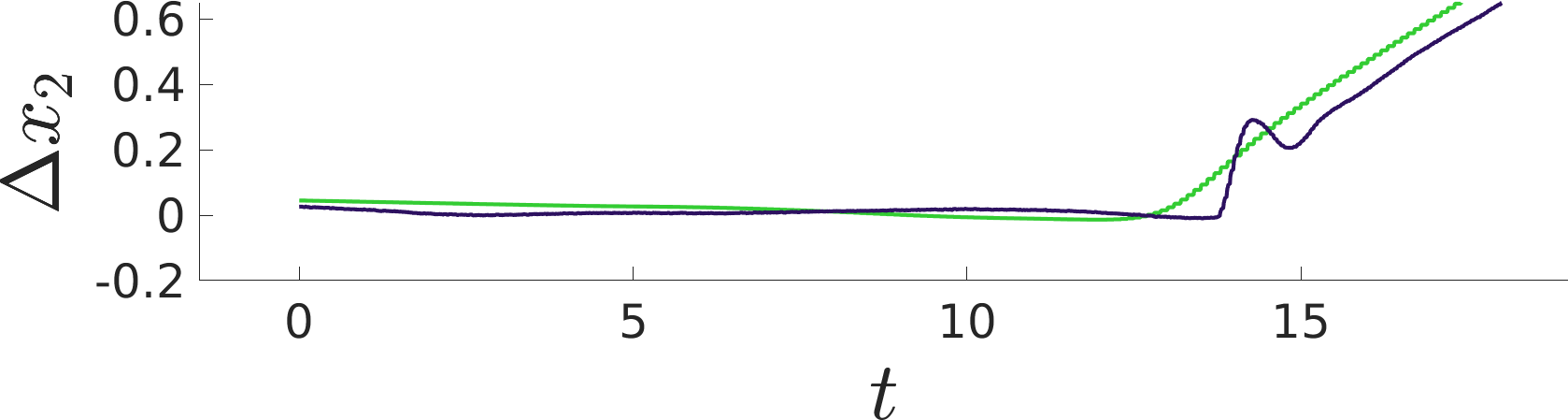}
		\includegraphics[width=\textwidth]{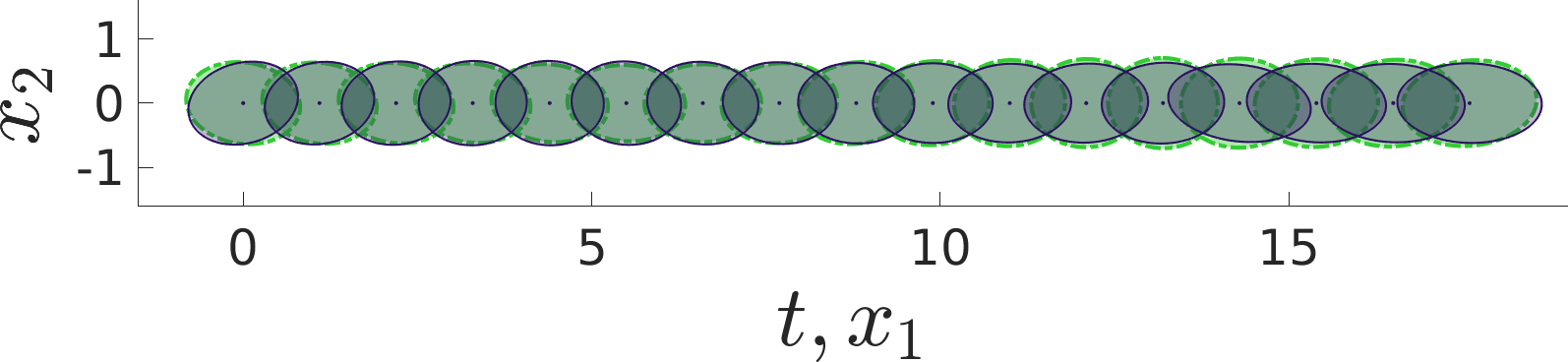}
		\includegraphics[width=\textwidth]{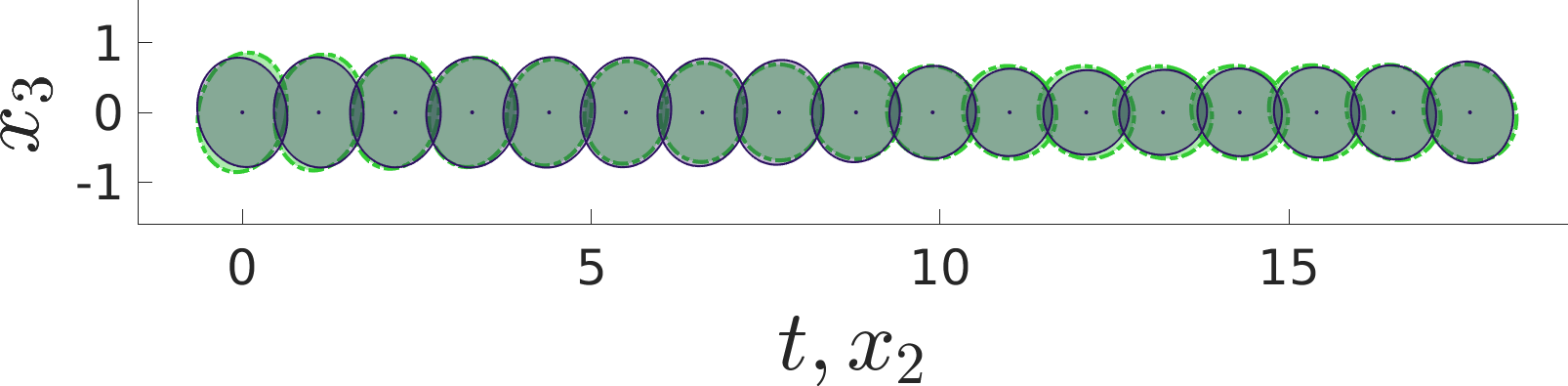}
		\caption{Reproduction with Baxter}
		\label{Fig:ManipTransferReproBaxter}
	\end{subfigure}
	\begin{subfigure}[b]{0.32\textwidth}
		\centering
		\includegraphics[width=\textwidth]{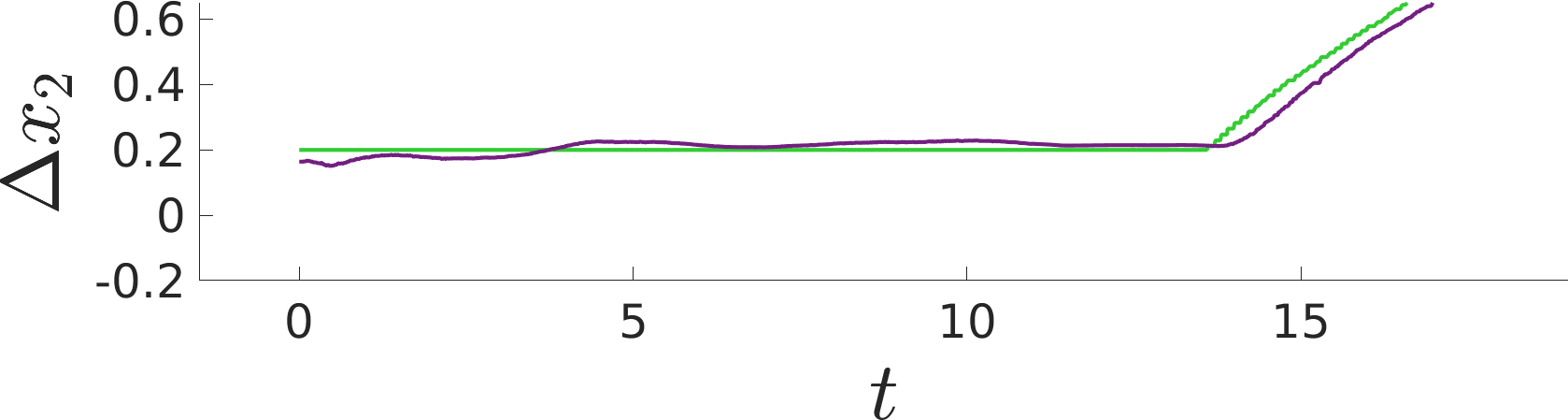}
		\includegraphics[width=\textwidth]{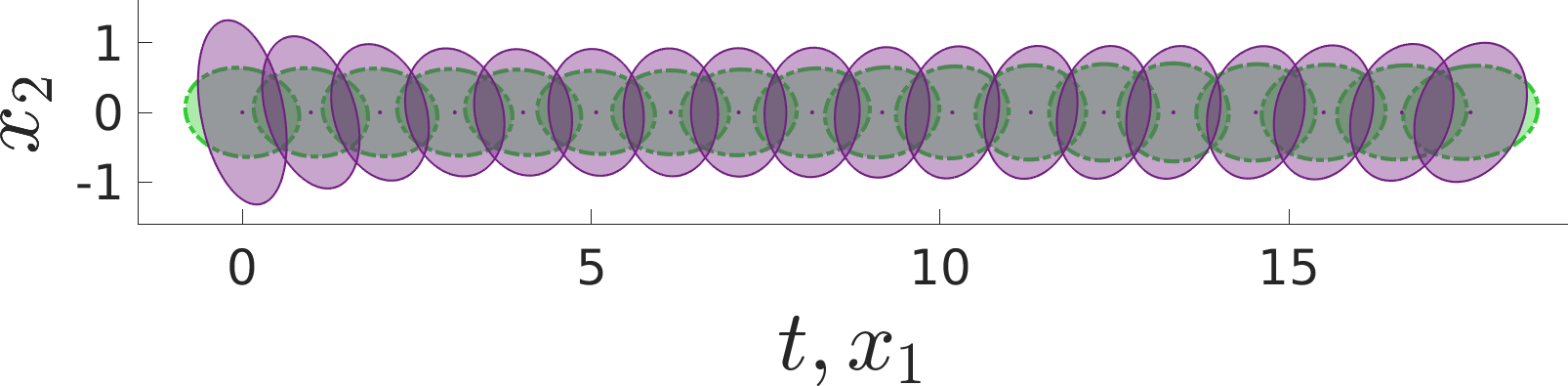}			\includegraphics[width=\textwidth]{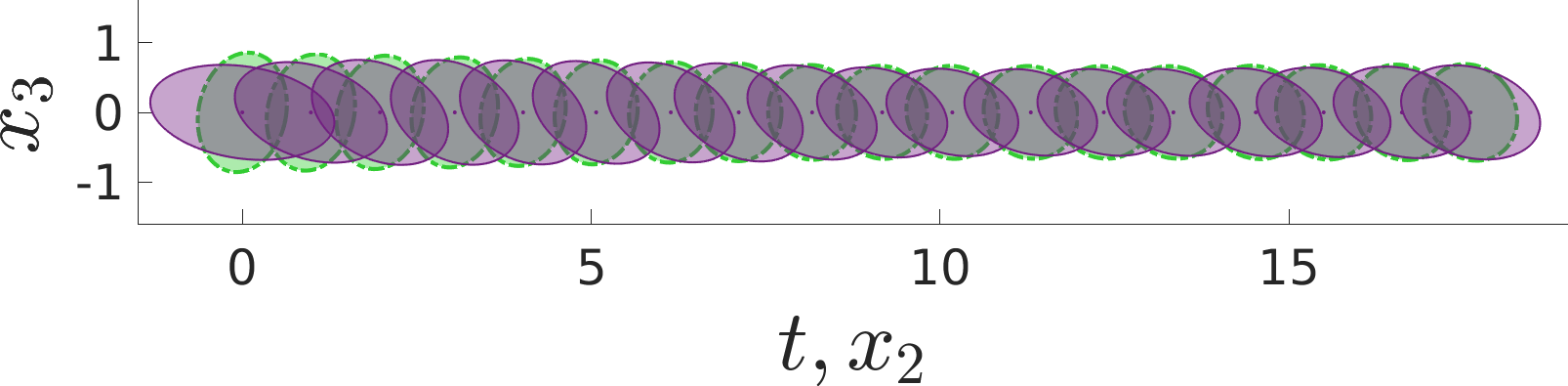}
		\caption{Reproduction with the Panda robots}
		\label{Fig:ManipTransferReproFranka}
	\end{subfigure}
	\caption{\textbf{(\emph{a}) Demonstrations and GMM encoding the unplugging task.} The \emph{top} graph shows the demonstrated relative end-effectors position for the Baxter robot (in gray) and components of the 4-states GMM (in blue). Only the most representative dimension is displayed. The distance between the two arms increases when the cable is unplugged from the socket. The \emph{middle-bottom} graphs show the demonstrated force manipulability profile (in gray) and centers of the 4-states GMM in the SPD manifold over time (in purple). \textbf{(\emph{b}) Reproduction of the unplugging task with Baxter.} The desired and reproduced trajectories are represented in green and dark blue respectively. The \emph{top} graph shows the desired and reproduced relative position between the end-effectors along the second dimension. The \emph{middle-bottom} graphs show the desired and reproduced (overlapping) manipulability ellipsoids. \textbf{(\emph{c}) Reproduction of the unplugging task with the two Panda robots.} The desired and reproduced trajectories are represented in green and purple respectively. The \emph{top} graph shows the desired and reproduced relative position between the end-effectors along the second dimension. The \emph{middle-bottom} graphs shows the desired and reproduced manipulability ellipsoids. The position $\bm{x}$ and time $t$ are given in meters and seconds.}
	\label{Fig:ManipTransfer}
\end{figure*}

Figure~\ref{Fig:ManipTransferExpGMM} displays the two demonstrations recorded by kinesthetically guiding the Baxter robot along with the components of the GMM encoding $\bm{\Delta x}_t$ and the centers of the components of the geometry-aware GMM encoding $\bm{M}^{\bm{F}}_a$. The first and third dimensions of $\bm{\Delta x}_t$ are not represented as they do not vary significantly during the experiment. Figure~\ref{Fig:ManipTransferReproBaxter} shows the relative Cartesian position and manipulability ellipsoid profile to be tracked and the reproduction results when the Baxter robot executed the task. Baxter successfully tracked the desired manipulability ellipsoid while maintaining the required relative distance between its end-effectors. 

Figure~\ref{Fig:ManipTransferReproFranka} shows the relative Cartesian position between the arms and the manipulability ellipsoid profile obtained during the reproduction of the task by the two Panda robots. These successfully achieved the required task and tracked the desired manipulability ellipsoid profile obtained from model trained with the data recorded on the Baxter robot. 
Note the manipulability matching is not exact in this case due to the differences between Baxter and the Panda robots. Indeed, even if the actuation capabilities of each robot are taken into account in our manipulability transfer framework, the capabilities of the two dual-arm system differ due to other physical specificities, e.g. the relative position of the bases of the arms.

\section{Discussion}
\label{sec:Discuss}
Our tracking formulation enables robots to modify their posture in an exponentially stable way so that desired manipulability ellipsoids are tracked, either as a main control task or as a redundancy resolution problem where the manipulability tracking is considered a secondary objective. 
Compared to state-of-the-art manipulability-based optimization schemes, our tracking formulation allows the reproduction of any manipulability ellipsoid beyond the maximization of manipulability parameters.
The proposed tracking approach covers different manipulability ellipsoids proposed in the literature, such as velocity, force and dynamic manipulability ellipsoids~\citep{Doty95:Manipulabilty}. A relevant aspect about our approach is their generic structure, which means that we can track manipulability ellipsoids for a large variety of robots, as reported in the previous section, where a robotic hand, a Centauro robot, a humanoid and two different bimanual setups were used to test our tracking approach. This shows that our approach can be used in a large variety of contexts and that many further applications can be considered. 

The manipulability transfer results reported in Section \ref{subsec:Exper_Transfer} showed the effectiveness of the proposed approach for transferring manipulability ellipsoids between robots that differ in their kinematic structure, which has remained a challenge in the robot learning community. Our learning framework allows a robot to learn posture-dependent task requirements without explicitly encoding a model in the joint space of the demonstrator, which would require complex kinematic mapping algorithms and would make task analysis less interpretable at first sight. In addition, the proposed framework extends the robot learning capabilities beyond the transfer of trajectory, force and impedance.

It is important to emphasize the fact that the manipulability tracking precision strongly depends on the number of DoFs when the task is considered a secondary objective, as the higher it is, the more capable the robot is to perform more than one task simultaneously. Note that, in the case of legged robots (which are often characterized by a high number of DoFs), the manipulability tracking may still be slightly compromised because of the set of constraints imposed by the balancing task, as observed in Section \ref{subsec:Exper_COM}. However, if these robots are provided with the possibility of modifying their feet position while keeping balance, then the manipulability tracking may be further improved. This clearly requires more sophisticated balancing controllers, but gives robots more freedom to adapt their posture and achieve better manipulability tracking. Notice that in the case of robotic hands, a similar behavior arises when the finger tips are constrained according to some grasping requirements, which might affect the manipulability tracking when projected into the nullspace of the primary task.

It is important to notice that the proposed manipulability tracking approach is a local method in the sense that the solution depends on the current configuration of the robot expressed through the Jacobian. This makes the tracking convergence dependent on the current configuration of the robot, which sometimes may limit the tracking performance. However, the robot may achieve a better tracking performance if it is allowed to look for other initial postures. As an example, the robot may not track precisely the desired manipulability ellipsoids for a given initial posture, due for instance to its joint limits. However, if the robot slightly modifies its initial posture, it may find a better starting configuration to subsequently minimize the error between the desired and current manipulability ellipsoids in a larger proportion, even if the new initial posture initially increases this error.

Overall, the proposed manipulability transfer framework may be exploited in a large variety of applications, where the posture of the robot may have an impact on its performance while executing the task. In addition to varying the robot posture for task compatibility, tracking a desired manipulability profile as a secondary task may typically complement a main control task to avoid singularity, handle perturbations during task execution, optimize the execution time or minimize the energy consumption~\citep{Kim10:ManipGait}. 
In particular, manipulability transfer may be utilized from a motion planning point of view. To do so, the robot may first track a desired manipulability as main control task in a planning phase, where the robot adapts its posture in order to anticipate the next action. Following this planning phase, the robot executes the desired action with a posture adapted to the task requirements. In this phase, the desired manipulability is tracked as a secondary task.
Moreover, in the context of rehabilitation and assistance, the proposed learning and tracking formulations may be exploited in  control strategies for exoskeletons. In~\citep{Petric19:ExoskeletonMuscManip}, the exoskeleton posture is optimized to achieve an isotropic manipulability by sensing the human muscular manipulability. In this setting, a varying exoskeleton manipulability profile may be retrieved using GMR as a function of the sensed muscular manipulability. 

From a mathematical point of view, it is worth highlighting the importance of considering the structure of the data we work with. While alternative solutions to handle SPD matrices are present in literature (e.g. those using Cholesky decomposition), we showed that Euclidean manipulability-tracking controllers lead to unstable behaviors in contrast to the stable behavior displayed by our geometry-aware controller. Equally important, the manipulability ellipsoids profiles retrieved by the geometry-aware and Euclidean GMR were similar only around the mean of the GMM components, but diverged when moving away from it. This is because the estimated output in Euclidean space is only a valid approximation for input data lying close to the mean, as reported in Section~\ref{sec:GeomImportance}. Therefore, geometry-awareness is crucial for successful learning and tracking of manipulability ellipsoids. 

\section{Conclusions and Future Work}
\label{sec:ConclFut}
This article presented a novel framework for transferring manipulability ellipsoids to robots. The proposed approach is built on a probabilistic learning model that allows the encoding and retrieval manipulability ellipsoids, and on the extension of the classical inverse kinematics problem to manipulability ellipsoids, by establishing a mapping between a change of manipulability ellipsoid and the robot joint velocity. We exploited tensor representation and Riemannian manifolds to build a geometry-aware learning framework and exponentially stable tracking controllers and showed the importance of geometry-awareness for manipulability transfer. We then showed that our manipulability transfer framework allows the exploitation of task variations recovered by the learning approach to characterize the precision of the manipulability tracking problem. 
This approach enables the learning of posture-dependent task requirements. It provides a skill transfer strategy going beyond the imitation of trajectory, force or impedance behaviors. Furthermore, it allows manipulability transfer between agents of different embodiments, while taking into account their individual characteristics and is adapted to complex scenarios involving any manipulability ellipsoid shape and various types of robots.

Future work will explore manipulability transfer between humans and robots. Following this research direction, we recently proposed a statistical analysis of single and dual-arm manipulability ellipsoids for human movements, accompanied by two human-to-robot manipulability transfer experiments. The corresponding results will be detailed in a forthcoming publication. We will also investigate manipulability transfer strategies where the desired manipulability would be optimized in function of the robot. The objective would be to adapt the manipulability ellipsoid to exploit the capabilities of the learner in situations in which this learner can reach a better manipulability than the teacher for the task at hand.

\begin{acks}
This work was supported by the Swiss National Science Foundation (SNSF/DFG project TACT-HAND).
\end{acks}

\bibliographystyle{SageH}
\bibliography{References}

\begin{appendices}
\section{Derivative of a matrix w.r.t. a vector}
\label{sec:AppendixProofsDerivatives}
\paragraph{Left multiplication by a constant matrix (Eq.~\eqref{Eq:LeftMultDeriv})}
\begin{equation*}
\frac{\partial \bm{AY}}{\partial \bm{x}} = \frac{\partial \bm{Y}}{\partial \bm{x}} \times_1 \bm{A}
\end{equation*} 
\begin{proof}
	\begin{equation*}
	\left(\frac{\partial \bm{AY}}{\partial \bm{x}}\right)_{ljk} = \frac{\partial}{\partial x_k} \sum_{i}a_{li}y_{ij} = \sum_{i}a_{li}\frac{\partial y_{ij}}{\partial x_k}
	\end{equation*} 
\end{proof}
\paragraph{Right multiplication by a constant matrix (Eq.~\eqref{Eq:RightMultDeriv})}
\begin{equation*}
\frac{\partial \bm{YB}}{\partial \bm{x}} = \frac{\partial \bm{Y}}{\partial \bm{x}} \times_2 \bm{B}^\trsp
\end{equation*} 
\begin{proof}
	\begin{equation*}
	\left(\frac{\partial \bm{YB}}{\partial \bm{x}}\right)_{ilk} = \frac{\partial}{\partial x_k} \sum_{i}y_{ij}b_{jl} = \sum_{j}b_{jl}\frac{\partial y_{ij}}{\partial x_k}
	\end{equation*}
\end{proof}

\paragraph{Derivative of the inverse of a matrix (Eq.~\eqref{Eq:InvDeriv})}
\begin{equation*}
\frac{\partial \bm{Y}^{-1}}{\partial \bm{x}} = - \frac{\partial \bm{Y}}{\partial \bm{x}}^\trsp \times_1 \bm{Y}^{-1} \times_2 \bm{Y}^{-\trsp}
\end{equation*}
\begin{proof}
	We compute the derivative of the definition of the inverse $\bm{Y}^{-1}\bm{Y}=\bm{I}$ as
	\begin{equation*}
	\frac{\partial}{\partial \bm{x}} (\bm{Y}^{-1}\bm{Y})= \frac{\partial}{\partial \bm{x}}(\bm{I}),
	\end{equation*}
	\begin{equation*}
	\frac{\partial\bm{Y}^{-1}}{\partial \bm{x}} \times_2 \bm{Y}^\trsp + \frac{\partial\bm{Y}}{\partial \bm{x}} \times_1 \bm{Y}^{-1} = \bm{0}.
	\end{equation*}
	Then, by isolating $\frac{\partial\bm{Y}^{-1}}{\partial \bm{x}}$, we obtain
	\begin{equation*}
	\frac{\partial\bm{Y}^{-1}}{\partial \bm{x}} = -\frac{\partial\bm{Y}}{\partial \bm{x}}^\trsp \times_1 \bm{Y}^{-1}\times_2 \bm{Y}^{-\trsp}.
	\end{equation*}
\end{proof}
	
\section{Symbolic manipulability Jacobian for a serial kinematic chain}
\label{sec:AppendixManipJacobian}
The computation of the manipulability Jacobian involves computing the derivative of the robot Jacobian w.r.t. the joint angles. Those derivatives can be computed in a symbolic form as shown in~\citep{Bruyninckx96}. We remind here the symbolic derivative for the hybrid representation of the Jacobian $\bm{J}\in\mathbb{R}^{6\times n}$ that is used in the computation of the manipulability Jacobian $\bm{\mathcal{J}}$.

The $i$-th column of the Jacobian is denoted by
\begin{equation}
\bm{J}^i = \left(\begin{matrix}
\bm{w}^i\\
\bm{v}^i\\
\end{matrix}\right),
\label{Eq:JacobianCol}
\end{equation}
with $\bm{w}^i\in\mathbb{R}^3$ and $\bm{v}^i\in\mathbb{R}^3$ the rotational and translational components of the Jacobian.

The derivative of the Jacobian w.r.t. the joint angles is a third order tensor $\frac{\partial\bm{J}}{\partial\bm{q}} \in\mathbb{R}^{6\times n \times n}$ with mode-1 fibers or columns
\begin{equation}
\left(\frac{\partial \bm{J}}{\partial \bm{q}}\right)_{:ij} =\;
\frac{\partial \bm{J}^i}{\partial q^j} \;= \;
\left\{
\begin{array}{ll}
\bm{P}_\Delta(\bm{J}^j)\bm{J}^i &\text{ if } j\leq i\\
- \bm{M}_\Delta(\bm{J}^j)\bm{J}^i &\text{ if } j > i\\
\end{array}
\right. ,
\label{Eq:JacobianJointDerCol}
\end{equation}
where
\begin{equation}
\bm{P}_\Delta (\bm{J}^j) = 
\left( \begin{matrix}
[\bm{w}^j \times] & \bm{0}_{3\times 3} \\
\bm{0}_{3\times 3} & [\bm{w}^j \times]\\
\end{matrix}
\right),
\label{Eq:JacPmatrix}
\end{equation}
\begin{equation}
\bm{M}_\Delta (\bm{J}^j) = 
\left( \begin{matrix}
\bm{0}_{3\times 3} & \bm{0}_{3\times 3} \\
[\bm{v}^j \times] & \bm{0}_{3\times 3}\\
\end{matrix}
\right),
\label{Eq:JacMmatrix}
\end{equation}
and $\times$ the cross product between two vectors. The notation $[\bm{w}^j \times]$ in a matrix denotes that the corresponding component of the result of the right-multiplication of the matrix by a vector is equal to the cross product between $\bm{w}^j$ and the corresponding vector component, e.g. $\bm{P}_\Delta(\bm{J}^j)\bm{J}^i = \left( \begin{matrix}
\bm{w}^j \times \bm{w}^i\\
\bm{w}^j \times \bm{v}^i\\
\end{matrix}\right)$.

Note that the time derivative of the Jacobian can therefore be computed as
\begin{equation}
\frac{d \bm{J}}{dt} = \sum_{j=1}^{n} \frac{\partial \bm{J}}{\partial q^j} \dot{q}_j.
\label{Eq:JacobianTimeDer}
\end{equation}

\section{Symbolic dynamic manipulability Jacobian for a serial kinematic chain}
\label{sec:AppendixDynManipJacobian}
The derivative of the robot inertia matrix w.r.t. joint angles is necessary for the computation of the dynamic manipulability Jacobian. It can be computed in closed form as follows.

The inertia matrix $\bm{\Lambda}(\bm{q})\in\mathbb{R}^{n \times n}$ can be written as
\begin{equation}
\bm{\Lambda}(\bm{q}) = \sum_{i=1}^{n} \bm{J}_i^\trsp 
\left( \begin{matrix}
\bm{\Lambda}_i & \bm{0} \\
\bm{0} & m_i \bm{I} \\
\end{matrix} \right)
\bm{J}_i,
\label{Eq:InertiaMat}
\end{equation}
where $\bm{J}_i$, $\bm{\Lambda}_i$ and $m_i$ are the Jacobian, inertia matrix and mass of link $i$, respectively~\citep{Park95,Murray94}.

The derivative of the inertia matrix is the third order tensor $\frac{\partial\bm{\Lambda}}{\partial\bm{q}}\in\mathbb{R}^{n\times n \times n}$ computed as  
\begin{equation}
\frac{\partial\bm{\Lambda}}{\partial\bm{q}} = \sum_{i=1}^{n} 
\frac{\partial \bm{J}_i^\trsp}{\partial \bm{q}} \times_2
\bm{J}_i^\trsp
\bm{M}_i
+ \frac{\partial \bm{J}_i}{\partial \bm{q}} \times_1
\bm{J}_i^\trsp
\bm{M}_i,
\label{Eq:InertiaMatJointDer}
\end{equation}
where $\bm{M}_i = \left( \begin{matrix}
\bm{\Lambda}_i & \bm{0} \\
\bm{0} & m_i \bm{I} \\
\end{matrix} \right)$ and $\frac{\partial \bm{J}_i}{\partial \bm{q}}$ is computed with Eq.~\eqref{Eq:JacobianJointDerCol}.

%
%

\section{Symbolic derivative of the manipulability Jacobian for a serial kinematic chain}
\label{sec:AppendixDiffManipJacobian}
In some cases, e.g. in the acceleration tracking controller, the time derivative of the manipulability Jacobian is required. This time derivative can be computed symbolically by exploiting the first and second derivative of the Jacobian w.r.t. the joint angles.

The time derivative of the velocity manipulability Jacobian $\bm{\mathcal{J}}^{\bm{\dot{x}}}\in\mathbb{R}^{6\times n \times n}$ defined as
\begin{equation}
\bm{\mathcal{J}}^{\bm{\dot{x}}} = \frac{\partial\bm{J}}{\partial \bm{q}}\times_2\bm{J} + \frac{\partial\bm{J}^\trsp}{\partial \bm{q}}\times_1\bm{J},
\label{Eq:VelocityManipJacobianAppendix}
\end{equation}
is obtained by exploiting the chain rule as
\begin{align}
&\frac{\partial \bm{\mathcal{J}}^{\bm{\dot{x}}}}{\partial t} = \frac{\partial}{\partial t}\left( \frac{\partial \bm{J}}{\partial \bm{q}} \times_2 \bm{J} + \frac{\partial \bm{J}^\trsp}{\partial \bm{q}} \times_1 \bm{J} \right) \\
&= \frac{\partial^2 \bm{J}}{\partial t \partial \bm{q}} \times_2 \bm{J} 
+ \frac{\partial \bm{J}}{\partial \bm{q}} \times_2  \frac{\partial \bm{J}}{\partial t} 
+ \frac{\partial^2 \bm{J}^\trsp}{\partial t \partial \bm{q}} \times_1 \bm{J}
+ \frac{\partial \bm{J}^\trsp}{\partial \bm{q}} \times_1  \frac{\partial \bm{J}}{\partial t}.
\label{Eq:ManipJacobianTimeDer}
\end{align}

The time derivative of the Jacobian is given by Eq.~\eqref{Eq:JacobianTimeDer} and the time derivative of the derivative of the Jacobian w.r.t. joint angles is given by
\begin{equation}
\frac{\partial^2 \bm{J}}{\partial t \partial q_j} = \sum_{k=1}^{n} \frac{\partial^2 \bm{J}}{\partial q_k \partial q_j} \dot{q}_k,
\label{Eq:JacobianTimeJointDer}
\end{equation}
where the second derivative of the Jacobian w.r.t. the joint angles is a fourth order tensor $\frac{\partial^2\bm{J}}{\partial\bm{q}^2} \in\mathbb{R}^{6\times n \times n \times n}$ with mode-1 fibers or columns
\begin{align}
&\left(\frac{\partial^2 \bm{J}}{\partial \bm{q}^2}\right)_{:ijk} =\;
\frac{\partial^2 \bm{J}^i}{\partial q^k \partial q^j} \;= \nonumber \\ 
&\left\{
\begin{small}
\begin{array}{ll}
\big( \bm{P}_\Delta(\bm{J}^j) \bm{P}_\Delta(\bm{J}^k) \big)\bm{J}^i 
+ \bm{P}_\Delta(\bm{J}^j) \big( \bm{P}_\Delta(\bm{J}^k) \bm{J}^i \big)
&\text{ if } k\!\leq \!j \!\leq\! i\\
\bm{P}_\Delta(\bm{J}^j) \big( \bm{P}_\Delta(\bm{J}^k) \bm{J}^i \big)
&\text{ if } j\!\leq\! k\! \leq \!i\\
- \bm{P}_\Delta(\bm{J}^j) \big( \bm{M}_\Delta(\bm{J}^k) \bm{J}^i \big)
&\text{ if } j \!\leq\! i \!<\! k \\
-\big( \bm{P}_\Delta(\bm{J}^k) \bm{M}_\Delta(\bm{J}^j) \big) \bm{J}^i 
- \bm{M}_\Delta(\bm{J}^j) \big( \bm{P}_\Delta(\bm{J}^k) \bm{J}^i \big)
&\text{ if } k\! \leq\! i\! < \!j \\
-\big( \bm{P}_\Delta(\bm{J}^k) \bm{M}_\Delta(\bm{J}^j) \big) \bm{J}^i 
&\text{ if } i\! <\! k \!<\! j \\
- \big( \bm{P}_\Delta(\bm{J}^j) \bm{M}_\Delta(\bm{J}^k) \big) \bm{J}^i 
&\text{ if } i\! < \!j\! \leq \! k \\
\end{array}
\end{small}
\right.
\label{Eq:JacobianJointSecDerCol}
\end{align}
where $\bm{P}_\Delta(\bm{J}^j)$ and $\bm{M}_\Delta(\bm{J}^k)$ are defined as in \eqref{Eq:JacPmatrix} and \eqref{Eq:JacMmatrix}, respectively. The time derivative of the force manipulability Jacobian $\bm{\mathcal{J}}^{\bm{F}}$ and the manipulability Jacobian $\bm{\mathcal{J}}^{\ddot{\bm{x}}}$ corresponding to the dynamic manipulability ellipsoid can be computed symbolically in a similar way using Eqs.~\eqref{Eq:JacobianTimeDer} and ~\eqref{Eq:JacobianTimeJointDer}. Moreover, their w.r.t. joint angles can be computed symbolically using the chain rules, Eqs.~\eqref{Eq:JacobianJointDerCol} and~\eqref{Eq:JacobianJointSecDerCol}.

\end{appendices}

\end{document}